\documentclass[lettersize,journal]{IEEEtran}
\usepackage{amsmath,amsfonts}
\usepackage{amssymb}
\usepackage{algorithmic}
\usepackage{algorithm}
\usepackage{array}
\newcolumntype{P}[1]{>{\centering\arraybackslash}p{#1}}
\newcolumntype{L}[1]{>{\raggedright\arraybackslash}p{#1}}
\usepackage{textcomp}
\usepackage{stfloats}
\usepackage{url}
\usepackage{verbatim}
\usepackage{graphicx}
\usepackage{cite}
\usepackage{xcolor}
\usepackage{multirow}
\usepackage{subcaption}
\usepackage{booktabs}
\usepackage[colorlinks=true]{hyperref}
\begin{document}

\title{End-to-End 3-D Spatiotemporal Perception with Multimodal Fusion and V2X Collaboration}

\author{Zhenwei Yang$^{1}$,Yibo Ai$^{1,*}$,Weidong Zhang$^{1,*}$ \\
\thanks{$^{*}$is the corresponding author. Contact: wdzhang@ustb.edu.cn; zwdpaper@163.com; ybai@ustb.edu.cn.}
\thanks{$^{1}$National Center for Materials Service Safety (NCMS), University of Science and Technology Beijing (USTB), Beijing 100083, China.}
\thanks{Copyright \copyright~2025 IEEE. Personal use of this material is permitted. 
However, permission to use this material for any other purposes must be obtained from the IEEE by sending a request to pubs-permissions@ieee.org.}
}

\maketitle

\begin{center}
    \small
    \textit{This work has been accepted for publication in IEEE Internet of Things Journal.} \\
    \textit{DOI: \href{https://doi.org/10.1109/JIOT.2026.3694808}{10.1109/JIOT.2026.3694808}}
\end{center}

\vspace{1cm}

\begin{abstract}
    Multiview cooperative perception and multimodal fusion are essential for reliable 3-D spatiotemporal understanding in autonomous driving, especially in cases with occlusions, limited viewpoints, and communication delays in vehicle-to-everything (V2X) scenarios.
    In this paper, Cross-modal End-to-End Tracking for V2X (XET-V2X), a multimodal fused end-to-end tracking framework for V2X collaboration that unifies multiview multimodal sensing within a shared spatiotemporal representation, is proposed.
    To efficiently align heterogeneous viewpoints and modalities, XET-V2X introduces a dual-layer spatial cross-attention module based on multiscale deformable attention.
    Multiview image features are aggregated to enhance semantic consistency, followed by point cloud fusion guided by the updated spatial queries, enabling effective cross-modal interaction while reducing computational overhead.
    Experiments based on the real-world V2X Sequential Perception Dataset (V2X-Seq-SPD) dataset and 
    two simulated V2X-Sim-derived subsets, namely the vehicle-to-vehicle (V2X-Sim-V2V) and vehicle-to-infrastructure (V2X-Sim-V2I) subsets, 
    demonstrate consistent improvements in detection and tracking performance under varying communication delays,
    with XET-V2X achieving up to 15--20\% relative gains in mean average precision (mAP) and average multi-object tracking accuracy (AMOTA) over single-view or single-modal baselines, while also outperforming representative tracking-by-detection cooperative perception methods.
\end{abstract}


\section{Introduction}
3-D spatiotemporal perception has become a foundational capability for intelligent systems as advances in artificial intelligence, 
IoT infrastructures, and multisource sensing continue to reshape autonomous perception. 
By jointly modeling spatial structure and temporal evolution, 3-D spatiotemporal perception enables a continuous understanding of object geometry,
motion trajectories, and behavioral patterns. This paradigm has proven essential for autonomous driving, 
where safety-critical applications such as 3-D detection, multiobject tracking, trajectory forecasting,
and motion planning require reliable scene representations over time.

A large body of prior research has explored improvements in both spatial and temporal aspects of perception. 
Single-vehicle systems that fuse data from cameras, LiDAR, and inertial sensors have demonstrated strong performance in static reconstruction, 
dynamic obstacle detection, and sequential prediction. 
Moreover, temporal fusion techniques, particularly Transformer-based bird's-eye-view (BEV) models, have significantly enhanced multiframe detection stability,
cross-frame feature alignment, and long-term trajectory reasoning. 
However, most existing temporal BEV models are designed for single-vehicle perception and do not explicitly address cross-agent collaboration under communication latency.
Beyond single-agent sensing, multiview cooperative perception in vehicle-to-everything (V2X) scenarios has emerged as a complementary strategy to overcome occlusions and restricted fields of view.
By exchanging raw sensor observations, intermediate features, or high-level perceptual results across connected automated vehicles (CAVs) and roadside units (RSUs), 
cooperative systems have achieved an extended perception range and reliable detection of hidden objects.
Representative studies have focused on cooperative LiDAR perception~\cite{chen2019cooper,chen2019f},
communication-efficient feature sharing~\cite{wang2020v2vnet,hu2022where2comm}, and large-scale benchmarks such as DAIR-V2X~\cite{yu2023v2x}, V2X-Sim~\cite{li2022v2x}, and V2V4Real~\cite{yu2023v2x,li2022v2x,xu2023v2v4real},
which have accelerated progress in collaborative perception research.

\begin{figure}[t]
	\centering
	\includegraphics[width=0.48\textwidth]{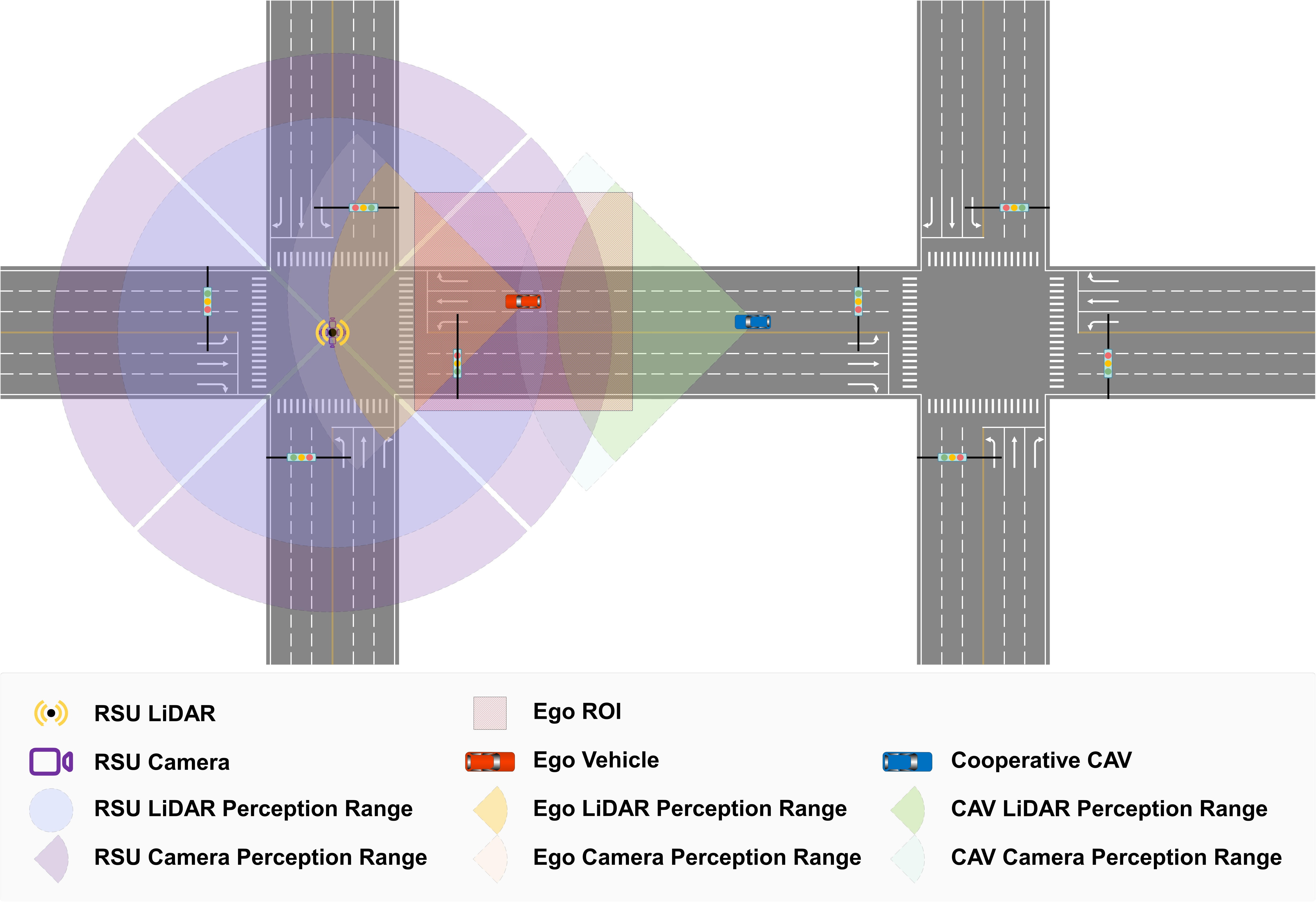}
    \caption{\textbf{Vehicle-to-Everything (V2X) Cooperative Perception Diagram.}
    The red vehicle denotes the ego vehicle, while the blue vehicles represent cooperative CAVs. 
    Both the ego vehicle and cooperative CAVs are equipped with forward-facing fan-shaped cameras and LiDAR sensors with front-view coverage. 
    The RSU is equipped with fan-shaped cameras covering the four approaches of the intersection and a $360^\circ$ LiDAR sensor, providing comprehensive infrastructure-side perception.}
	\label{fig:xet_v2x_diagram}
\end{figure}

Despite these advances, several limitations remain. Single-vehicle perception still suffers from issues related to view occlusion, blind spots,
and geometric distortions that hinder reliable understanding in dense urban environments. 
Existing cooperative approaches often adopt tracking-by-detection pipelines,
where multiview detection and temporal association are optimized in separate stages.
Such decoupled designs introduce redundant computations,
limit cross-stage information flow and may result in error accumulation across the detection, fusion, and tracking stages.
Moreover, most V2X research has focused on either spatial fusion or modality fusion and rarely integrated temporal modeling into a unified cross-agent framework.
Heterogeneous sensors across agents further complicate cross-view geometric alignment, spatiotemporal synchronization, 
and feature consistency, particularly under communication constraints or domain shifts. 
These challenges collectively restrict the robustness, continuity, and scalability of the current 3-D spatiotemporal perception systems.

To address these gaps, we propose a unified end-to-end framework that jointly integrates multiview collaboration, 
multimodal fusion, and temporal modeling within a single learnable architecture. 
Unlike conventional tracking-by-detection pipelines, the proposed framework performs spatial perception, cross-agent fusion, 
and temporal reasoning tasks in a unified optimization process, as illustrated in Fig.~\ref{fig:xet_v2x_diagram}.
By jointly exploiting complementary geometric cues from distributed LiDAR systems and semantic information from multiple cameras,
the proposed framework establishes a shared BEV representation that supports cross-view consistency and long-horizon motion reasoning. 
A dual-layer cross-modal and cross-view attention mechanism enables fine-scale interaction between heterogeneous features, 
and Transformer-based temporal encoding captures long-term trajectories and mitigates occlusion-induced discontinuities.
Extensive experiments based on the V2X-Seq-SPD~\cite{yu2023v2x} and V2X-Sim~\cite{li2022v2x} benchmarks demonstrate substantial improvements in both detection and tracking performance over single-view, unimodal, and tracking-by-detection cooperative perception baselines.

The contributions of this study are summarized as follows:
\begin{itemize}
    \item We propose a unified end-to-end 3-D spatiotemporal perception framework that jointly integrates multiview collaboration, multimodal fusion, and temporal modeling within a single optimization process.
    \item We design a dual-layer cross-modal-cross-view interaction module and a shared BEV representation that achieve robust geometric alignment and complete information transfer across distributed agents.
    \item We demonstrate state-of-the-art performance for large-scale V2X benchmarks and show consistent improvements over single-view perception baselines, unimodal perception systems, and tracking-by-detection cooperative perception methods.
\end{itemize}

The remainder of this paper is organized as follows. Section II provides a review of the related works relevant to this study.
In Section III, we present XET-V2X, our proposed multimodal fused end-to-end tracking framework for V2X collaboration.
Section IV details the experimental results and analysis. Finally, Section V concludes the paper.

\section{Related Works}
\subsection{Multiview Cooperative Perception}
Multiview cooperative perception in V2X scenarios aggregates observations from spatially distributed agents to extend the perception range, 
mitigate occlusions, and improve scene understanding under limited viewpoints. 
Early studies primarily explored cooperative warning systems~\cite{seeliger2014advisory,kim2020shared}, 
raw sensor sharing~\cite{ye2020cooperative,arnold2020cooperative}, intermediate feature fusion~\cite{xiao2018multimedia}, 
and decision-level collaboration~\cite{yee2018collaborative,miller2020cooperative,arnold2020cooperative}. 
Representative approaches include point cloud sharing~\cite{chen2019cooper,chen2019f} and neural communication frameworks such as V2VNet~\cite{wang2020v2vnet} and Where2Comm~\cite{hu2022where2comm}. 
From the perspective of information exchange, existing cooperative perception methods can generally be grouped into raw-data-level, 
feature-level, and object-level collaboration, corresponding to different trade-offs among information fidelity, communication cost, and deployment complexity. 
Among them, feature-level collaboration has become the dominant paradigm because it preserves rich contextual information while remaining substantially more communication-efficient than raw sensor sharing. 
To further address practical bandwidth constraints, communication-efficient strategies such as dynamic communication, sparse region selection, 
and information filling with codebooks have also been proposed~\cite{liu2020who2com,liu2020when2com,hu2022where2comm,hu2024communication}.

Existing multiview modeling paradigms generally fall into BEV-based and query-based methods. 
In BEV-based methods, such as BEVDet~\cite{huang2021bevdet} and BEVDepth~\cite{li2023bevdepth}, perspective image features are projected into a unified BEV space. 
In cooperative settings, PillarGrid~\cite{bai2022pillargrid} extends this idea by fusing LiDAR data from connected vehicles and infrastructure. 
Alternatively, Transformer-based methods introduce learnable queries for cross-view reasoning. 
Object-query-based approaches such as PETR~\cite{liu2022petr}, PETRv2~\cite{liu2023petrv2}, and DETR3D~\cite{wang2022detr3d} directly predict 3-D objects from sparse queries, 
while BEV-query-based frameworks such as the BEVFormer series~\cite{li2024bevformer,yang2023bevformer} construct spatiotemporal BEV representations through learnable BEV queries. 
Meanwhile, public benchmarks such as DAIR-V2X~\cite{yu2022dair}, OpenV2V~\cite{xu2022opv2v}, V2X-Sim\cite{li2022v2x}, V2X-Seq\cite{yu2023v2x}, and V2V4Real~\cite{xu2023v2v4real} have substantially accelerated research in this area.

Despite this progress, practical V2X collaboration still faces several key challenges, 
including pose and calibration errors, communication latency, asynchronous observations, and heterogeneous sensing configurations. 
Recent studies have therefore begun to move beyond idealized homogeneous settings toward more realistic collaborative perception, 
for example through latency-aware synchronization~\cite{lei2022latency}, feature-flow-based temporal compensation~\cite{yu2023flow}, 
asynchronous multiagent detection~\cite{dao2024practical}, and extensible heterogeneous perception~\cite{lu2024heal}. 
Nevertheless, most existing approaches remain modular and primarily focus on spatial collaboration, with limited support for unified spatiotemporal optimization across agents.

\subsection{Multimodal Fusion Perception}
Multimodal 3-D object detection leverages heterogeneous sensors to exploit the complementary strengths of cameras and LiDAR. 
Camera images provide rich semantic, texture, and appearance cues but are sensitive to illumination changes and lack reliable absolute depth. 
LiDAR directly captures 3-D geometric structure with accurate metric depth, 
yet its point clouds are sparse and relatively weak in semantic representation. 
Therefore, the core problem of multimodal fusion lies in how to achieve stable alignment, effective interaction, 
and robust joint reasoning across heterogeneous data structures, sampling densities, and coordinate systems.

According to the fusion stage, existing methods can generally be categorized into early fusion, intermediate fusion, and late fusion. 
Early fusion methods enrich point clouds with image-derived semantic cues before deep 3-D feature extraction, 
as implemented in PointPainting~\cite{vora2020pointpainting}, MVXNet~\cite{sindagi2019mvx}, AVOD~\cite{ku2018joint}, and PointAugmenting~\cite{wang2021pointaugmenting}. 
Intermediate fusion methods allow deeper interaction between modality-specific representations to refine detections and alleviate misalignment issues, 
such as CenterFusion~\cite{nabati2021centerfusion}, AutoAlignV2~\cite{chen2022deformable}, and DeepFusion~\cite{li2022deepfusion}. 
Late or decision-level fusion methods combine high-level outputs from different modalities after independent perception, with CLOCs~\cite{pang2020clocs} being a representative example. 
These three strategies reflect different trade-offs among information preservation, robustness, and engineering complexity.

More recently, multimodal fusion has increasingly been reformulated as an end-to-end learning problem. 
Query-based Transformers such as TransFusion~\cite{bai2022transfusion}, FUTR3D~\cite{chen2023futr3d}, 
and DeepInteraction~\cite{yang2022deepinteraction} model fusion through shared object queries, 
while BEV-based fusion frameworks such as BEVFusion~\cite{liu2023bevfusion} further promote multimodal interaction in a unified BEV representation space. 
Subsequent studies such as CMT~\cite{yan2023cross}, UniBEV~\cite{wang2024unibev}, and GAFusion~\cite{li2024gafusion} continue to improve cross-modal interaction, 
unified BEV modeling, and robustness under incomplete or challenging sensing conditions. 
Although these methods have significantly improved single-agent multimodal perception, 
aligning heterogeneous sensor representations across modalities and viewpoints remains challenging in dynamic driving scenarios. 
Recent efforts in V2X multimodal collaboration, such as MDNet~\cite{he2025mdnet}, have begun to investigate this issue under collaborative settings. 
However, seamlessly extending multimodal alignment into a unified spatiotemporal tracking framework remains an open problem.

\subsection{Temporal Fusion and End-to-End Tracking}
Temporal 3-D perception introduces the time dimension on top of spatial scene understanding and aims to model the continuous evolution of object states in dynamic environments. 
Compared with single-frame 3-D perception, temporal perception must additionally maintain stable inter-frame associations,
 motion constraints, and state updates, thereby producing a continuous and temporally consistent scene representation. 
Existing methods can be broadly grouped into tracking-by-detection, joint detection-and-embedding, and end-to-end tracking paradigms.

Tracking-by-detection methods first detect objects frame by frame and then associate them through a separate tracking module. 
In 2D vision, representative methods include SORT~\cite{bewley2016simple} and DeepSORT~\cite{wojke2017simple}. 
In 3-D autonomous driving, AB3DMOT~\cite{weng2020ab3dmot}, CBMOT~\cite{benbarka2021score}, 
and SimpleTrack~\cite{pang2022simpletrack} provide strong and efficient baselines by combining 3-D detection with explicit motion modeling and data association. 
Joint detection-and-embedding methods further reduce the separation between detection and identity modeling by learning them in a shared framework. 
Representative works include JDE\cite{wang2020towards}, FairMOT~\cite{zhang2021fairmot}, CenterTrack~\cite{yin2021center}, OGR3MOT~\cite{zaech2022learnable}, MotionTrack~\cite{zhang2023motiontrack}, and ADA-Track~\cite{ding2024ada}. 
Although these methods improve efficiency and reduce hand-crafted association rules, their trajectory generation still relies on relatively explicit matching or motion-based post-processing.

End-to-end tracking methods further integrate trajectory maintenance and state propagation into a unified network. 
Transformer-based frameworks such as TrackFormer~\cite{meinhardt2022trackformer} and MOTR~\cite{zeng2022motr} explicitly propagate track queries across frames to achieve implicit identity continuation and state updating. 
In 3-D perception, temporal fusion has also been explored through sequential aggregation~\cite{piergiovanni20214d}, recurrent BEV updates~\cite{chang2024recurrentbev}, and joint V2X spatiotemporal fusion~\cite{zhou2024v2xpnp}. 
Related works such as BEVFormer~\cite{li2024bevformer,yang2023bevformer}, PETRv2~\cite{liu2023petrv2}, StreamPETR~\cite{wang2023exploring}, 
and Sparse4D v3~\cite{lin2023sparse4dv3} further demonstrate the potential of unified BEV or query-based representations for end-to-end temporal modeling. 
In the context of vehicle-infrastructure cooperation, recent studies have also begun to explore end-to-end temporal perception using LiDAR sequences~\cite{yang2025letvic}.

Despite substantial progress in multiview collaboration, multimodal fusion, and temporal perception, 
most existing methods still address these three aspects in isolation. 
They typically focus on either cooperative spatial perception, single-agent multimodal detection, or temporal modeling alone, 
without explicitly integrating them into a unified V2X spatiotemporal framework under realistic communication constraints. 
In contrast, our work introduces an end-to-end 3-D spatiotemporal perception framework that jointly performs cooperative perception, 
multimodal sensor fusion, and temporal aggregation within a unified learning architecture, 
thereby ensuring consistent perception across both agents and time.

\begin{figure*}[t]
	\centering
	\includegraphics[width=1.0\textwidth]{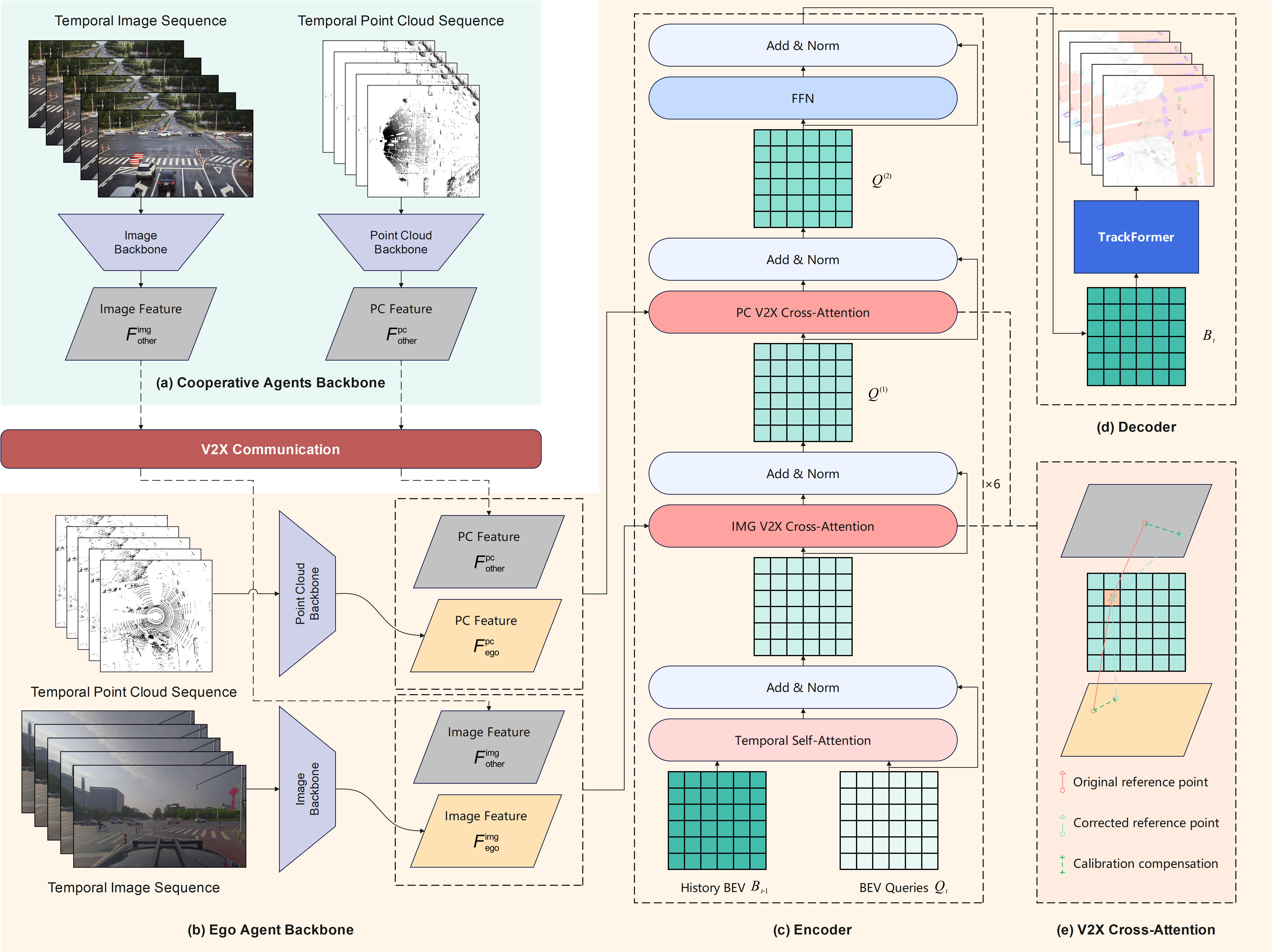}
    \caption{\textbf{Architecture of XET-V2X.} 
    The diagram illustrates the steps involved in the multiview multimodal cooperative perception framework.
    (a) and (b) Both the ego vehicle and cooperative agents (connected automated vehicles (CAVs) and roadside units (RSUs)) employ PointPillars to extract LiDAR point cloud features and ResNet-101 with a feature pyramid network (FPN) to extract multiscale image features.
    The intermediate feature-level representations from cooperative agents are then transmitted to the ego vehicle via vehicle-to-everything (V2X) communication.
    (c) and (e) Multiview multimodal fusion is implemented through a dual-layer V2X cross-attention module based on multiscale deformable attention.
    This module progressively fuses image features first, followed by point cloud features,
    and a calibration error compensation (CEC) module explicitly corrects spatial misalignment issues across heterogeneous viewpoints.
    (d) The decoder performs joint 3-D object detection and tracking on the basis of the MOTR framework in an end-to-end manner.}
	\label{fig:xet-v2x-architecture}
\end{figure*}

\section{Method}
In this section, the proposed multiview cooperative spatiotemporal perception model with multimodal fusion, termed \textbf{XET-V2X}, is presented.
Building upon an end-to-end 3-D temporal perception framework, XET-V2X integrates a multiview cooperative mechanism to dynamically fuse multimodal sensor data collected from an ego vehicle and other cooperative agents.
This design explicitly addresses practical challenges such as sensor calibration errors, the limited representational capacity of single-modality sensors, and the restricted field of view of individual sensing agents.
By jointly exploiting multiview, multimodal, and temporal information, XET-V2X achieves accurate object detection and tracking, thereby enhancing environmental perception and enabling safe and efficient autonomous driving.

\subsection{Task Formulation}\label{subsec:task_formulation}
In this work, a V2X-based multiview cooperative spatiotemporal perception task for autonomous driving is explored,
where an ego vehicle collaborates with other agents, such as CAVs and RSUs,
to perform 3-D object detection and tracking using temporally ordered multimodal sensor data.
Each agent is equipped with heterogeneous sensors, including cameras and LiDAR systems,
providing complementary geometric and semantic observations from diverse viewpoints.

Unlike single-vehicle perception, cooperative perception involves the explicit exploitation of spatial complementarity and temporal continuity across distributed sensing agents.
By sharing perception-relevant information through V2X communication, the system aims to mitigate occlusion, extend perception range, 
and enhance the spatiotemporal consistency of object states in complex traffic environments.

\subsubsection{Temporal Modeling and Notation}
After preprocessing, sensor observations from the ego vehicle and other cooperative agents are organized into a temporal sequence of length $N$,
indexed by $k = 1,2,\dots,N$.
Let $t_{\mathrm{ego},k}$ and $t_{\mathrm{other},k}$ denote the acquisition times of the ego-view and other-view observations for index $k$, respectively.
The temporal indices satisfy a strictly increasing order for each agent,
and $t_{\mathrm{other},k} \leq t_{\mathrm{ego},k}$ holds because of the communication latency in V2X transmission.

We assume that each agent internally synchronizes its onboard sensors (e.g., camera and LiDAR),
ensuring that multimodal data collected by the same agent at index $k$ correspond to the same timestamp.
This assumption is consistent with practical vehicular and RSU sensing systems
and does not require strict interagent time alignment.

\subsubsection{Input Representation}
At time index $k$, the cooperative perception system receives the following:
\begin{itemize}
    \item Ego-view multimodal observations
    $(S_{\mathrm{ego},k}^\mathrm{pc}, S_{\mathrm{ego},k}^\mathrm{img})$,
    together with the corresponding calibration parameters
    $(M_{\mathrm{ego},k}^\mathrm{pc}, M_{\mathrm{ego},k}^\mathrm{img})$;
    \item Multimodal observations from other cooperative agents
    $(S_{\mathrm{other},k}^\mathrm{pc}, S_{\mathrm{other},k}^\mathrm{img})$,
    with calibration parameters
    $(M_{\mathrm{other},k}^\mathrm{pc}, M_{\mathrm{other},k}^\mathrm{img})$,
    transmitted to the ego vehicle via V2X communication.
\end{itemize}

These observations jointly encode geometric structure and semantic appearance from multiple viewpoints over time,
forming the basis for cooperative spatiotemporal perception.

\subsubsection{Output Definition}
At ego time $t_{\mathrm{ego},k}$, the system outputs a set of detected and tracked objects:
\begin{equation}
    \mathrm{O}_k=
    \Bigl\{\bigl(c_{i,k},\, p_{i,k},\, d_{i,k},\, \theta_{i,k},\, \mathcal{T}_i \bigr)\Bigr\}_{i=1}^{\mathcal{D}_k},
\end{equation}
where $c_{i,k}$ denotes the object category,
$p_{i,k}$ is its 3-D position,
$d_{i,k}$ is its 3-D dimensions,
$\theta_{i,k}$ is its orientation,
and $\mathcal{T}_i$ is a persistent tracking identity.
$\mathcal{D}_k$ denotes the number of objects perceived at time $k$.

\subsubsection{Ground Truth and Training Objective}
The ground truth at time $t_{\mathrm{ego},k}$ is based on aggregated annotations from all the cooperative agents and sensing modalities,
restricted to the egocentric region of interest $\mathrm{ROI}_{\mathrm{ego},k}$:
\begin{equation}
    \mathrm{GT}_k=
    \Bigl(
    \mathrm{GT}_{\mathrm{ego},k}^\mathrm{pc} \cup
    \mathrm{GT}_{\mathrm{ego},k}^\mathrm{img} \cup
    \mathrm{GT}_{\mathrm{other},k}^\mathrm{pc} \cup
    \mathrm{GT}_{\mathrm{other},k}^\mathrm{img}
    \Bigr)
    \cap
    \mathrm{ROI}_{\mathrm{ego},k},
\end{equation}
which can be equivalently represented as follows:
\begin{equation}
    \mathrm{GT}_k=
    \bigl\{(c_{i,k},\, p_{i,k},\, d_{i,k},\, \theta_{i,k},\, \mathcal{T}_i)\bigr\}_{i=1}^{\mathcal{G}_k},
\end{equation}
where $\mathcal{G}_k$ is the number of ground-truth objects at time $k$.

Based on $\{\mathrm{O}_k\}_{k=1}^N$ and $\{\mathrm{GT}_k\}_{k=1}^N$,
the cooperative perception process jointly optimizes the detection accuracy, localization precision,
and temporal identity consistency under multiview and multiagent settings.

\subsection{Motivation and Design Rationale for Multiview Cooperative Multimodal Fusion}
As formulated in Section~\ref{subsec:task_formulation}, the objective of V2X-based cooperative spatiotemporal perception is to jointly implement 3-D object detection and tracking by leveraging temporally ordered multimodal observations from distributed sensing agents.
Compared with single-vehicle perception, this task introduces two fundamental challenges:
(i) effectively exploiting spatial complementarity across heterogeneous viewpoints and
(ii) maintaining consistent object representations over time in cases with communication latency and sensing uncertainty.

From a spatial perspective, perception from a single viewpoint is inherently limited by occlusion, a restricted field of view, and range constraints.
Even with multicamera or LiDAR-camera setups onboard a single vehicle, blind spots and long-range perception degradation remain unavoidable,
especially in complex traffic scenarios, such as those at intersections and in dense urban environments.
V2X-based multiview collaboration enables an ego vehicle to incorporate complementary observations from RSUs and CAVs,
thereby extending perception coverage and alleviating occlusions through viewpoint diversity.
This spatial redundancy forms the basis for complete and reliable 3-D scene understanding.

From a modal perspective, LiDAR and camera sensors exhibit fundamentally different yet complementary characteristics.
LiDAR provides accurate 3-D geometry and depth measurements, which are critical for precise localization and size estimation,
while cameras capture valuable appearance, texture, and semantic cues that facilitate object classification and attribute recognition.
Relying on a single modality leads to inherent vulnerabilities under adverse lighting or weather conditions.
By fusing LiDAR and camera data across multiple cooperative agents, the perception system is robust to modality-specific degradation,
making it suitable for real-world autonomous driving scenarios.

From a temporal perspective, object detection and tracking are intrinsically dynamic tasks.
Isolated framewise perception is insufficient for ensuring identity consistency, motion continuity, and stable state estimation.
Temporal modeling allows the system to exploit motion patterns, smooth noisy observations,
and associate objects across time in cases of asynchronous data arrival in V2X communication.
Importantly, cooperative temporal fusion does not require strict interagent synchronization;
instead, bounded communication delays can be flexibly considered while preserving spatiotemporal consistency for the ego vehicle.

On the basis of these considerations, a multiview cooperative multimodal fusion paradigm
in which temporally ordered LiDAR and camera observations from an ego vehicle and other V2X agents are jointly modeled is adopted.
The design explicitly integrates spatial complementarity, modality diversity, and temporal continuity,
forming a unified perception framework that addresses the core limitations of single-view, single-modality, and frame-centric approaches.
This motivation is the basis of the architectural choices presented in Sections~\ref{subsec:model_architecture}--\ref{subsec:e2e_3d_det_and_trk_decoder}.

\subsection{Model Architecture}\label{subsec:model_architecture}
The overall architecture of XET-V2X is illustrated in Fig.\ref{fig:xet-v2x-architecture}.
The core design of the model lies in multiview multimodal feature extraction and multiview multimodal feature fusion.
For both the ego view and other cooperative views (i.e., RSUs or CAVs), 
XET-V2X employs PointPillars~\cite{lang2019pointpillars} to extract point cloud features and ResNet~\cite{he2016deep} to extract image features, yielding high-dimensional feature representations.
Two stacked multiview cooperative cross-attention modules are subsequently applied to fuse features across viewpoints and modalities.
Finally, object detection and tracking are performed on the basis of the MOTR framework~\cite{zeng2022motr}, enabling joint detection and tracking in an end-to-end manner.

\subsection{Multiview Multimodal Feature Extraction}\label{subsec:multiview_multimodal_feature_extraction}

\subsubsection{Multiview Point Cloud Feature Extraction}
\begin{figure}[t]
	\centering
	\includegraphics[width=0.24\textwidth]{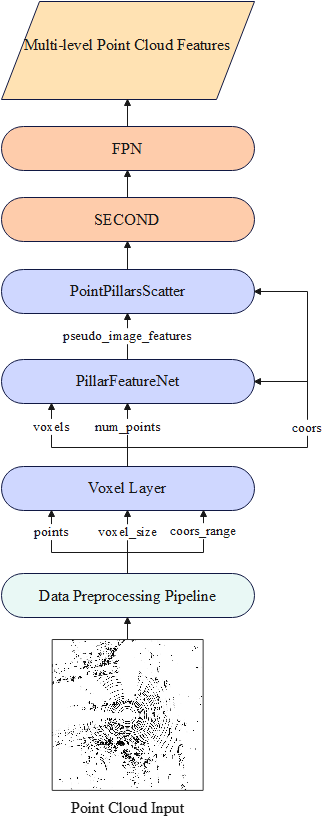}
    \caption{\textbf{Point Cloud Backbone.} The point cloud backbone extracts bird's-eye-view (BEV) features from point cloud data.}
	\label{fig:pc-backbone}
\end{figure}

Fig.\ref{fig:pc-backbone} illustrates the point cloud feature encoding backbone for an ego vehicle and cooperative agents in the XET-V2X framework,
corresponding to the architectures shown in Fig. \ref{fig:xet-v2x-architecture}~(a) and (b).
This backbone includes a hierarchical feature encoding pipeline that transforms raw point clouds into multiscale BEV feature representations.
Specifically, raw point cloud data are first preprocessed and then voxelized into a regular 3-D grid structure.
A pillar-based feature encoding module (PillarFeatureNet) is subsequently employed to extract geometric features from nonempty pillars.
These extracted sparse features are then distributed into a dense 2D grid space via a scatter operation (PointPillarsScatter) to form a BEV pseudoimage representation.
Finally, a 2D SECOND-based backbone comprising standard dense convolutional layers and a feature pyramid network (FPN) are applied to perform multiscale feature aggregation and enhancement.
A hierarchical feature pyramid with varying receptive fields is then constructed to improve perception for objects of different sizes.

In this work, efficient point cloud processing is implemented on the basis of the PointPillars framework~\cite{lang2019pointpillars},
which significantly reduces computational complexity while preserving essential 3-D geometric information through pillarwise partitioning (pillar size: $0.2\,\mathrm{m} \times 0.2\,\mathrm{m} \times 8\,\mathrm{m}$).
As illustrated in the processing pipeline, the raw 3-D point cloud $(x, y, z)$ is first preprocessed, followed by voxelization within modality-specific regions of interest (ROIs) defined according to sensor placement:
\begin{itemize}
    \item Ego-vehicle ROI in the V2X-Seq-SPD dataset~\cite{yu2023v2x}: $x \in [-51.2, 51.2]$\,m, $y \in [-51.2, 51.2]$\,m, $z \in [-5.0, 3.0]$\,m;
    \item Roadside ROI in the V2X-Seq-SPD dataset~\cite{yu2023v2x}: $x \in [0, 102.4]$\,m, $y \in [-51.2, 51.2]$\,m, $z \in [-5.0, 3.0]$\,m;
    \item Ego-vehicle ROI in the V2X-Sim dataset~\cite{li2022v2x}: $x \in [-51.2, 51.2]$\,m, $y \in [-51.2, 51.2]$\,m, $z \in [-3.0, 5.0]$\,m;
    \item Other-agent ROI in the V2X-Sim dataset~\cite{li2022v2x}: $x \in [-51.2, 51.2]$\,m, $y \in [-51.2, 51.2]$\,m, $z \in [-3.0, 5.0]$\,m.
\end{itemize}

The voxel size is set to $0.2$\,m along the $x$ and $y$ axes and $8$\,m along the $z$ axis.
Accordingly, both ego-view and other-view BEV grids have a spatial resolution of $512 \times 512$.
Each voxel stores the points falling within it along with their corresponding coordinate information.

During feature encoding, PillarFeatureNet employs multilayer perceptrons (MLPs) to encode geometric features and transform unstructured point cloud data into structured sparse feature pillars.
The PointPillarsScatter module then distributes these sparse features into a dense 2D grid space via a scatter operation, forming a BEV pseudoimage.
A 2D SECOND-based backbone comprising standard dense convolutional layers is subsequently applied to efficiently extract high-level spatial semantics from the pseudoimage.
Finally, the FPN performs top-down fusion with lateral connections to produce a four-level feature pyramid $\{P1, P2, P3, P4\}$, corresponding to perception resolutions of $0.4$\,m, $0.8$\,m, $1.6$\,m, and $3.2$\,m, respectively.
This design effectively leverages highly optimized 2D convolution steps to achieve an optimal balance between computational efficiency and multiscale perception capability.

\subsubsection{Multiview Image Feature Extraction}
\begin{figure}[t]
	\centering
	\includegraphics[width=0.24\textwidth]{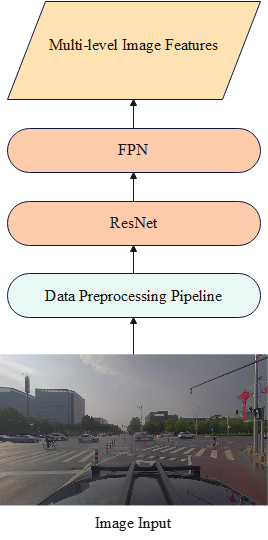}
    \caption{\textbf{Image Backbone.} The image backbone extracts multilevel features from camera images.}
	\label{fig:img-backbone}
\end{figure}

Fig.\ref{fig:img-backbone} shows the image feature encoding backbone for both vehicle-mounted and roadside cameras in the XET-V2X framework, 
corresponding to the architectures illustrated in Fig.\ref{fig:xet-v2x-architecture}~(a) and (b).
This backbone follows a hierarchical feature encoding paradigm that converts raw images into multiscale feature representations.
Specifically, the input images are first subjected to standard preprocessing and augmentation operations, including resizing, normalization, and padding.
A residual network (ResNet) is then employed to extract feature maps at different semantic levels.
An FPN subsequently fuses multilevel features through top-down pathways and lateral connections,
generating a set of multiscale image features with unified channel dimensions and varying spatial resolutions to improve the detection of objects at different scales.

In XET-V2X, the image branch is initialized from pre-trained visual backbone weights and is built based on the ResNet-101 architecture~\cite{he2016deep}, 
with multiscale feature fusion realized via an FPN.
During preprocessing, all the input images are resized, normalized, and padded to match the network input requirements.
The ResNet-101 backbone consists of four residual stages, from which feature maps C3, C4, and C5 (with channel dimensions of 512, 1024, and 2048, respectively) are extracted from the second, third, and fourth stages (i.e., conv3\_x, conv4\_x, and conv5\_x, respectively).
Deformable convolutions (DCNv2) are incorporated into the conv4\_x and conv5\_x stages to increase robustness to geometric deformations. 
To stabilize training and reduce the memory footprint, the entire image backbone is frozen during the learning process. 
In contrast, the image FPN remains trainable and is optimized jointly with the LiDAR branch, the temporal and multiview fusion modules, and the final decoder in the end-to-end training framework. 
The FPN module subsequently performs top-down channel mapping and spatial interpolation with lateral connections, producing four pyramid levels $\{P3, P4, P5, P6\}$ with unified channel dimensions.
These multiscale features provide rich representations ranging from fine-scale to coarse-level semantics for downstream detection and segmentation heads,
effectively improving performance for objects of varying sizes while maintaining computational efficiency.

\subsection{Multiview Multimodal Feature Transmission}\label{subsec:multiview_multimodal_feature_transmission}
\label{subsec:xet-v2x-feature-transmission}
To enable cooperative perception in cases with practical V2X communication constraints,
XET-V2X adopts feature-level transmission as the primary information exchange mechanism among distributed agents.
As illustrated in Fig.~\ref{fig:xet-v2x-architecture}~(a) and (b),
image and point cloud features extracted by remote agents (e.g., CAVs or RSUs)
are transmitted to the ego vehicle for cooperative fusion.

In V2X systems, transmissible information can generally be categorized into three levels:
raw sensor data, instance-level perception results, and feature-level intermediate representations.
These forms are characterized by different trade-offs among communication bandwidth, information preservation, and privacy~\cite{fan2023quest,chen2023transiff}.
Raw sensor data (e.g., images, video streams, and point clouds) preserve complete scene information
but require extremely high communication bandwidth and raise severe privacy concerns.
Instance-level results, such as detected bounding boxes and semantic labels,
have minimal transmission cost but discard substantial contextual information.
Feature-level intermediate representations extracted by deep neural networks
provide a balanced solution by maintaining rich semantic and spatial cues
while significantly reducing transmission overhead when properly optimized~\cite{yu2023flow,chen2023transiff,hu2024communication}.

Motivated by the above considerations, XET-V2X adopts multiscale feature transmission for both image and LiDAR modalities. 
As summarized in Table~\ref{tab:communication_payload}, transmitting the complete set of four-scale FPN features corresponds to an upper-bound communication cost of approximately 66.75 MB/frame in uncompressed fp16 format, 
whereas transmitting only the higher-level semantic features (Scales 2--4) reduces the payload to approximately 16.49 MB/frame while retaining sufficient spatial context for collaborative fusion. 
Under the V2X-Seq-SPD setting (10 Hz), these two configurations correspond to approximately 667.5 MB/s and 164.9 MB/s, respectively; under the V2X-Sim benchmarks (5 Hz), the corresponding rates are approximately 333.75 MB/s and 82.45 MB/s. 
From the perspective of multi-agent scalability, the total uncompressed communication cost grows approximately linearly with the number of collaborating agents, i.e., $\approx N \times 16.49$ MB/frame under the practical Scales 2--4 transmission strategy in a standard broadcast-based setting.
Under this setting, the image feature payload (approximately 5.48 MB) is even smaller than that of an uncompressed raw RGB image (approximately 5.98 MB), 
indicating that feature-level transmission does not introduce extra communication burden for the image modality.

\begin{table}[t]
    \centering
    \caption{\textbf{Communication Payload of Multi-scale Features per Collaborating Agent.} Here, $C$ denotes the channel dimension of each feature map.}
    \label{tab:communication_payload}
    \begin{tabular}{L{0.08\textwidth}L{0.05\textwidth}P{0.08\textwidth}P{0.03\textwidth}P{0.06\textwidth}P{0.05\textwidth}}
        \toprule
        \textbf{Modality} & \textbf{Feature Scale} & \textbf{Spatial Resolution} & \textbf{$C$} & \textbf{Bytes / Element} & \textbf{Payload (MB)} \\
        \midrule
        Image FPN & Scale 1 & $136 \times 240$ & 256 & 2 (fp16) & $16.71$ \\
        Image FPN & Scale 2 & $68 \times 120$  & 256 & 2 (fp16) & $4.18$ \\
        Image FPN & Scale 3 & $34 \times 60$   & 256 & 2 (fp16) & $1.04$ \\
        Image FPN & Scale 4 & $17 \times 30$   & 256 & 2 (fp16) & $0.26$ \\
        \midrule
        LiDAR FPN & Scale 1 & $256 \times 256$ & 256 & 2 (fp16) & $33.55$ \\
        LiDAR FPN & Scale 2 & $128 \times 128$ & 256 & 2 (fp16) & $8.39$ \\
        LiDAR FPN & Scale 3 & $64 \times 64$   & 256 & 2 (fp16) & $2.10$ \\
        LiDAR FPN & Scale 4 & $32 \times 32$   & 256 & 2 (fp16) & $0.52$ \\
        \bottomrule
    \end{tabular}
\end{table}

Compared with raw data sharing and instance-level result sharing, feature-level transmission provides a more practical trade-off among information richness, privacy preservation, and communication efficiency. 
A key motivation is privacy: transmitting raw high-resolution images or point clouds may directly expose sensitive information such as faces, 
license plates, and detailed scene layouts, whereas intermediate neural features abstract these low-level details while preserving the semantic and geometric cues required for perception. 
In addition, for the LiDAR modality, transmitting features shifts the heavy backbone computation to the sender side and thereby reduces the computational burden on the receiving ego vehicle. 
Another important advantage is compressibility. The transmitted neural features are highly compatible with existing feature-compression methods. 
In particular, FFNet~\cite{yu2023flow} showed that a lightweight feature compression and transmission module can reduce the transmission cost of spatial feature flows from 243 Mb to 0.12 Mb. 
Because our framework also adopts feature-level communication, the FFNet compression module is directly applicable to our pipeline and can be used to further reduce the bandwidth required for sharing spatial features. 
This compatibility is especially important for bandwidth-limited multi-agent V2X deployment and further supports feature-level transmission as a practical design choice for scalable multiview multimodal perception.

By transmitting multimodal features and performing cross-agent interactions at the ego vehicle,
XET-V2X achieves efficient information sharing while preserving strong perception capability,
thereby supporting scalable multiview multimodal perception in distributed V2X systems.

\subsection{Multiview Multimodal Feature Fusion}\label{subsec:multiview_multimodal_feature_fusion}
Conventional multiview feature fusion methods face two major challenges.
First, the application of dense global attention mechanisms to high-dimensional features incurs substantial computational overhead.
Second, spatial calibration discrepancies across heterogeneous viewpoints and sensing modalities lead to feature misalignment,
which decreases the effectiveness of fusion.
These challenges are further amplified in V2X scenarios, where observations originate from multiple agents and multiple sensing modalities.
To address these issues, the proposed XET-V2X framework introduces a dual-layer spatial cross-attention module based on multiscale deformable attention,
enabling efficient and robust multiview multimodal feature fusion.

\subsubsection{Calibration Error Compensation}
Residual extrinsic calibration errors between agents can lead to cross-view feature misalignment in V2X cooperative perception.
To address this issue, we incorporate a lightweight calibration error compensation (CEC) mechanism
into the reference-point generation process of the proposed V2XCrossAttn module.
The CEC design is inspired by our prior vehicle--infrastructure cooperative temporal perception work (LET-VIC)~\cite{yang2025letvic},
and is further adapted in this work to support multi-view multimodal feature interaction.

For each BEV reference point $\mathrm{Ref}_i^{bev}$, the CEC learns per-point planar offsets for both the ego (vehicle) side and the other-agent side:
\begin{align}
\label{eq:cec-ego-error-offsets}
\begin{bmatrix}x_i^{ego}\\y_i^{ego}\end{bmatrix}=R_{bev2ego}\cdot{\mathrm{Ref}}_i^{bev}+T_{bev2ego}+\begin{bmatrix} {\Delta x}_i^{ego} \\ {\Delta y}_i^{ego} \end{bmatrix},
\end{align}
\begin{align}
\label{eq:cec-other-error-offsets}
\begin{bmatrix}x_i^{other}\\y_i^{other}\end{bmatrix}=R_{bev2other}\cdot{\mathrm{Ref}}_i^{bev}+T_{bev2other}+\begin{bmatrix} {\Delta x}_i^{other} \\ {\Delta y}_i^{other} \end{bmatrix}.
\end{align}
The compensated coordinates are then normalized to $[0, 1]$ according to the configured BEV ranges:
\begin{align}
\label{eq:cec-normalized-ego}
{P}_i^{ego}=\begin{bmatrix} \cfrac{x_i^{ego}-x_{min}^{ego}}{x_{max}^{ego}-x_{min}^{ego}} & \cfrac{y_i^{ego}-y_{min}^{ego}}{y_{max}^{ego}-y_{min}^{ego}} \end{bmatrix},
\end{align}
\begin{align}
\label{eq:cec-normalized-other}
{P}_i^{other}=\begin{bmatrix} \cfrac{x_i^{other}-x_{min}^{other}}{x_{max}^{other}-x_{min}^{other}} & \cfrac{y_i^{other}-y_{min}^{other}}{y_{max}^{other}-y_{min}^{other}} \end{bmatrix}.
\end{align}
Here, $\Delta x,\Delta y$ are learnable parameters optimized end-to-end together with the perception network. In XET-V2X, we apply CEC when computing the other-agent reference points $P^{other}$ (and optionally the ego reference points) used by deformable cross-attention.

\subsubsection{Deformable V2X Cross-Attention Process}
Given a set of BEV queries $Q^{bev}$, the proposed V2X cross-attention process aggregates features from the ego agent and other cooperative agents.
Each agent provides multimodal features, including image features and point cloud features.
Formally, for a specific BEV query $Q_q^{bev}$ and its corresponding reference points, the V2X cross-attention process is expressed as follows:
\begin{align}
\label{eq:v2x-cross-attention}
&\text{V2XCrossAttn}(Q_q^{bev},F^{ego},F^{other},P^{ego},P^{other}) \notag \\
&=\sum_{i=1}^{N_{ref}} \Bigg[\frac{1}{1+\mathcal{M}_{qi}^{other}}\Big(\text{MSDeformAttn}(Q_q^{bev}, P_{qi}^{ego}, F^{ego}) \notag \\
&+ \mathcal{M}_{qi}^{other}\cdot\text{MSDeformAttn}(Q_q^{bev}, P_{qi}^{other}, F^{other})\Big)\Bigg],
\end{align}
where $q$ is the index of the BEV query, $i$ is the index of the sampled 3-D reference points along the vertical pillar of the query, and $N_{ref}$ represents the total number of reference points per query.
$F^{ego}$ and $F^{other}$ represent the multimodal feature sets from the ego agent and other cooperative agents, respectively.
$P_{qi}^{ego}$ and $P_{qi}^{other}$ are the normalized coordinates of the $i$-th reference point for the $q$-th query projected to the corresponding feature spaces of the ego and cooperative agents, respectively.
$\mathcal{M}_{qi}^{other} \in \{0, 1\}$ is a validity mask that excludes reference points falling outside the spatial support of the other agents' features.

The core operation in Eq.~\ref{eq:v2x-cross-attention} is the multiscale deformable attention step (MSDeformAttn),
which efficiently aggregates sparse local features across multiple scales.
Given a query feature $Q_q$, a reference point $P_q$, and a multiscale feature map $\{F^l\}_{l=1}^L$, the operation is formulated as follows:
\begin{align}\label{eq:msdeformattn}
\text{MSDeformAttn}&(Q_q, P_q, \{F^l\}_{l=1}^L)=\sum_{m=1}^M W_m\Bigg[\sum_{l=1}^L \sum_{k=1}^K A_{mlqk} \notag \\
&\cdot W_m' F^l \big(\mathcal{R}_l(P_q) + \Delta p_{mlqk}\big)\Bigg],
\end{align}

where $m$ is the index of the attention heads, $l$ is the index of the feature levels, and $k$ is the index of the sampling points.
$\Delta p_{mlqk}$ and $A_{mlqk}$ denote the learnable sampling offsets and normalized attention weights, respectively, corresponding to the $q$-th query.
$\mathcal{R}_l(\cdot)$ is a function that rescales the normalized coordinates to the spatial resolution of the $l$-th feature level.
$W_m$ and $W_m'$ are learnable linear projection weights.
This formulation allows the attention mechanism to adaptively focus on informative regions while compensating for spatial misalignment between collaborative agents, achieving both efficiency and robustness.

\subsubsection{Asynchronous Temporal Compensation}
In practical V2X scenarios, cooperative perception inherently suffers from transmission delays and unaligned sensor sampling frequencies, leading to asynchronous observations.
Theoretically, our cooperative feature fusion does not require strict inter-agent hardware synchronization. Let $t_e$ and $t_c$ denote the timestamps of the ego vehicle and the collaborative agent, respectively,
with an asynchronous time gap $\Delta t = t_e - t_c$. While standard pose-based alignment can transform the cooperative features into the ego coordinate system, 
it cannot account for object motion during the time gap, resulting in a spatial displacement $\Delta d \approx v \cdot \Delta t$ for moving objects (where $v$ is the object's velocity) in the produced feature maps.

From a mathematical perspective, the MSDeformAttn operation (Eq.~\ref{eq:msdeformattn}) inherently mitigates the effects of this temporal delay without requiring explicit, heuristic motion compensation modules. 
Because MSDeformAttn dynamically predicts content-dependent sampling offsets $\Delta p_{mlqk}$ based on query features, 
the network implicitly learns to optimize these spatial offsets to approximate the temporal displacement ($\Delta p_{mlqk} \approx \Delta d$). 
Instead of relying on perfectly aligned features at $\Delta t = 0$, this data-driven spatial adaptability allows the network to adaptively adjust its receptive field to "search" for the shifted features of delayed moving objects, 
fundamentally relaxing the constraint of strict inter-agent synchronization.

It is worth noting that while asynchronous delays inevitably lead to a degradation in evaluation metrics due to information staleness at large time gaps, 
our experimental results demonstrate that the proposed framework is highly resilient. Within a practical delay range, 
the multiview cooperative multimodal fusion model consistently maintains superior perception performance compared to single-view or single-modal baselines.

\subsubsection{Progressive Multimodal Fusion}
As illustrated in Fig.\ref{fig:xet-v2x-architecture}~(c) and (e),
the proposed multiview multimodal fusion approach is implemented through two cascaded spatial cross-attention layers.
The first layer performs multiview fusion within the image modality,
aligning and aggregating visual semantics from different viewpoints to construct an image-enhanced BEV representation.
The second layer subsequently fuses point cloud features under the guidance of the updated BEV queries,
thereby enabling explicit cross-modal interaction and enforcing spatial consistency between image and LiDAR modalities.
This progressive fusion process can be expressed as follows:
\begin{equation}
Q^{(1)} = \text{V2XCrossAttn}(Q^{bev}, F^{ego}_{img}, F^{other}_{img}),
\end{equation}
\begin{equation}
Q^{(2)} = \text{V2XCrossAttn}(Q^{(1)}, F^{ego}_{lidar}, F^{other}_{lidar}),
\end{equation}
where $Q^{(2)}$ serves as the final multiview multimodal BEV representation.

The experimental results show that the image-first, point-cloud-second fusion order yields slightly improved performance across all the datasets,
while the reverse order remains competitive.
This finding indicates that the proposed framework is robust to fusion order variations.
Accordingly, the image-first fusion strategy is adopted as the default configuration in this study.

Overall, the proposed dual-layer deformable cross-attention module enables deep integration of multiview and multimodal information
within a unified and computationally efficient framework,
providing robust BEV features for end-to-end 3-D spatiotemporal perception in V2X cooperative environments.

\subsection{End-to-End 3-D Detection and Tracking Decoder}\label{subsec:e2e_3d_det_and_trk_decoder}
As illustrated in Fig.~\ref{fig:xet-v2x-architecture}(d), the final stage of XET-V2X involves joint 3-D object detection and tracking based on an end-to-end Transformer decoder derived from the MOTR framework~\cite{zeng2022motr}.
Unlike conventional tracking-by-detection pipelines, where detection and tracking are optimized separately with additional postprocessing,
the proposed design integrates both tasks into a unified decoding process.

The decoder operates based on the fused multiview multimodal BEV representation produced by the preceding cooperative fusion module.
Two types of queries are maintained in the decoding process: detection queries and tracking queries. 
Detection queries are responsible for discovering newly appearing objects in the current frame by interacting with the updated BEV features. 
In contrast, tracking queries propagate object states from previous frames and preserve identity consistency over time. 
By jointly updating these two query sets, the decoder can simultaneously handle object discovery and temporal associations.

During the inference process, tracking queries inherit historical object embeddings and spatial information,
enabling the model to maintain stable trajectories across frames. 
Moreover, detection queries are used to complement the tracking process by capturing objects that newly enter the scene or were previously unobserved because of occlusions or limited viewpoints.
This collaborative query mechanism allows the framework to naturally model long-term temporal dependencies in cooperative perception scenarios.

The entire detection and tracking process is optimized end to end using a unified training objective defined on the matched detection and tracking queries. 
In the current implementation, the effective nonzero supervision terms consist of object classification and 3-D bounding-box regression, 
while temporal identity consistency is learned implicitly through clip-level matching, track-query propagation, and unified sequence modeling across frames. 
Such a design enables the decoder to leverage both spatial information from cooperative perception and temporal cues from sequential observations, 
thereby improving robustness in complex V2X environments with dynamic traffic participants and partial observations.

During training, all trainable components other than the frozen image backbone are optimized jointly, including the image FPN, the LiDAR perception branch, 
the temporal self-attention module, the dual-layer multiview cross-attention fusion modules, and the final Transformer decoder.

\section{Experiments}
\subsection{Experimental Settings}
\subsubsection{Dataset Configuration}\label{subsubsec:xet-v2x datasets config}

Autonomous driving is a data-driven technology-based process that requires large-scale and diverse datasets for model training and iterative refinement.
However, the distribution of real-world driving data is inherently long-tailed: common scenarios appear frequently, 
whereas safety-critical corner cases occur with extremely low probability. 
This challenge is further amplified when high-level driving functions are deployed only in geographically restricted areas, 
limiting exposure to rare yet crucial situations. Although the current systems can perform reliably in relatively structured environments (e.g., along highways),
their performance in complex urban scenarios remains insufficient, primarily because of the lack of long-tail data.
Achieving a safety level surpassing that of human drivers is estimated to require tens of billions of testing kilometers~\cite{feng2023dense},
making real-world data collection prohibitively expensive and time-consuming.

To comprehensively evaluate the proposed model under diverse conditions and cooperative paradigms, 
we employ a dataset suite covering both real-world and simulated V2X scenarios, including V2I and V2V collaboration:
\begin{itemize}
    \item \textbf{Real-world evaluation}:  
    V2X-Seq-SPD~\cite{yu2023v2x} is used to assess model performance under realistic traffic conditions, 
    where inherent sensor noise and annotation inconsistencies provide stringent conditions for assessing practical robustness.
    \item \textbf{Simulation-based evaluation}:  
    V2X-Sim-V2V and V2X-Sim-V2I, constructed from V2X-Sim~\cite{li2022v2x}, yield noise-free ground truth labels,
    enabling controlled analysis of the model architecture, multiview collaboration, and upper-bound performance without real-world artifacts.
    \item \textbf{Complementary collaboration paradigms}:  
    V2I datasets (V2X-Seq-SPD, V2X-Sim-V2I) are established via distributed and infrastructure-assisted perception strategies,
    whereas V2V datasets (V2X-Sim-V2V) reflect fully distributed multiagent collaboration.
    Together, they form a holistic benchmark for assessing V2X collaborative perception schemes.
\end{itemize}
All the datasets are unified into the nuScenes format to standardize data loading and the annotation structure.
A consolidated dataset comparison is shown in Table~\ref{label:datasets comp}.

\begin{table}[t]
    \centering
    \small
    \caption{\textbf{Comparison of Datasets.}
    The table summarizes key characteristics and statistics of the three datasets, including collaboration type and data scale. 
    Note that all compared datasets share the same sensing modalities (Point Cloud, Image) captured by identical sensors (LiDAR, Camera) on both ego and other vehicles.}
    \label{label:datasets comp}
    \setlength{\tabcolsep}{2pt}
    \begin{tabular}{L{0.15\textwidth}|P{0.105\textwidth}P{0.11\textwidth}P{0.105\textwidth}}
        \toprule
        Attribute & V2X-Seq-SPD & V2X-Sim-V2V & V2X-Sim-V2I \\
        \midrule
        Collaboration Type & V2I & V2V & V2I \\
        Sampling Rate & 10Hz & 5Hz & 5Hz \\
        Total Scenes & 67 & 36 & 65 \\
        Training Scenes & 46 & 24 & 46 \\
        Test Scenes & 21 & 12 & 19 \\
        Total Frames & 10761 & 13600 & 6500 \\
        Training Frames & 7445 & 2400 & 4600 \\
        Test Frames & 3316 & 1200 & 1900 \\
        Ego Samples & 149672 & 33671 & 64091 \\
        Other-agent Samples & 199067 & 33389 & 110082 \\ 
        Cooperative Samples & 281078 & 51172 & 108367 \\
        \bottomrule
    \end{tabular}
\end{table}

\subsubsection{Evaluation Metrics}
To evaluate system performance in cooperative 3-D perception, 
we employ mAP to assess the 3-D object detection accuracy,
AMOTA to measure overall multiobject tracking effectiveness across recall levels,
and AMOTP to quantify tracking localization precision.

\subsubsection{Implementation Details of XET-V2X}\label{subsubsec:xet-v2x implementation details}
The proposed XET-V2X framework is implemented under the Transformer architecture~\cite{vaswani2017attention}. 
The model integrates multiview features from both ego and cooperative agents
using synchronized LiDAR and image sequences for spatiotemporal 3-D perception.

The point clouds are voxelized using a pillar-based representation with a voxel size of [0.2, 0.2, 8].
For the V2X-Seq-SPD dataset~\cite{yu2023v2x}, the LiDAR range for CAVs is set to [-51.2, -51.2, -5.0, 51.2, 51.2, 3.0], 
while for RSUs, it is set to [0, -51.2, -5.0, 102.4, 51.2, 3.0]. For the V2X-Sim-V2V and V2X-Sim-V2I datasets,
the LiDAR range for both CAVs and RSUs is set to [-51.2, -51.2, -3.0, 51.2, 51.2, 5.0].

PillarFeatureNet encodes per-pillar features (4-dimensional point attributes),
which is followed by a SECOND backbone with a layer configuration [3,5,5] and output channels [64, 128, 256].
The multiscale LiDAR features are aggregated by an FPN into 256-dimensional representations.

The images are processed using a ResNet-101 backbone~\cite{he2016deep} with multiscale feature extraction.
The resulting feature maps are fused via an FPN into 256-dimensional outputs.
Standard image normalization and GridMask augmentation are applied during training.

The temporal module consists of six encoder layers and six decoder layers.
Each encoder layer includes temporal self-attention to model historical dependencies, 
followed by two-stage spatial cross-attention to fuse multimodal spatial features.
The decoder adopts a dual-attention structure and supports 900 learnable object queries.
The BEV feature map is configured to a resolution of $200\times200$ with 256 channels.

A unified multitask head performs 3-D object detection and online tracking, and a HungarianAssigner3DTrack-based matcher is applied for bipartite matching within the ClipMatcher framework. 
The final training objective is defined as: 
\begin{equation}
    L_{\mathrm{total}}=\lambda_{\mathrm{cls}}L_{\mathrm{cls}}+\lambda_{\mathrm{bbox}}L_{\mathrm{bbox}},
\end{equation}
where $L_{\mathrm{cls}}$ is the Focal Loss used for classification and $L_{\mathrm{bbox}}$ is the L1 loss used for 3-D bounding-box regression. 
Following the final training configuration, we set $\lambda_{\mathrm{cls}}=2.0$ and $\lambda_{\mathrm{bbox}}=0.25$, with Focal Loss parameters $\gamma=2.0$ and $\alpha=0.25$. 
Tracking is trained jointly through sequence-level matching and track-query propagation rather than through an additional standalone identity loss term. 
The tracker incorporates both semantic consistency and geometric association.

Each input sequence contains five frames, with the first four serving as historical context.
The AdamW optimizer with a learning rate of $2\times10^{-4}$ and a weight decay of 0.01 is used for training.
A cosine annealing schedule with 500 warm-up iterations, 10 epochs, and gradient clipping (norm 35) is applied.
During inference, the model outputs both detection and tracking predictions in a single forward pass.

\begin{table}[t]
    \centering
    \small
    \caption{\textbf{Comparison of Different Methods.} 
    This table summarizes the configurations of the proposed XET-V2X and baseline models across four dimensions: End-to-End (E2E) formulation for perception, Multi-View collaboration, Multi-Modal fusion, and specific configurations of Viewpoints (Ego vs. Others) and Sensory Modalities (LiDAR vs. Image). All models are trained for 10 epochs.}
    \label{tab:xet-v2x baseline models comparison}
    \setlength{\tabcolsep}{4pt}
    \begin{tabular}{L{0.16\textwidth}P{0.02\textwidth}P{0.03\textwidth}P{0.03\textwidth}P{0.022\textwidth}P{0.03\textwidth}P{0.033\textwidth}P{0.03\textwidth}}
        \toprule
        \multirow{2}{*}{Model} & \multirow{2}{*}{E2E} & Multi & Multi & \multicolumn{2}{c}{Viewpoints} & \multicolumn{2}{c}{Modalities} \\
        \cmidrule(lr){5-6} \cmidrule(lr){7-8}
        & & View & Modal & Ego & Others & LiDAR & Image \\
        \midrule
        ImvoxelNet-V~\cite{rukhovich2022imvoxelnet} & & & & \checkmark & & & \checkmark \\
        PointPillars-V~\cite{lang2019pointpillars} & & & & \checkmark & & \checkmark & \\
        MVXNet-V~\cite{sindagi2019mvx} & & & \checkmark & \checkmark & & \checkmark & \checkmark \\
        ImvoxelNet-V2X~\cite{rukhovich2022imvoxelnet} & & \checkmark & & \checkmark & \checkmark & & \checkmark \\
        PointPillars-V2X~\cite{lang2019pointpillars} & & \checkmark & & \checkmark & \checkmark & \checkmark & \\
        MVXNet-V2X~\cite{sindagi2019mvx} & & \checkmark & \checkmark & \checkmark & \checkmark & \checkmark & \checkmark \\
        CET-V & \checkmark & & & \checkmark & & & \checkmark \\
        LET-V~\cite{yang2025letvic} & \checkmark & & & \checkmark & & \checkmark & \\
        XET-V & \checkmark & & \checkmark & \checkmark & & \checkmark & \checkmark \\
        CET-V2X & \checkmark & \checkmark & & \checkmark & \checkmark & & \checkmark \\
        LET-V2X~\cite{yang2025letvic} & \checkmark & \checkmark & & \checkmark & \checkmark & \checkmark & \\
        \textbf{XET-V2X (Ours)} & \checkmark & \checkmark & \checkmark & \checkmark & \checkmark & \checkmark & \checkmark \\
        \bottomrule
    \end{tabular}
\end{table}

\begin{table*}[htpb!]
    \centering
    \small
    \caption{\textbf{Performance comparison under different communication delay conditions on three Vehicle-to-Everything (V2X) benchmarks.} 
    All compared models are trained for 10 epochs. The baseline models, including ImvoxelNet~\cite{rukhovich2022imvoxelnet}, PointPillars~\cite{lang2019pointpillars}, and MVXNet~\cite{sindagi2019mvx}, 
    follow a tracking-by-detection paradigm using AB3DMOT~\cite{weng2020ab3dmot}. 
    Latency is measured in terms of frame-level delay, where one frame corresponds to 100~ms for V2X-Seq-SPD~\cite{yu2023v2x} (10~Hz), and 200~ms for both V2X-Sim-V2V and V2X-Sim-V2I (5~Hz).}
    \label{tab:xet-v2x experiment results}
    \begin{tabular}{L{0.16\textwidth}P{0.05\textwidth}P{0.05\textwidth}P{0.06\textwidth}P{0.06\textwidth}P{0.05\textwidth}P{0.06\textwidth}P{0.06\textwidth}P{0.05\textwidth}P{0.06\textwidth}P{0.06\textwidth}}
        \toprule
        \multirow{2}{*}{Model} & Latency & \multicolumn{3}{c}{V2X-Seq-SPD} & \multicolumn{3}{c}{V2X-Sim-V2V} & \multicolumn{3}{c}{V2X-Sim-V2I} \\
        \cmidrule(lr){3-5} \cmidrule(lr){6-8} \cmidrule(lr){9-11}
        & (frames) & mAP$\uparrow$ & AMOTA$\uparrow$ & AMOTP$\downarrow$ & mAP$\uparrow$ & AMOTA$\uparrow$ & AMOTP$\downarrow$ & mAP$\uparrow$ & AMOTA$\uparrow$ & AMOTP$\downarrow$ \\
        \midrule
        ImvoxelNet-V~\cite{rukhovich2022imvoxelnet} & - & 0.042 & 0.066 & 1.902 & - & - & - & - & - & - \\
        PointPillars-V~\cite{lang2019pointpillars} & - & 0.370 & 0.413 & 1.179 & - & - & - & - & - & - \\
        MVXNet-V~\cite{sindagi2019mvx} & - & 0.408 & 0.442 & 1.125 & - & - & - & - & - & - \\
        ImvoxelNet-V2X~\cite{rukhovich2022imvoxelnet} & 0 & 0.211 & 0.284 & 1.548 & - & - & - & - & - & - \\
        ImvoxelNet-V2X~\cite{rukhovich2022imvoxelnet} & 1 & 0.190 & 0.279 & 1.560 & - & - & - & - & - & - \\
        ImvoxelNet-V2X~\cite{rukhovich2022imvoxelnet} & 2 & 0.172 & 0.245 & 1.599 & - & - & - & - & - & - \\
        PointPillars-V2X~\cite{lang2019pointpillars} & 0 & 0.548 & 0.579 & 0.774 & - & - & - & - & - & - \\
        PointPillars-V2X~\cite{lang2019pointpillars} & 1 & 0.510 & 0.576 & 0.789 & - & - & - & - & - & - \\
        PointPillars-V2X~\cite{lang2019pointpillars} & 2 & 0.486 & 0.528 & 0.876 & - & - & - & - & - & - \\
        MVXNet-V2X~\cite{sindagi2019mvx} & 0 & 0.616 & 0.648 & 0.670 & - & - & - & - & - & - \\
        MVXNet-V2X~\cite{sindagi2019mvx} & 1 & 0.571 & 0.626 & 0.734 & - & - & - & - & - & - \\
        MVXNet-V2X~\cite{sindagi2019mvx} & 2 & 0.543 & 0.591 & 0.781 & - & - & - & - & - & - \\
        CET-V & - & 0.198 & 0.241 & 1.618 & 0.217 & 0.190 & 1.646 & 0.177 & 0.136 & 1.726 \\
        LET-V~\cite{yang2025letvic} & - & 0.449 & 0.437 & 1.175 & 0.424 & 0.315 & 1.401 & 0.379 & 0.308 & 1.432 \\
        XET-V & - & 0.490 & 0.469 & 1.122 & 0.510 & 0.459 & 1.168 & 0.412 & 0.364 & 1.335 \\
        CET-V2X & 0 & 0.179 & 0.169 & 1.630 & 0.313 & 0.291 & 1.419 & 0.714 & 0.782 & 0.567 \\
        CET-V2X & 1 & 0.188 & 0.169 & 1.631 & 0.299 & 0.272 & 1.465 & 0.638 & 0.723 & 0.706 \\
        CET-V2X & 2 & 0.187 & 0.169 & 1.633 & 0.278 & 0.240 & 1.499 & 0.572 & 0.626 & 0.826 \\
        LET-V2X~\cite{yang2025letvic} & 0 & 0.668 & 0.616 & 0.859 & 0.502 & 0.391 & 0.865 & 0.653 & 0.562 & 0.671 \\
        LET-V2X~\cite{yang2025letvic} & 1 & 0.608 & 0.592 & 0.951 & 0.462 & 0.373 & 0.898 & 0.573 & 0.511 & 0.817 \\
        LET-V2X~\cite{yang2025letvic} & 2 & 0.546 & 0.543 & 1.009 & 0.451 & 0.346 & 0.938 & 0.508 & 0.430 & 0.901 \\
        \textbf{XET-V2X (Ours)} & 0 & \textbf{0.795} & \textbf{0.787} & \textbf{0.591} & \textbf{0.766} & \textbf{0.731} & \textbf{0.677} & \textbf{0.858} & \textbf{0.819} & \textbf{0.476} \\
        \textbf{XET-V2X (Ours)} & 1 & \textbf{0.743} & \textbf{0.761} & \textbf{0.668} & \textbf{0.714} & \textbf{0.698} & \textbf{0.760} & \textbf{0.746} & \textbf{0.776} & \textbf{0.584} \\
        \textbf{XET-V2X (Ours)} & 2 & \textbf{0.688} & \textbf{0.722} & \textbf{0.730} & \textbf{0.668} & \textbf{0.652} & \textbf{0.804} & \textbf{0.664} & \textbf{0.668} & \textbf{0.712} \\
        \bottomrule
    \end{tabular}
\end{table*}

\subsubsection{Baseline Model Configuration}\label{sec:xet-v2x baseline settings}
To comprehensively evaluate the improvements achieved with XET-V2X in end-to-end 3-D spatiotemporal perception, the following baselines are adopted.
All the baselines (CET-V/LET-V~\cite{yang2025letvic}/XET-V and their V2X counterparts) are controlled variants derived from the same end-to-end architecture with identical training schedules and backbone settings; they differ only in the enabled viewpoints/modalities and corresponding fusion modules, and we select hyperparameters on the basis of validation performance to ensure fair comparison.
\begin{itemize}
  \item \textbf{ImvoxelNet-V}~\cite{rukhovich2022imvoxelnet}: Employs only the ego-view monocular image stream for 3-D object detection
  and uses AB3DMOT~\cite{weng2020ab3dmot} as the tracking module to enable temporal perception; represents a single-view, single-modality (image) tracking-by-detection framework.
  \item \textbf{PointPillars-V}~\cite{lang2019pointpillars}: Uses only the ego-view LiDAR stream for 3-D object detection, 
  with AB3DMOT~\cite{weng2020ab3dmot} applied for object tracking to capture temporal dynamics; represents a single-view, single-modality (LiDAR) tracking-by-detection framework.
  \item \textbf{MVXNet-V}~\cite{sindagi2019mvx}: Uses both ego-view images and ego-view LiDAR as inputs for multimodal 3-D object detection, 
  followed by AB3DMOT~\cite{weng2020ab3dmot} for tracking; represents a single-view, multimodal (image + LiDAR) tracking-by-detection framework.
  \item \textbf{ImvoxelNet-V2X}~\cite{rukhovich2022imvoxelnet}: Incorporates ego-view images and other-agent images and performs cooperative perception via late fusion at the detection stage;
  AB3DMOT~\cite{weng2020ab3dmot} is then used for tracking to model temporal continuity; represents a multiview cooperative, single-modality (image) tracking-by-detection framework with late fusion.
  \item \textbf{PointPillars-V2X}~\cite{lang2019pointpillars}: Incorporates ego-view LiDAR data and other-agent LiDAR data, implements cooperative perception via late fusion for detection,
  and applies AB3DMOT~\cite{weng2020ab3dmot} for tracking; represents a multiview cooperative, single-modality (LiDAR) tracking-by-detection framework with late fusion.
  \item \textbf{MVXNet-V2X}~\cite{sindagi2019mvx}: Incorporates both images and LiDAR data from the ego vehicle and other agents, implements multimodal cooperative detection via late fusion,
  and employs AB3DMOT~\cite{weng2020ab3dmot} for tracking to achieve temporal perception; represents a multiview cooperative, multimodal (image + LiDAR) tracking-by-detection framework with late fusion.
  \item \textbf{CET-V}: Employs only the ego-view monocular image stream and outputs detection and tracking results in an end-to-end manner;
  represents a single-view, single-modality (image) end-to-end framework.
  \item \textbf{LET-V}~\cite{yang2025letvic}: Uses only the ego-view narrow-FoV LiDAR with end-to-end detection and tracking;
  represents a single-view, single-modality (LiDAR) end-to-end perception scheme.
  \item \textbf{XET-V}: Uses both ego-view images and ego-view LiDAR data as inputs, without cooperative viewpoints; represents a single-view, multimodal (image + LiDAR) end-to-end perception scheme.
  \item \textbf{CET-V2X}: Incorporates ego-view images and multiview camera images from cooperative agents, without using LiDAR;
  represents a multiview cooperative, single-modality (image) end-to-end perception scheme.
  \item \textbf{LET-V2X}~\cite{yang2025letvic}: Incorporates ego-view LiDAR and multiview LiDAR data from cooperative agents, without using images;
represents a multiview cooperative, single-modality (LiDAR) end-to-end perception scheme.
\end{itemize}
These baselines are used for the systematic assessment of XET-V2X in terms of perception accuracy, robustness, multiview collaboration,
and cross-agent feature fusion, as summarized in Table~\ref{tab:xet-v2x baseline models comparison}.

\subsection{Quantitative Evaluation}\label{subsec:xet-v2x experiment results}
\textbf{Latency simulation.} For a frame-level delay $d\in\{0,1,2\}$, we simulate communication latency by time-shifting the other-agent features by $d$ frames (i.e., at the ego time index $k$, we use the most recent received other-agent observation at index $k-d$), while the ego observation remains at index $k$. This ensures that no future information is used and directly reflects asynchronous feature arrival under V2X transmission.

\subsubsection{Overall Performance}
\begin{figure*}[htbp!]
    \centering
    \begin{subfigure}[b]{0.33\textwidth}
        \centering
        \includegraphics[width=\linewidth]{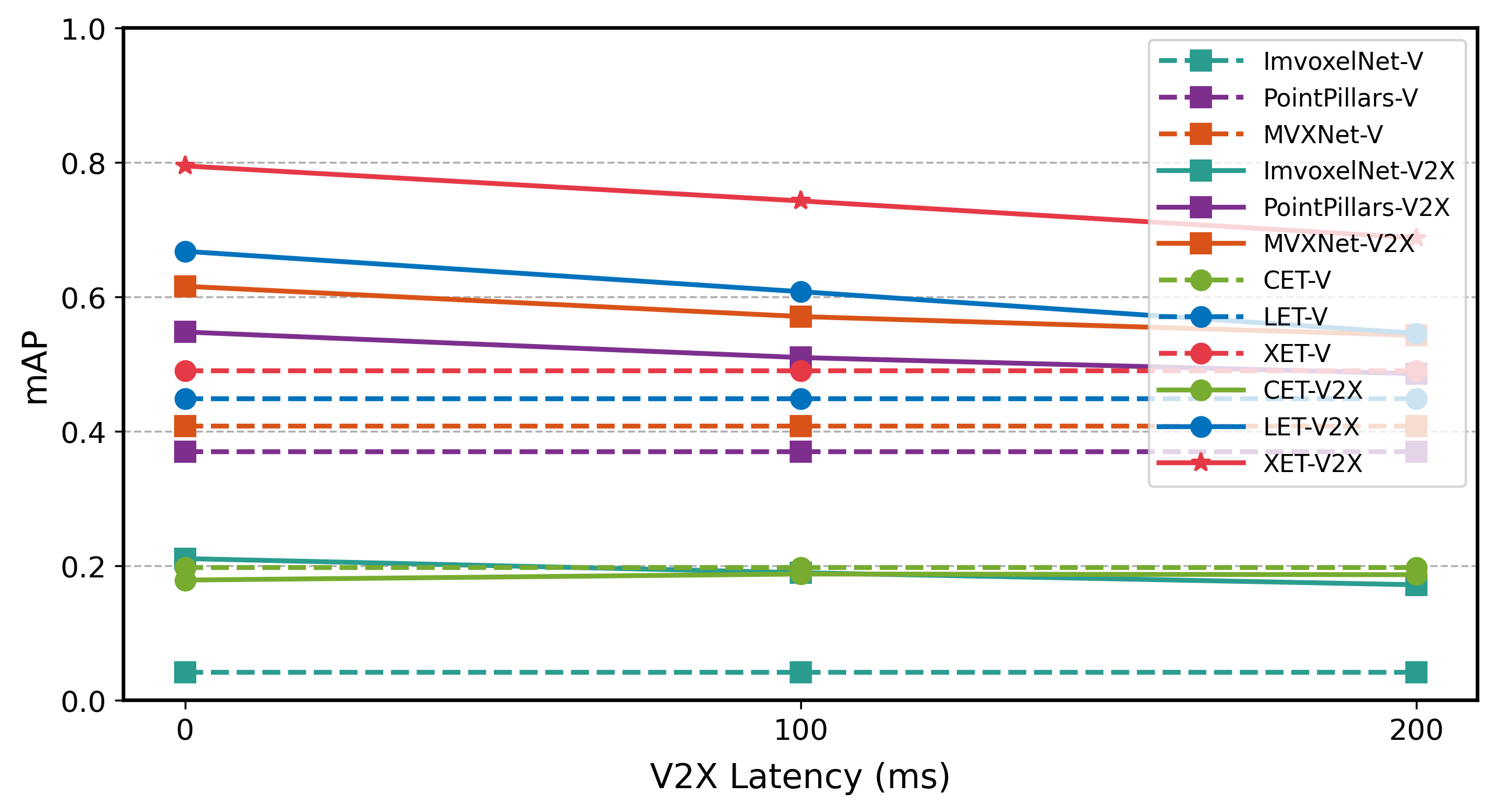}
        \subcaption{V2X-Seq-SPD: mAP}
        \label{fig:combined_a}
    \end{subfigure}%
    \begin{subfigure}[b]{0.33\textwidth}
        \centering
        \includegraphics[width=\linewidth]{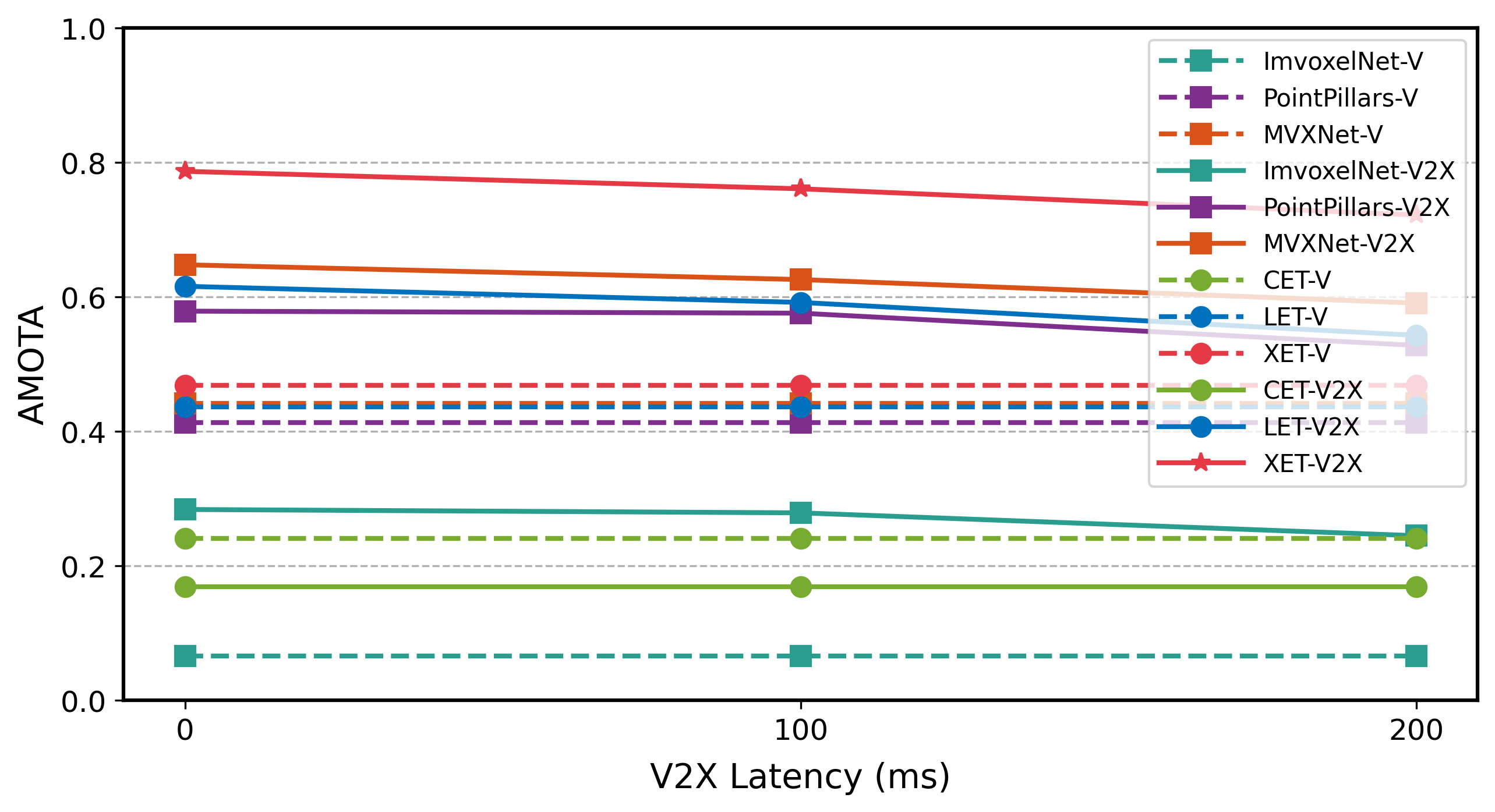}
        \subcaption{V2X-Seq-SPD: AMOTA}
        \label{fig:combined_b}
    \end{subfigure}%
    \begin{subfigure}[b]{0.33\textwidth}
        \centering
        \includegraphics[width=\linewidth]{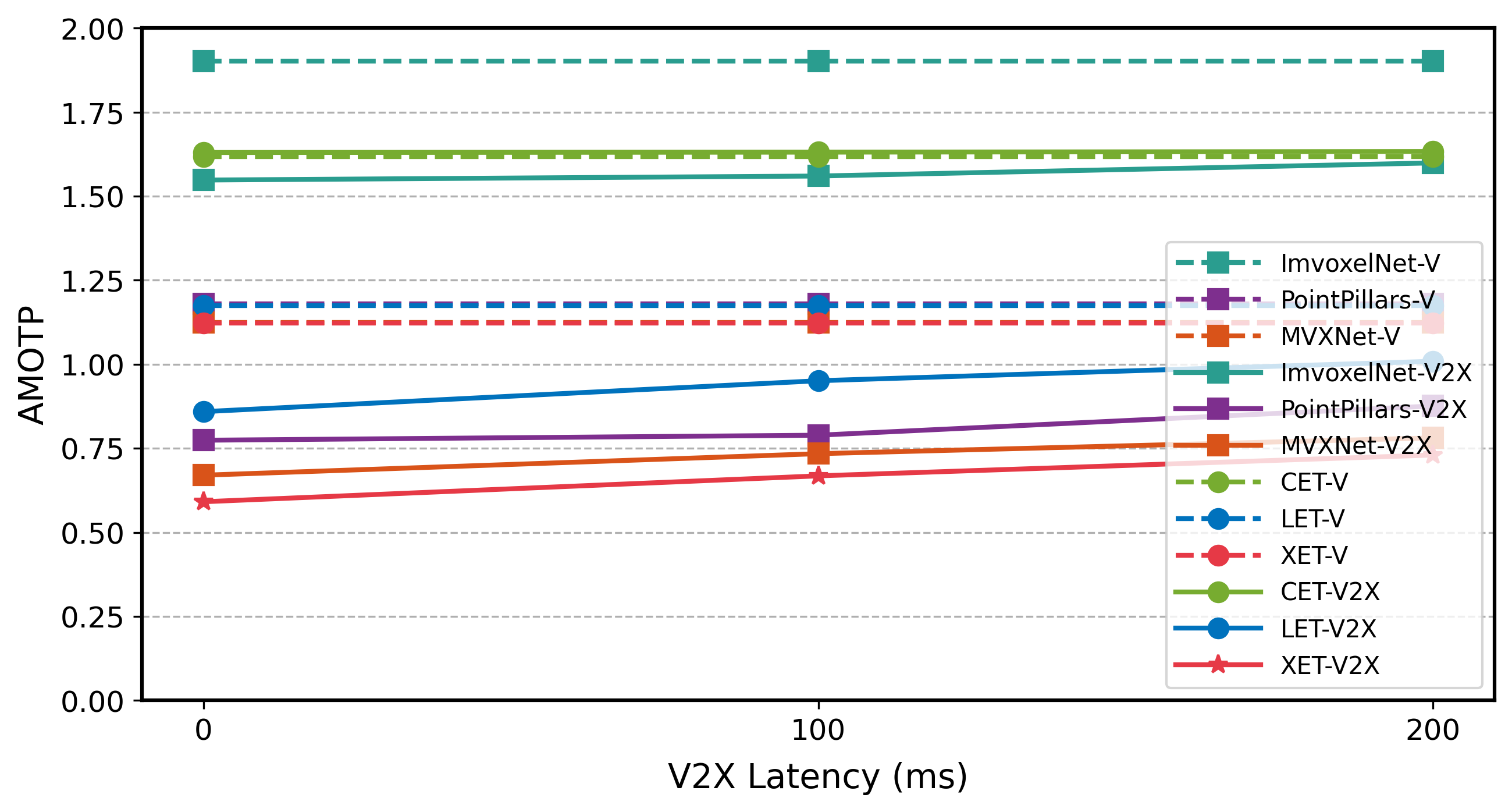}
        \subcaption{V2X-Seq-SPD: AMOTP}
        \label{fig:combined_c}
    \end{subfigure}
    \\
    \begin{subfigure}[b]{0.33\textwidth}
        \centering
        \includegraphics[width=\linewidth]{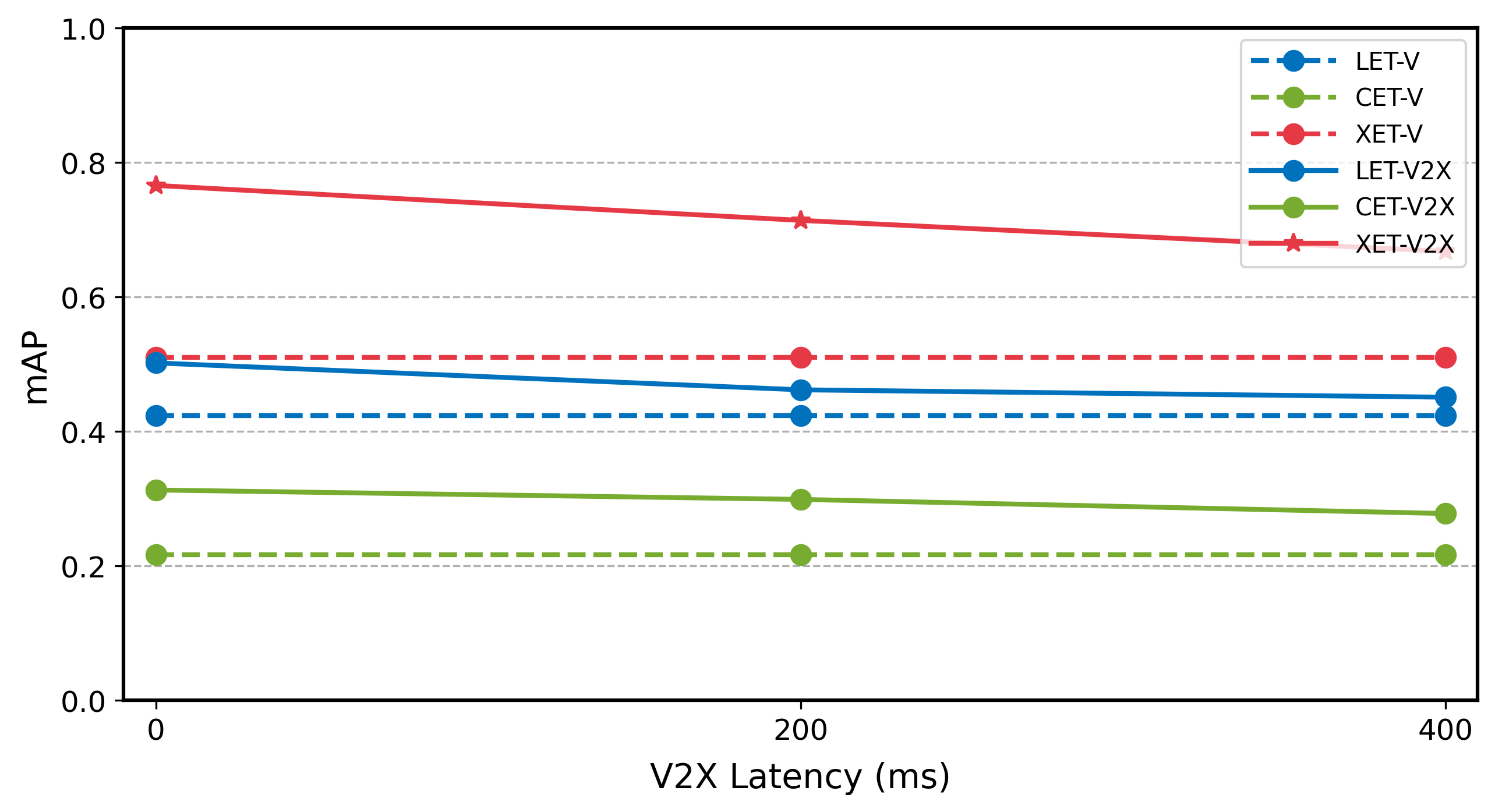}
        \subcaption{V2X-Sim-V2V: mAP}
        \label{fig:combined_d}
    \end{subfigure}%
    \begin{subfigure}[b]{0.33\textwidth}
        \centering
        \includegraphics[width=\linewidth]{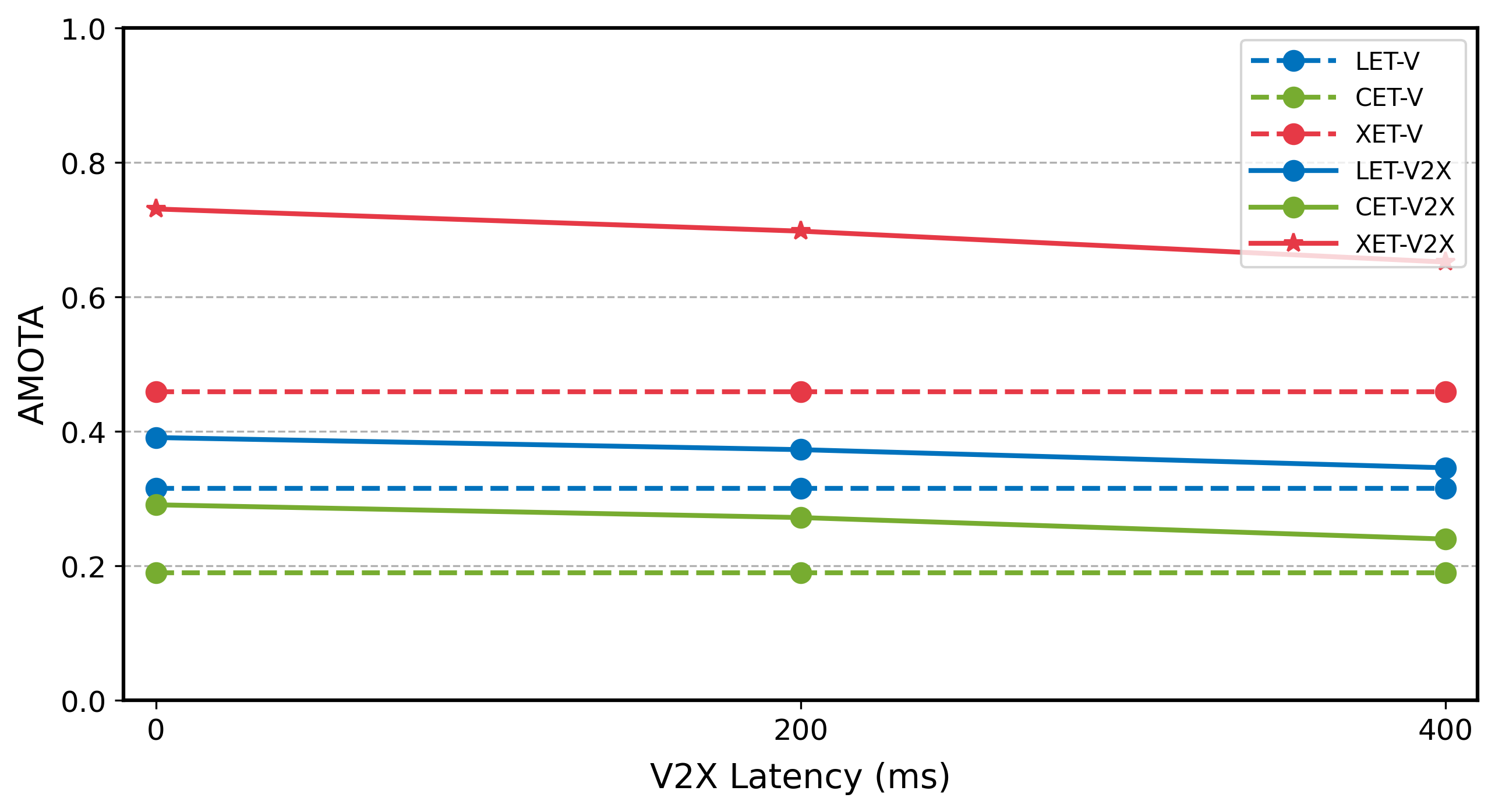}
        \subcaption{V2X-Sim-V2V: AMOTA}
        \label{fig:combined_e}
    \end{subfigure}%
    \begin{subfigure}[b]{0.33\textwidth}
        \centering
        \includegraphics[width=\linewidth]{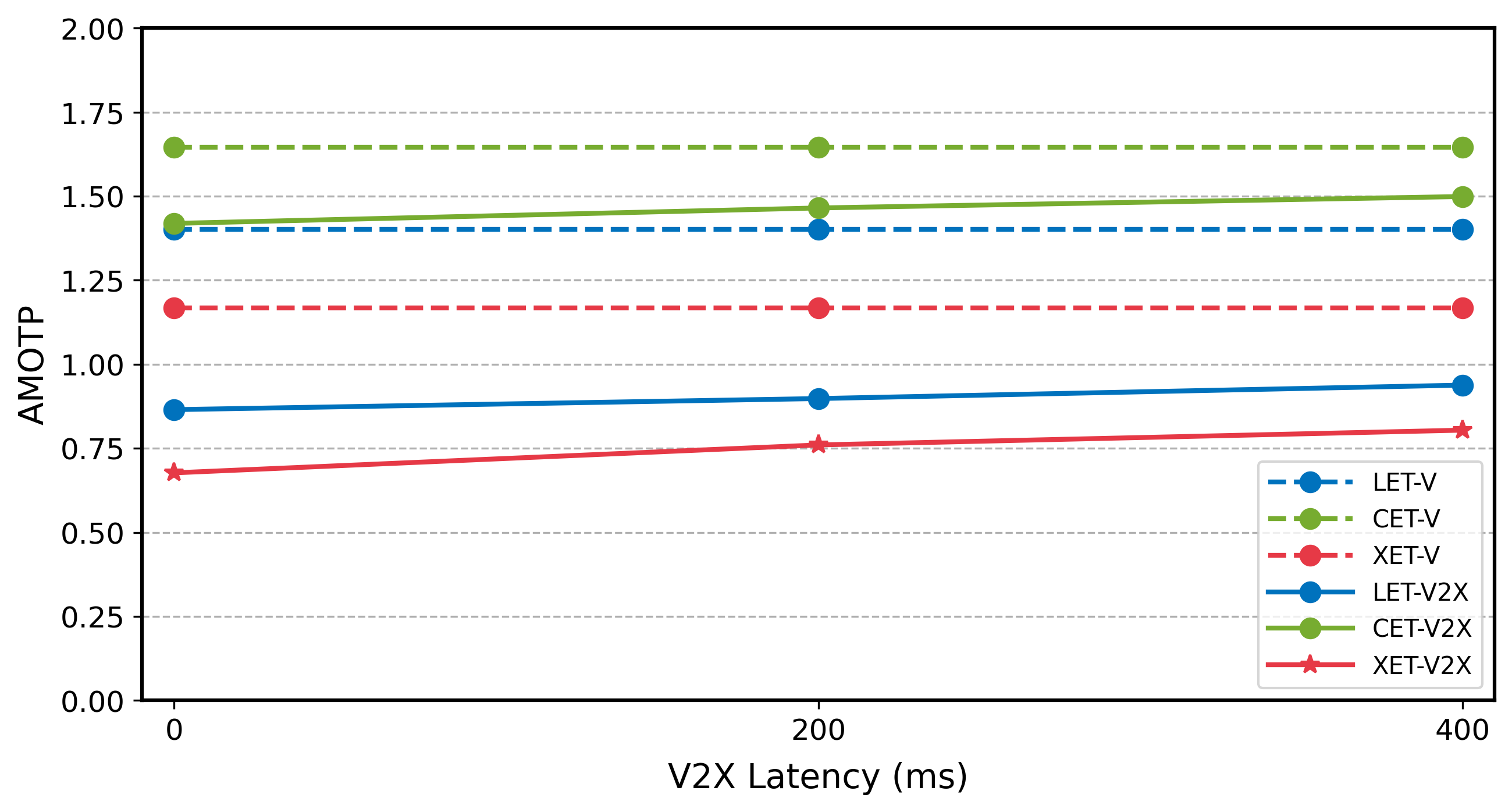}
        \subcaption{V2X-Sim-V2V: AMOTP}
        \label{fig:combined_f}
    \end{subfigure}
    \\
    \begin{subfigure}[b]{0.33\textwidth}
        \centering
        \includegraphics[width=\linewidth]{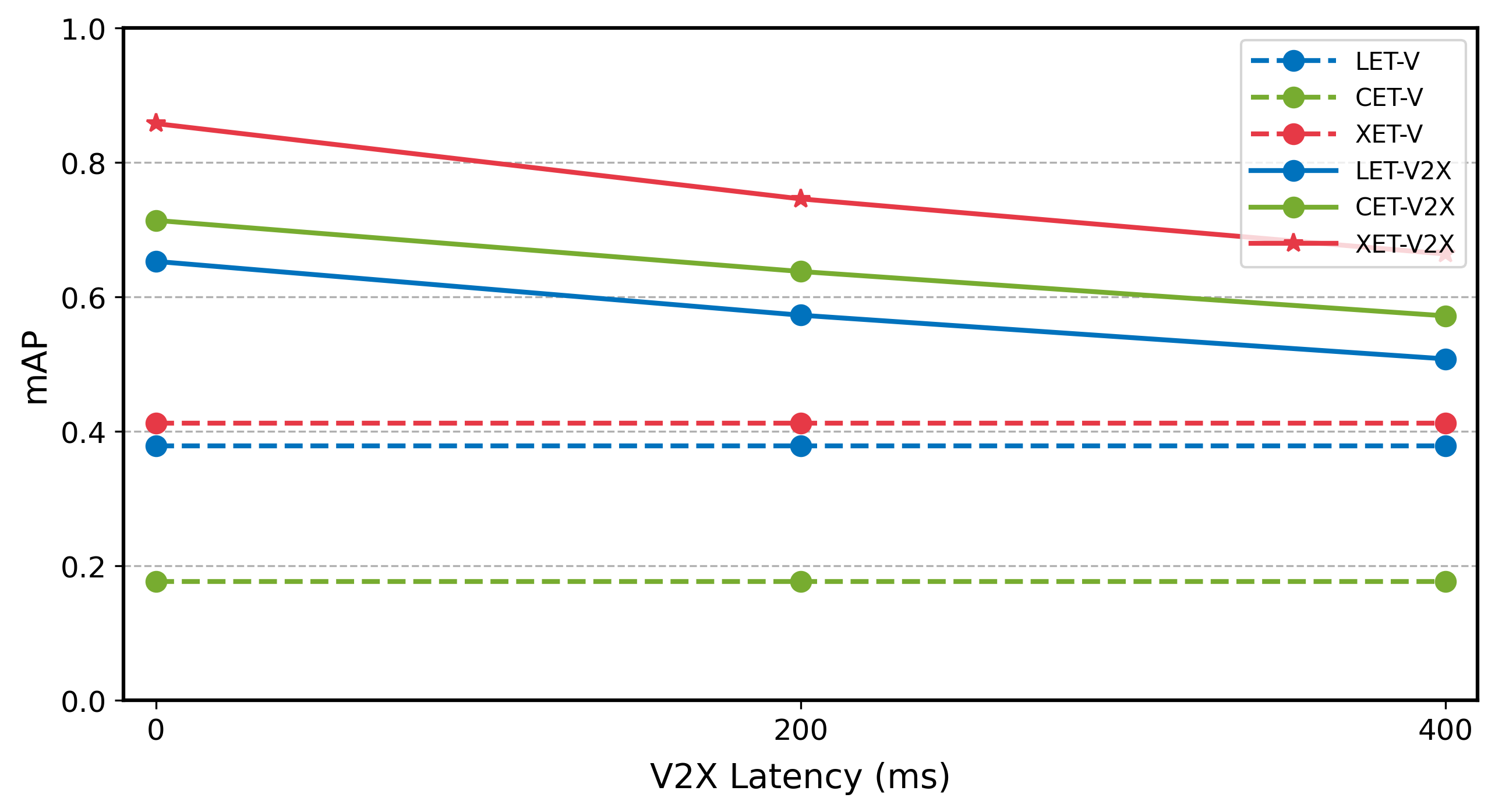}
        \subcaption{V2X-Sim-V2I: mAP}
        \label{fig:combined_g}
    \end{subfigure}%
    \begin{subfigure}[b]{0.33\textwidth}
        \centering
        \includegraphics[width=\linewidth]{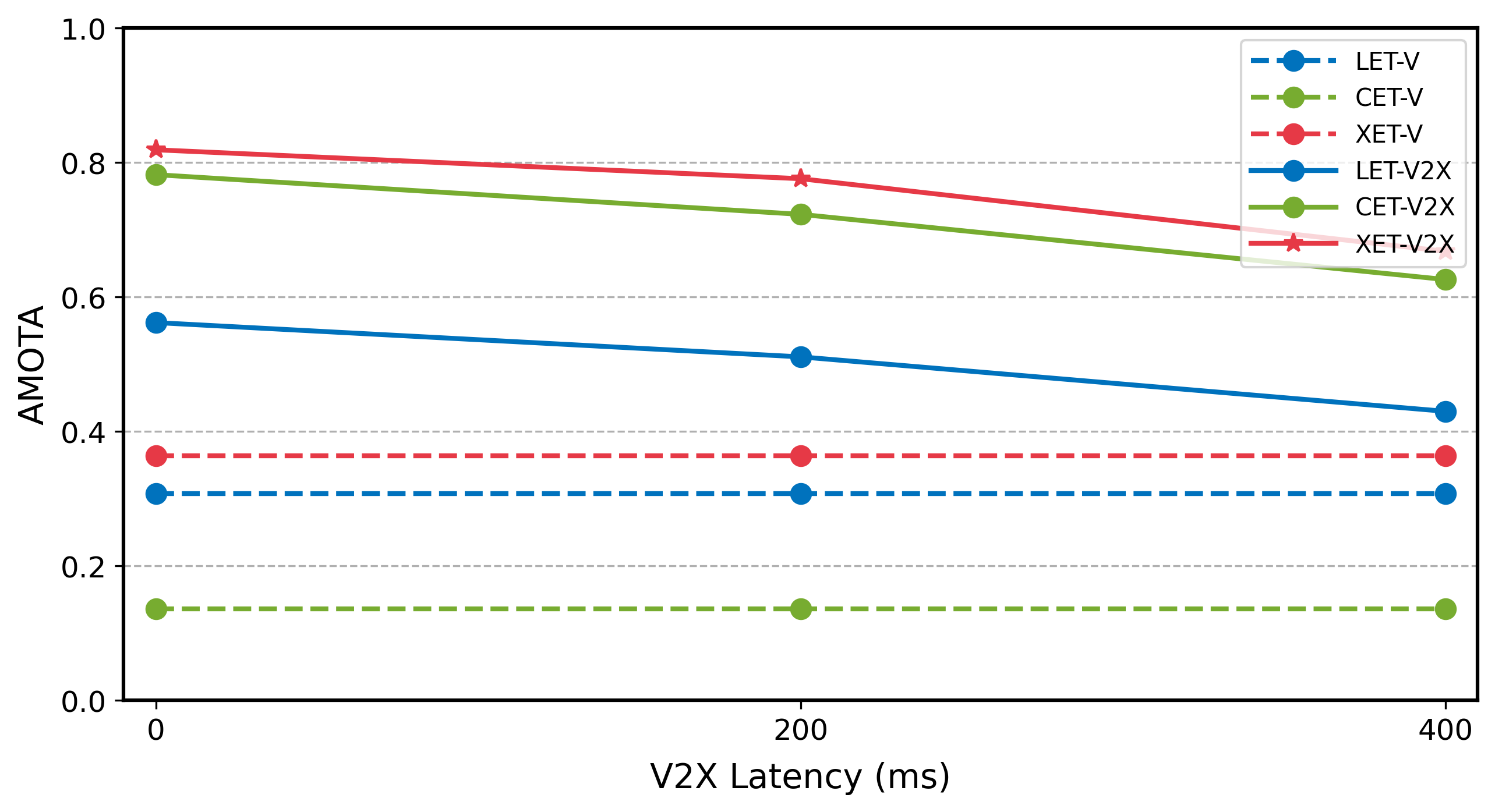}
        \subcaption{V2X-Sim-V2I: AMOTA}
        \label{fig:combined_h}
    \end{subfigure}%
    \begin{subfigure}[b]{0.33\textwidth}
        \centering
        \includegraphics[width=\linewidth]{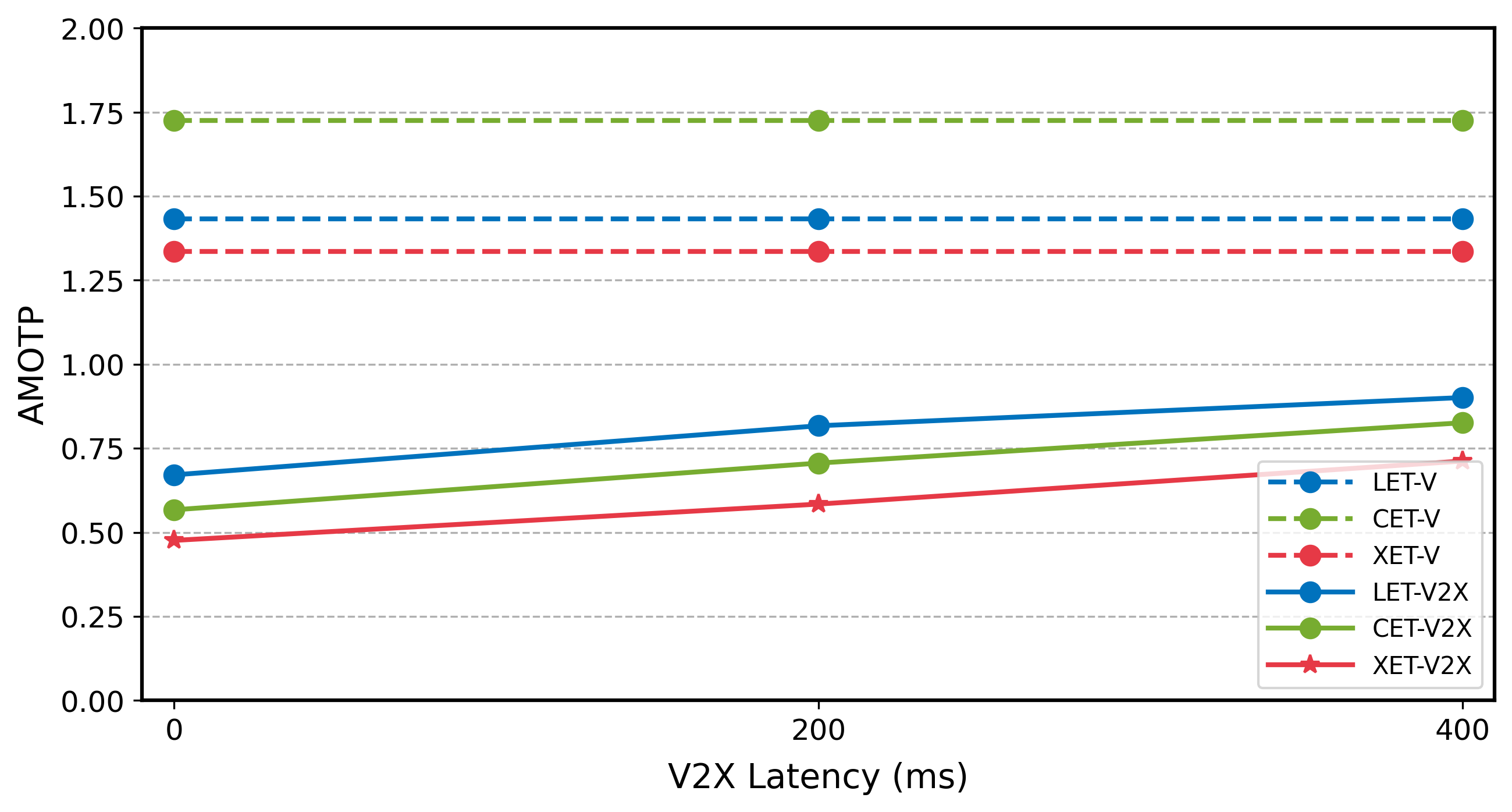}
        \subcaption{V2X-Sim-V2I: AMOTP}
        \label{fig:combined_i}
    \end{subfigure}
    \caption{\textbf{Comprehensive performance comparison of XET-V2X and baseline models under varying communication latency conditions across three datasets.} 
    The top row (a-c) includes evaluations for the V2X-Seq-SPD dataset~\cite{yu2023v2x}, the middle row (d-f) includes evaluations for the V2X-Sim-V2V dataset~\cite{li2022v2x},
    and the bottom row (g-i) includes evaluations for the V2X-Sim-V2I dataset~\cite{li2022v2x}.
    Across all subfigures, the horizontal axis represents different communication latency conditions, 
    while the vertical axis denotes the corresponding evaluation metrics. For the evaluation metrics, 
    high values indicate good performance for detection (mAP) and tracking accuracy (AMOTA),
    whereas low values indicate good performance for tracking precision (AMOTP).}
    \label{fig:combined_performance_comparison}
\end{figure*}

As reported in Table~\ref{tab:xet-v2x experiment results} and Fig.~\ref{fig:combined_performance_comparison},
the XET models outperform all the evaluated baselines for the real-world V2X-Seq-SPD dataset~\cite{yu2023v2x} and the simulated V2X-Sim benchmarks~\cite{li2022v2x}.
In addition to prior end-to-end methods (CET and LET), we further introduce several tracking-by-detection baselines, including ImVoxelNet~\cite{rukhovich2022imvoxelnet}, 
PointPillars~\cite{lang2019pointpillars}, and MVXNet~\cite{sindagi2019mvx}, which are combined with AB3DMOT~\cite{weng2020ab3dmot} to enable temporal perception.
These baselines cover single-view image-based, LiDAR-based, and multimodal perception settings, as well as their cooperative extensions under V2X communication, providing a comprehensive comparison across modalities and collaboration levels.

Across all the datasets, XET-V consistently outperforms LET-V and CET-V, whereas XET-V2X achieves the best results compared with LET-V2X~\cite{yang2025letvic} and CET-V2X in terms of mAP, AMOTA, and AMOTP.
Notably, the improvements remain stable under different communication latency conditions, demonstrating the robustness of the proposed framework in realistic V2X scenarios.
Compared with the tracking-by-detection baselines, the proposed XET-V2X displays clear advantages, highlighting the effectiveness of integrating multiview collaboration and multimodal fusion within a unified end-to-end architecture.
These results indicate that the proposed framework provides consistent performance gains across heterogeneous datasets and maintains strong generalization ability in both real-world and simulated environments.

\subsubsection{Quantitative Analysis of Multiview Collaboration}
The V2X-Seq-SPD dataset~\cite{yu2023v2x} provides a realistic vehicle-infrastructure cooperative perception benchmark,
characterized by illumination variations, frequent occlusions, and complex traffic dynamics.
As shown in Table~\ref{tab:xet-v2x experiment results} and Fig.~\ref{fig:combined_a}--\ref{fig:combined_c},
XET-V2X consistently outperforms the single-vehicle baseline XET-V under all latency settings.
In the case of zero latency, XET-V2X improves the mAP and AMOTA by 30.5\% and 31.8\%, respectively, while also reducing AMOTP compared to those in the baseline case.
These gains remain stable as the communication delay increases, demonstrating strong robustness to realistic V2X latency.

Similar trends are observed for the simulated cooperative perception benchmarks.
By introducing complementary viewpoints from multiple agents,
multiview collaboration effectively enlarges the observable region and mitigates occlusions in single-vehicle perception.
As a result, XET-V2X consistently outperforms XET-V across all delay conditions.

The results confirm that multiview collaboration significantly enhances perception performance, especially in complex traffic environments.

\subsubsection{Quantitative Analysis of Multimodal Fusion}
The results in Table~\ref{tab:xet-v2x experiment results} are used to further evaluate the effectiveness of multimodal fusion.
In the single-vehicle setting, XET-V consistently outperforms the unimodal baselines
LET-V (LiDAR-only) and CET-V (camera-only).
For the V2X-Seq-SPD dataset~\cite{yu2023v2x}, XET-V improves the mAP and AMOTA by 4.1\% and 3.2\%, respectively, compared to those of LET-V
and achieves substantially larger gains over CET-V,
indicating that the image and point cloud modalities provide complementary geometric and semantic information.

When cooperative perception is implemented, the advantages of multimodal fusion become more pronounced.
Under zero latency, XET-V2X outperforms LET-V2X~\cite{yang2025letvic} by 12.7\% for mAP and 17.1\% for AMOTA,
while also achieving clear improvements over CET-V2X.
Consistent performance gains are observed across different latency settings and simulated datasets~\cite{li2022v2x}.

These results demonstrate that multimodal fusion improves both perception accuracy and robustness in cooperative environments.

\subsubsection{End-to-End vs. Tracking-by-Detection Cooperative Perception}
We further compare the proposed framework with representative cooperative perception methods based on the tracking-by-detection paradigm,
including ImvoxelNet-V2X~\cite{rukhovich2022imvoxelnet}, PointPillars-V2X~\cite{lang2019pointpillars}, and MVXNet-V2X~\cite{sindagi2019mvx}.
These methods perform multiview detection followed by temporal association using AB3DMOT~\cite{weng2020ab3dmot}.

As presented in Table~\ref{tab:xet-v2x experiment results}, XET-V2X consistently achieves the best performance across all the datasets and latency settings.
For instance, on the basis of the V2X-Seq-SPD dataset~\cite{yu2023v2x} with zero latency, XET-V2X achieves an mAP of 0.795,
significantly outperforming MVXNet-V2X~\cite{sindagi2019mvx} (0.616), PointPillars-V2X~\cite{lang2019pointpillars} (0.548), and ImvoxelNet-V2X~\cite{rukhovich2022imvoxelnet} (0.211).
Similar improvements are observed in AMOTA and AMOTP.
Moreover, in the single-vehicle case, XET-V also achieves better performance than the corresponding tracking-by-detection baselines do,
further verifying the effectiveness of the proposed architecture.

The results indicate that decoupled detection and tracking pipelines are less effective for cooperative perception.
In contrast, the proposed framework jointly models spatial perception,
multiview fusion and temporal reasoning in an end-to-end manner,
leading to more effective cross-agent feature interactions and improved robustness.

The experiments demonstrate that end-to-end cooperative perception provides clear advantages over traditional tracking-by-detection approaches.

\subsection{Computational Efficiency Analysis}
\begin{table}[htpb!]
    \centering
    \small
    \caption{\textbf{Computational Efficiency and Macroarchitecture Ablation Analyses.} All the methods are run on the same NVIDIA RTX 4090 GPU.
    The table demonstrates the trade-off between modality/viewpoint choices and computational load.}
    \label{tab:efficiency_analysis}
    \begin{tabular}{L{0.13\textwidth}P{0.09\textwidth}P{0.09\textwidth}P{0.07\textwidth}}
        \toprule
        Model & Params (M) & FLOPs (G) & FPS \\
        \midrule
        CET-V & 33.38 & 424.92 & 5.6 \\
        LET-V & 30.10 & 306.10 & 6.2 \\
        XET-V & 35.36 & 564.60 & 4.7 \\
        CET-V2X & 33.38 & 702.95 & 4.5 \\
        LET-V2X~\cite{yang2025letvic} & 30.11 & 441.77 & 5.1 \\
        XET-V2X (Ours) & 35.36 & 990.10 & 3.5 \\
        \bottomrule
    \end{tabular}
\end{table}
To provide a comprehensive evaluation of our framework on the deployed computing platform, we conduct an efficiency analysis.
The inference speed (frames per second, FPS), computational load (FLOPs), and parameter count for all end-to-end model variants are evaluated for models run on the same NVIDIA RTX 4090 GPU,
as detailed in Table~\ref{tab:efficiency_analysis}. Note that traditional tracking-by-detection baselines are not strictly comparable in terms of end-to-end GPU latency because of their reliance on CPU-based tracking algorithms;
thus, we focus our efficiency analysis on the controlled end-to-end variants.

This systematic comparison naturally serves as a macroarchitecture ablation study of two key design choices: modality selection and cooperation level.
As shown in Table~\ref{tab:efficiency_analysis}, expanding from single-agent (V) to cooperative perception (V2X) primarily increases the number of computational FLOPs because of the processing of additional views,
but it introduces negligible additional parameters (e.g., maintaining 35.36 M for both XET-V and XET-V2X).
Although the comprehensive XET-V2X model requires more computational resources (990.1 GFLOPs), resulting in an inference speed of 3.5 FPS,
this represents a highly justifiable trade-off given its substantial and consistent accuracy improvements over the simpler variants.

From a deployment perspective, several practical considerations should be noted. 
Although feature-level transmission provides a more favorable trade-off between information richness and communication cost than raw data sharing, the actual bandwidth requirement can still become substantial as the number of collaborating agents increases. 
Moreover, in the current experimental setting, communication latency is simulated by frame-level time shifting rather than by a full network-stack emulation, and no dedicated learned feature compression module is integrated into the present framework. 
The reported results therefore reflect the perception-side robustness of the proposed model under delayed and asynchronous observations, rather than an end-to-end evaluation of a complete deployed V2X system. 
In addition, real-world deployment still depends on sufficiently accurate calibration and localization across agents, as well as adequate onboard or edge computing resources to support the transformer-based multimodal cooperative perception pipeline. 
These aspects constitute important directions for future work toward practical large-scale deployment.

\subsection{Ablation Study of the Multimodal Fusion Design}\label{subsubsec:xet-v2x ablation}
\begin{table*}[htpb!]
    \centering
    \caption{\textbf{Performance comparison for different fusion orders based on three vehicle-to-everything (V2X) benchmarks.}
    Fusion order denotes the feature refinement order in the cross-modal attention mechanism:
    \checkmark for Image $\rightarrow$ Point Cloud, blank for Point Cloud $\rightarrow$ Image.
    }
    \label{tab:xet-v2x ablation results}
    \begin{tabular}{P{0.1\textwidth}P{0.05\textwidth}P{0.05\textwidth}P{0.07\textwidth}P{0.07\textwidth}P{0.05\textwidth}P{0.07\textwidth}P{0.07\textwidth}P{0.05\textwidth}P{0.07\textwidth}P{0.07\textwidth}}
        \toprule
        Model & Fusion & \multicolumn{3}{c}{V2X-Seq-SPD} & \multicolumn{3}{c}{V2X-Sim-V2V} & \multicolumn{3}{c}{V2X-Sim-V2I} \\
        \cmidrule(lr){3-5} \cmidrule(lr){6-8} \cmidrule(lr){9-11}
        & Order & mAP$\uparrow$ & AMOTA$\uparrow$ & AMOTP$\downarrow$ & mAP$\uparrow$ & AMOTA$\uparrow$ & AMOTP$\downarrow$ & mAP$\uparrow$ & AMOTA$\uparrow$ & AMOTP$\downarrow$ \\
        \midrule
        \multirow{2}{*}{XET-V} & & 0.456 & 0.421 & 1.235 & 0.495 & 0.454 & 1.180 & 0.399 & 0.358 & 1.350 \\
        & \checkmark & 0.490 & 0.469 & 1.122 & 0.510 & 0.459 & 1.168 & 0.412 & 0.364 & 1.335 \\
        \multirow{2}{*}{XET-V2X} & & 0.707 & 0.663 & 0.823 & 0.735 & 0.700 & 0.748 & 0.797 & 0.813 & 0.508 \\
        & \checkmark & \textbf{0.795} & \textbf{0.787} & \textbf{0.591} & \textbf{0.766} & \textbf{0.731} & \textbf{0.677} & \textbf{0.858} & \textbf{0.819} & \textbf{0.476} \\
        \bottomrule
    \end{tabular}
\end{table*}
This ablation study was performed to assess the design choices used in the proposed multimodal fusion module,
with particular emphasis on the contribution of multimodal interactions and the feature fusion order
in the V2X spatial cross-attention layers.
Quantitative comparisons are reported based on three representative V2X benchmarks,
including a real-world sequential dataset and two simulated cooperative perception settings.

\subsubsection{Effect of the Fusion Order}
To further analyze the robustness of the proposed fusion design,
we evaluate two alternative feature fusion orders in the stacked V2X spatial cross-attention layers:
(i) point cloud features refined with image features (Point $\rightarrow$ Image),
and (ii) image features refined with point cloud features (Image $\rightarrow$ Point).
As shown in Table~\ref{tab:xet-v2x ablation results},
both fusion orders consistently outperform those used in single-modality and single-view baselines,
suggesting that the cross-attention-based fusion mechanism is not sensitive to a specific interaction sequence.

Nevertheless, the Image $\rightarrow$ Point fusion order exhibits slightly superior and more stable performance across all benchmarks.
In particular, XET-V2X with Image $\rightarrow$ Point fusion achieves the highest mAP and AMOTA for all three datasets
while also yielding low AMOTP values, indicating comparatively precise localization.
This trend is consistent across both real-world and simulated scenarios, 
suggesting that refining image features provides informative semantic guidance for subsequent point cloud feature enhancement.

On the basis of these observations, the Image $\rightarrow$ Point fusion order is adopted in the final model.
The ablation results indicate that the performance gains of XET primarily stem from effective cross-modal and cross-view interactions,
while the selected fusion order further improves the accuracy and robustness in V2X cooperative perception scenarios.

\subsection{Qualitative Visualization Results}
\begin{figure*}[htbp!]
    \captionsetup[subfigure]{font=footnotesize, skip=1pt}
    \captionsetup{skip=3pt}
    \centering
    \begin{subfigure}[b]{0.14285\textwidth}
        \includegraphics[width=\linewidth]{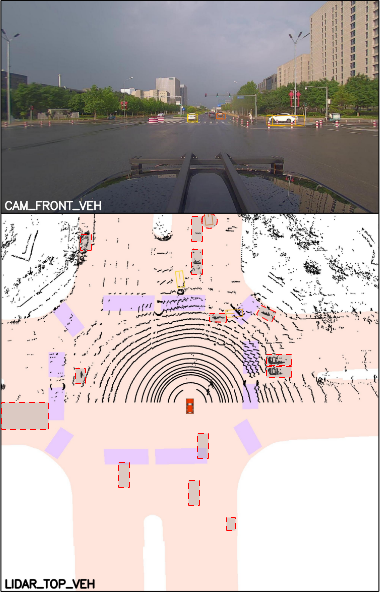}
        \subcaption{ImVoxelNet-V}
        \label{fig:v2x-seq-spd_imvoxelnet-v}
    \end{subfigure}%
    \begin{subfigure}[b]{0.14285\textwidth}
        \includegraphics[width=\linewidth]{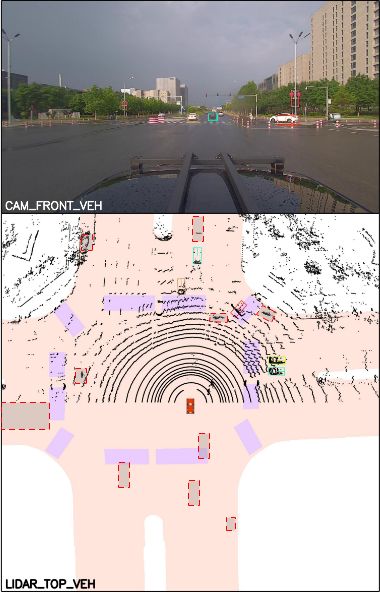}
        \subcaption{PointPillars-V}
        \label{fig:v2x-seq-spd_pointpillars-v}
    \end{subfigure}%
    \begin{subfigure}[b]{0.14285\textwidth}
        \includegraphics[width=\linewidth]{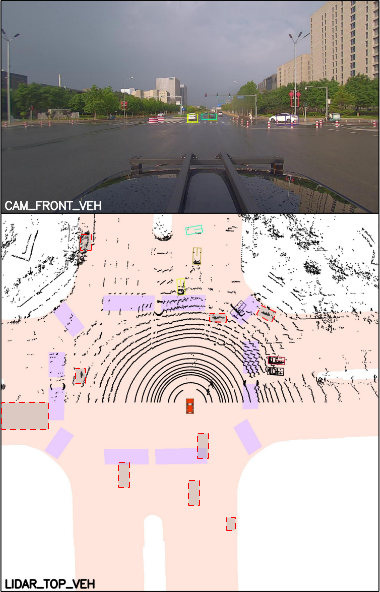}
        \subcaption{MVXNet-V}
        \label{fig:v2x-seq-spd_mvxnet-v}
    \end{subfigure}%
    \begin{subfigure}[b]{0.14285\textwidth}
        \includegraphics[width=\linewidth]{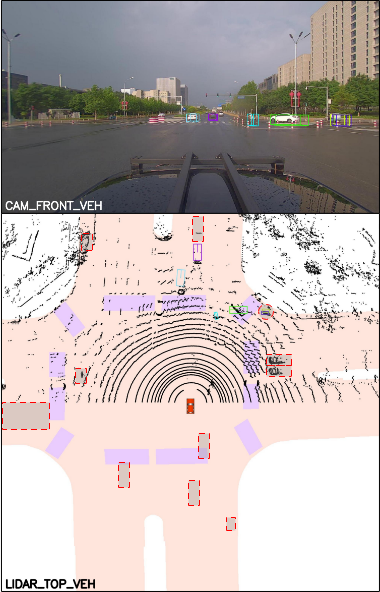}
        \subcaption{CET-V}
        \label{fig:v2x-seq-spd_cet-v}
    \end{subfigure}%
    \begin{subfigure}[b]{0.14285\textwidth}
        \includegraphics[width=\linewidth]{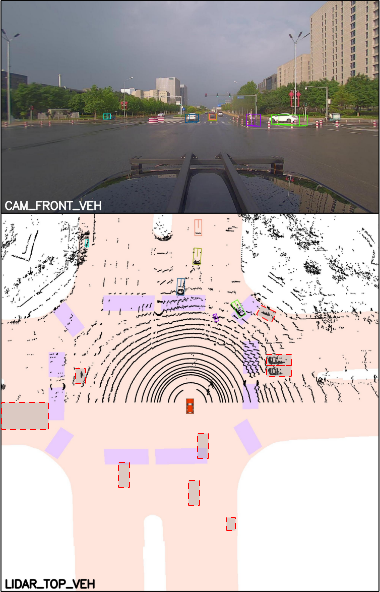}
        \subcaption{LET-V}
        \label{fig:v2x-seq-spd_let-v}
    \end{subfigure}%
    \begin{subfigure}[b]{0.14285\textwidth}
        \includegraphics[width=\linewidth]{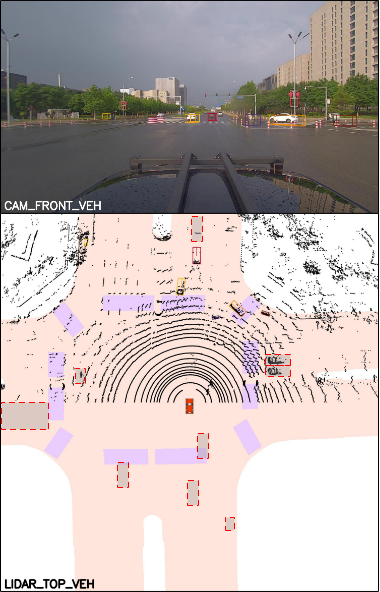}
        \subcaption{XET-V}
        \label{fig:v2x-seq-spd_xet-v}
    \end{subfigure}%
    \begin{subfigure}[b]{0.14285\textwidth}
        \makebox[\linewidth]{}
    \end{subfigure}
    \\[-0.5mm]
    \begin{subfigure}[b]{0.14285\textwidth}
        \includegraphics[width=\linewidth]{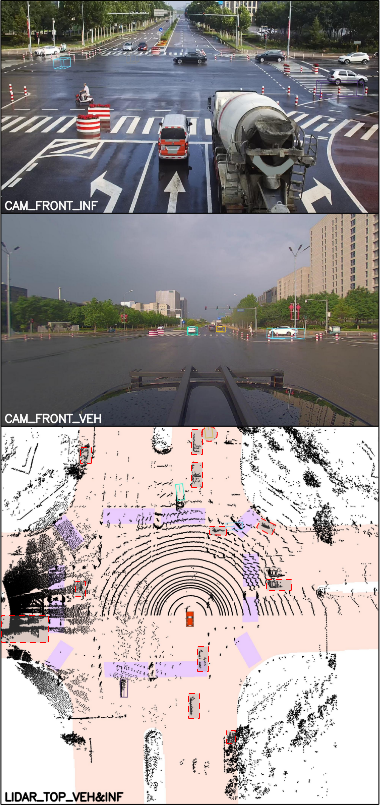}
        \subcaption{ImVoxelNet-V2X}
        \label{fig:v2x-seq-spd_imvoxelnet-v2x}
    \end{subfigure}%
    \begin{subfigure}[b]{0.14285\textwidth}
        \includegraphics[width=\linewidth]{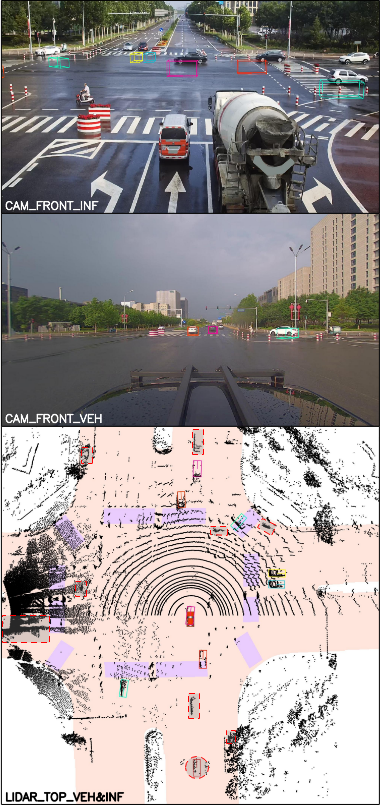}
        \subcaption{PointPillars-V2X}
        \label{fig:v2x-seq-spd_pointpillars-v2x}
    \end{subfigure}%
    \begin{subfigure}[b]{0.14285\textwidth}
        \includegraphics[width=\linewidth]{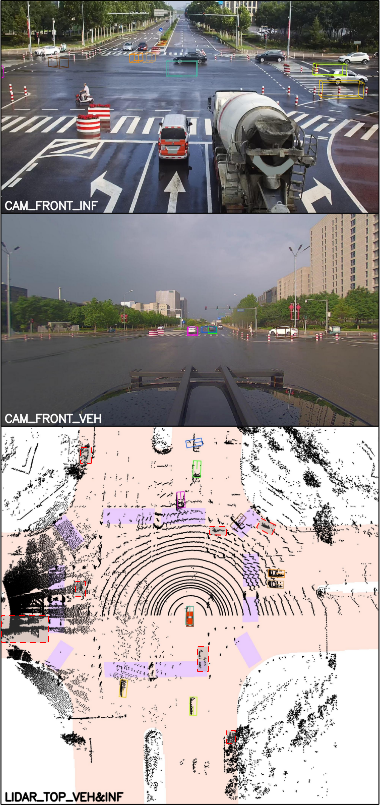}
        \subcaption{MVXNet-V2X}
        \label{fig:v2x-seq-spd_mvxnet-v2x}
    \end{subfigure}%
    \begin{subfigure}[b]{0.14285\textwidth}
        \includegraphics[width=\linewidth]{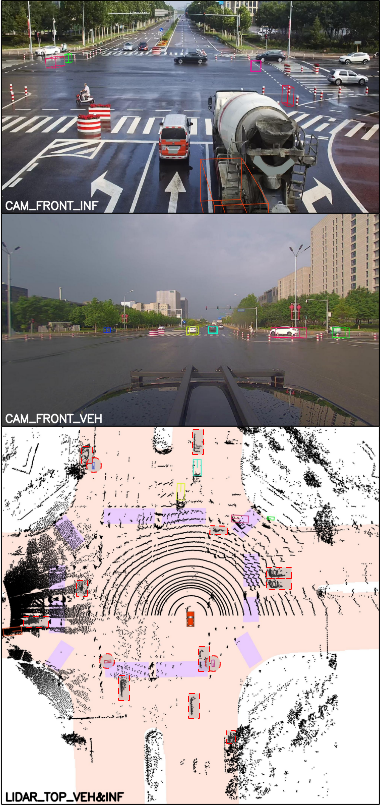}
        \subcaption{CET-V2X}
        \label{fig:v2x-seq-spd_cet-v2x}
    \end{subfigure}%
    \begin{subfigure}[b]{0.14285\textwidth}
        \includegraphics[width=\linewidth]{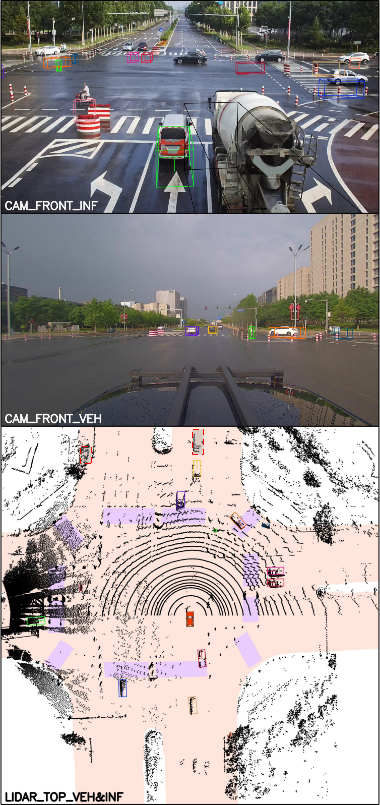}
        \subcaption{LET-V2X}
        \label{fig:v2x-seq-spd_let-v2x}
    \end{subfigure}%
    \begin{subfigure}[b]{0.14285\textwidth}
        \includegraphics[width=\linewidth]{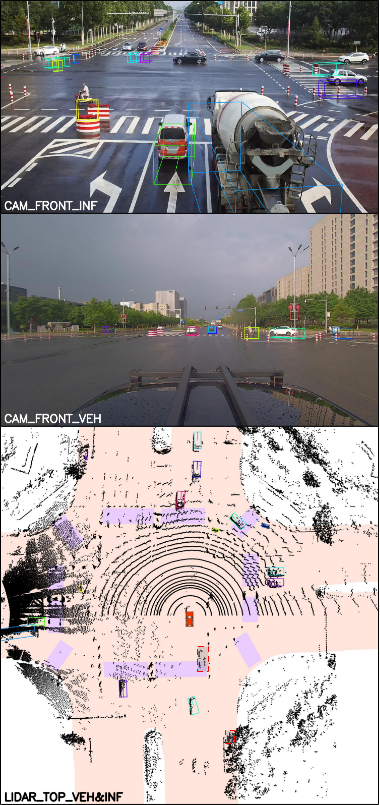}
        \subcaption{XET-V2X}
        \label{fig:v2x-seq-spd_xet-v2x}
    \end{subfigure}%
    \begin{subfigure}[b]{0.14285\textwidth}
        \includegraphics[width=\linewidth]{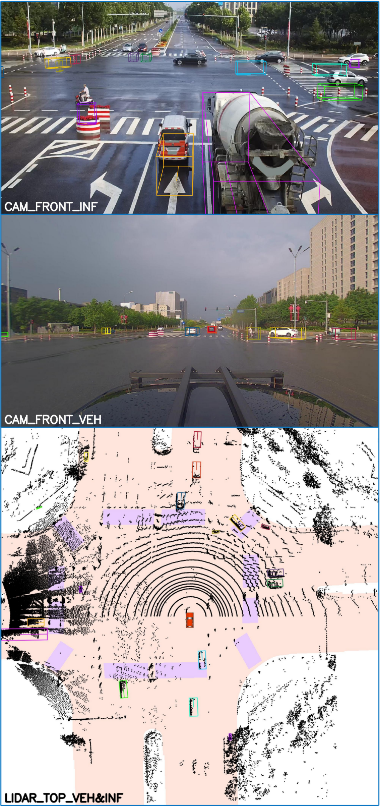}
        \subcaption{Ground Truth}
        \label{fig:v2x-seq-spd_ground-truth}
    \end{subfigure}
    \caption{\textbf{Detailed visualization of temporal perception results based on V2X-Seq-SPD.}
    To highlight fine-scale perception details, each subfigure displays the detection results for a single frame.
    For the single-vehicle (V) setting in (a)-(f), the visualizations comprise the vehicle-side image (top) and the vehicle-side point cloud (bottom).
    For the vehicle-to-everything (V2X) setting in (g)-(l), the visualizations incorporate the infrastructure-side image (top), the vehicle-side image (middle), and the fused vehicle-and-infrastructure point cloud projected into the vehicle coordinate system (bottom).
    (m) represents the ground truth.
    Semitransparent dashed boxes indicate false negatives, andsemitransparent dashed circles denote false positives.}
    \label{fig:visul_v2x-seq-spd}
\end{figure*}

\begin{figure*}[htbp!]
    \captionsetup[subfigure]{font=footnotesize, skip=1pt}
    \captionsetup{skip=3pt}
    \centering
    \begin{subfigure}[b]{0.48\textwidth}
        \includegraphics[width=\linewidth]{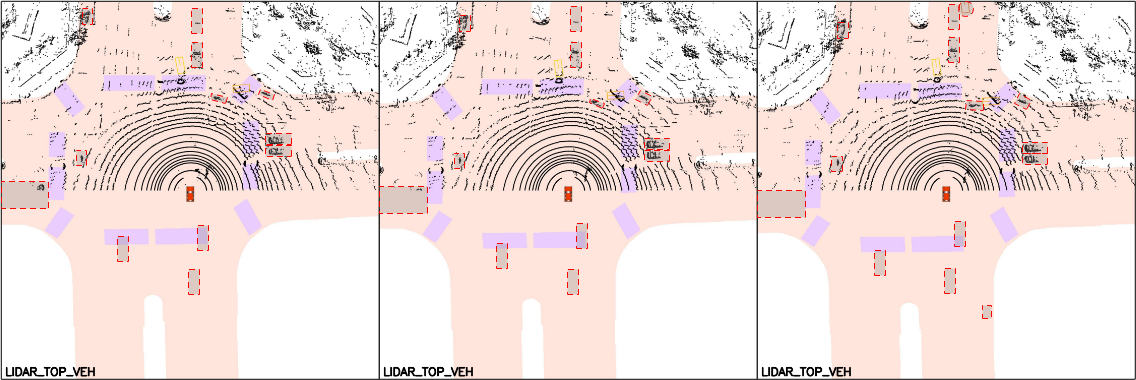}
        \subcaption{ImVoxelNet-V}
        \label{fig:v2x-seq-spd_imvoxelnet-v_seq}
    \end{subfigure}
    \begin{subfigure}[b]{0.48\textwidth}
        \includegraphics[width=\linewidth]{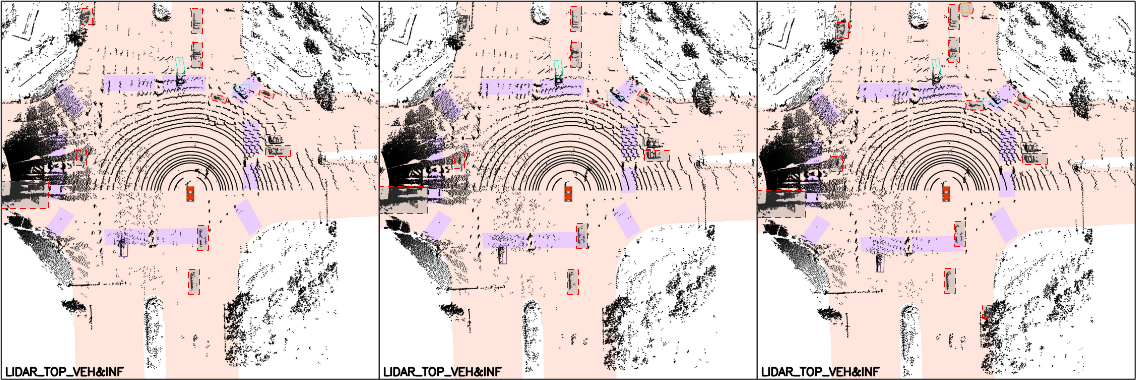}
        \subcaption{ImVoxelNet-V2X}
        \label{fig:v2x-seq-spd_imvoxelnet-v2x_seq}
    \end{subfigure}
    \\[-0.5mm]
    \begin{subfigure}[b]{0.48\textwidth}
        \includegraphics[width=\linewidth]{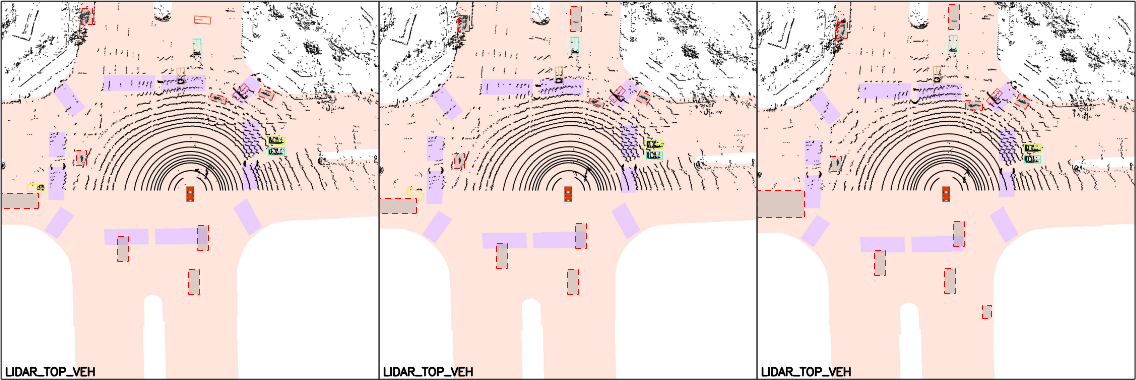}
        \subcaption{PointPillars-V}
        \label{fig:v2x-seq-spd_pointpillars-v_seq}
    \end{subfigure}
    \begin{subfigure}[b]{0.48\textwidth}
        \includegraphics[width=\linewidth]{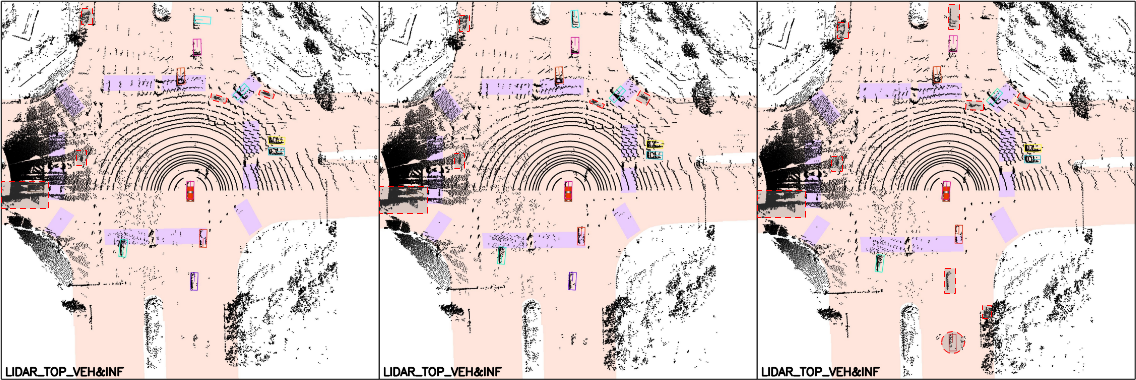}
        \subcaption{PointPillars-V2X}
        \label{fig:v2x-seq-spd_pointpillars-v2x_seq}
    \end{subfigure}
    \\[-0.5mm]
    \begin{subfigure}[b]{0.48\textwidth}
        \includegraphics[width=\linewidth]{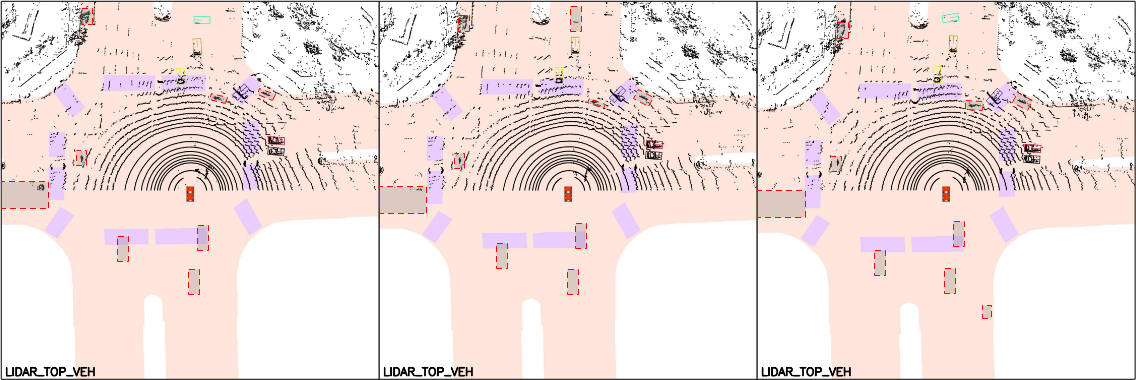}
        \subcaption{MVXNet-V}
        \label{fig:v2x-seq-spd_mvxnet-v_seq}
    \end{subfigure}
    \begin{subfigure}[b]{0.48\textwidth}
        \includegraphics[width=\linewidth]{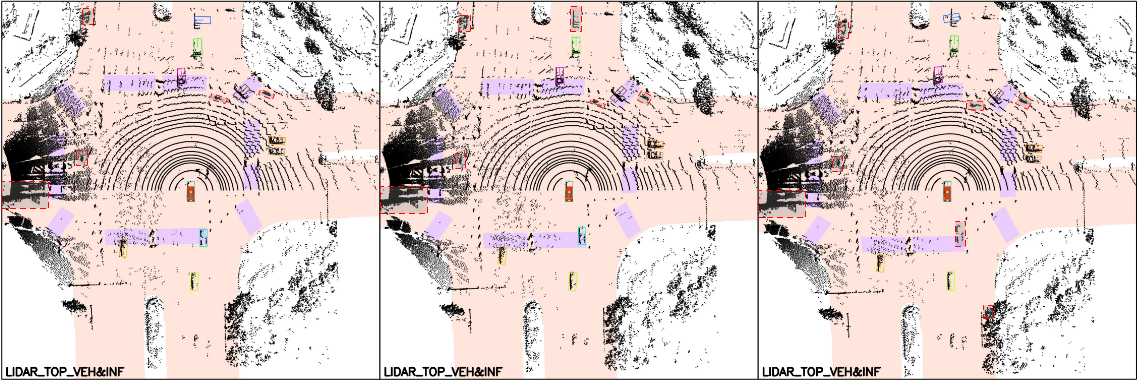}
        \subcaption{MVXNet-V2X}
        \label{fig:v2x-seq-spd_mvxnet-v2x_seq}
    \end{subfigure}
    \\[-0.5mm]
    \begin{subfigure}[b]{0.48\textwidth}
        \includegraphics[width=\linewidth]{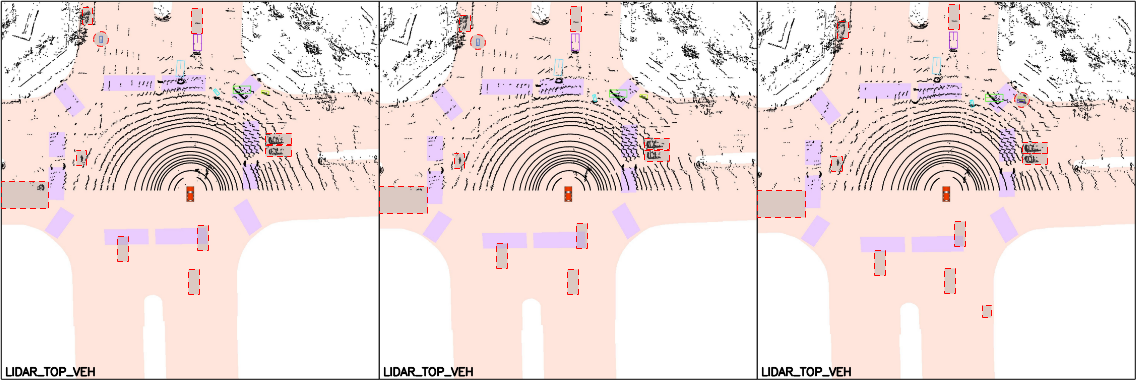}
        \subcaption{CET-V}
        \label{fig:v2x-seq-spd_cet-v_seq}
    \end{subfigure}
    \begin{subfigure}[b]{0.48\textwidth}
        \includegraphics[width=\linewidth]{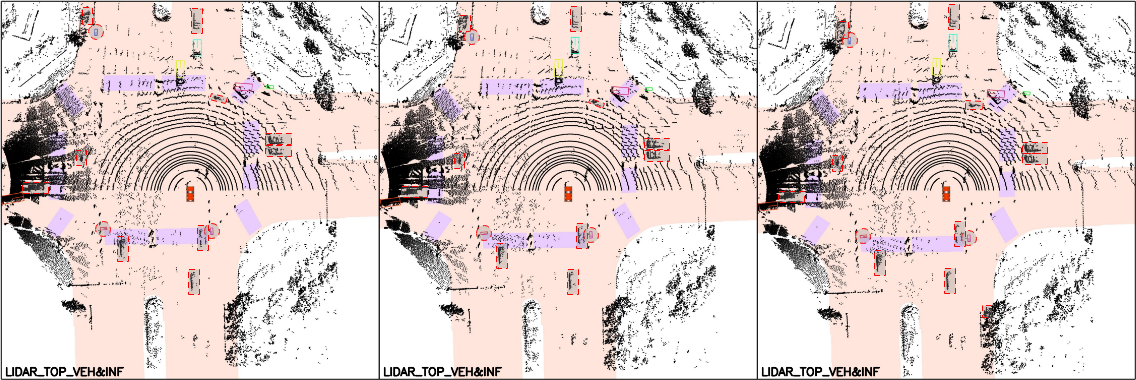}
        \subcaption{CET-V2X}
        \label{fig:v2x-seq-spd_cet-v2x_seq}
    \end{subfigure}
    \\[-0.5mm]
    \begin{subfigure}[b]{0.48\textwidth}
        \includegraphics[width=\linewidth]{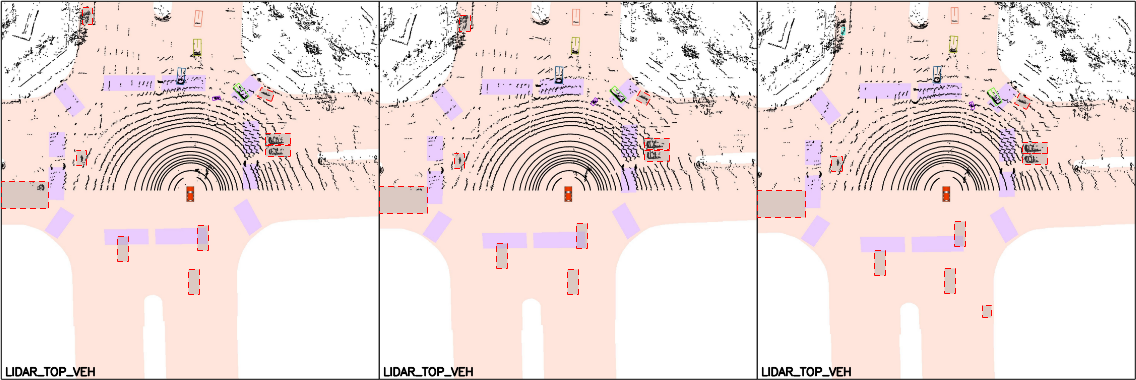}
        \subcaption{LET-V}
        \label{fig:v2x-seq-spd_let-v_seq}
    \end{subfigure}
    \begin{subfigure}[b]{0.48\textwidth}
        \includegraphics[width=\linewidth]{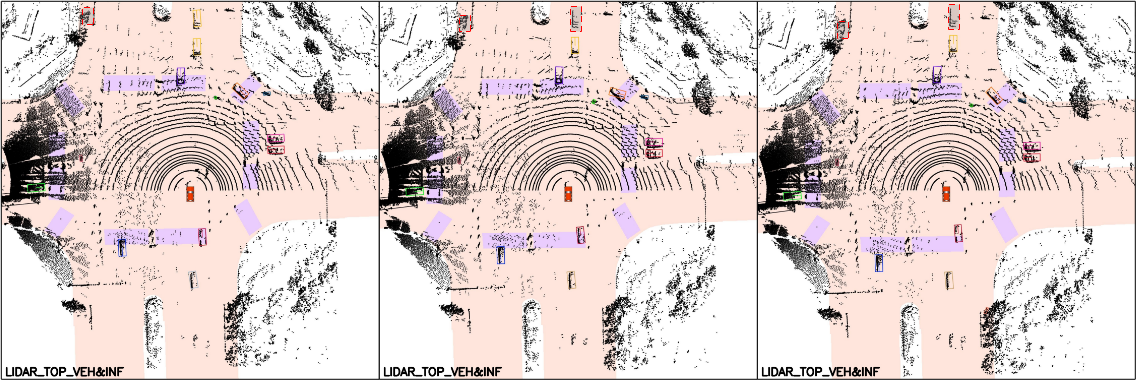}
        \subcaption{LET-V2X}
        \label{fig:v2x-seq-spd_let-v2x_seq}
    \end{subfigure}
    \\[-0.5mm]
    \begin{subfigure}[b]{0.48\textwidth}
        \includegraphics[width=\linewidth]{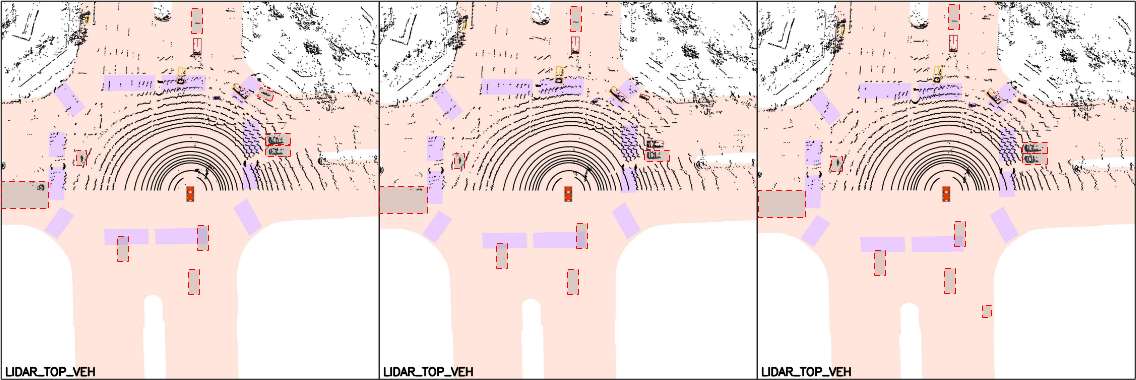}
        \subcaption{XET-V}
        \label{fig:v2x-seq-spd_xet-v_seq}
    \end{subfigure}
    \begin{subfigure}[b]{0.48\textwidth}
        \includegraphics[width=\linewidth]{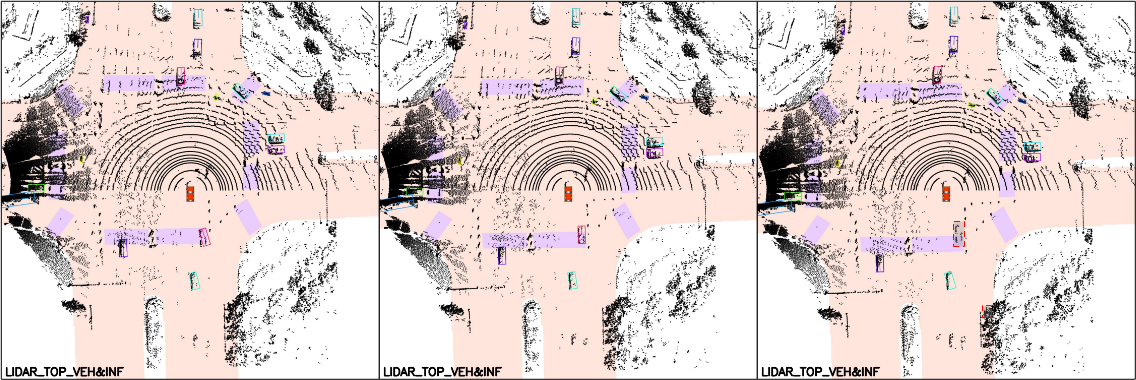}
        \subcaption{XET-V2X}
        \label{fig:v2x-seq-spd_xet-v2x_seq}
    \end{subfigure}
    \\[-0.5mm]
    \begin{subfigure}[b]{0.48\textwidth}
        \includegraphics[width=\linewidth]{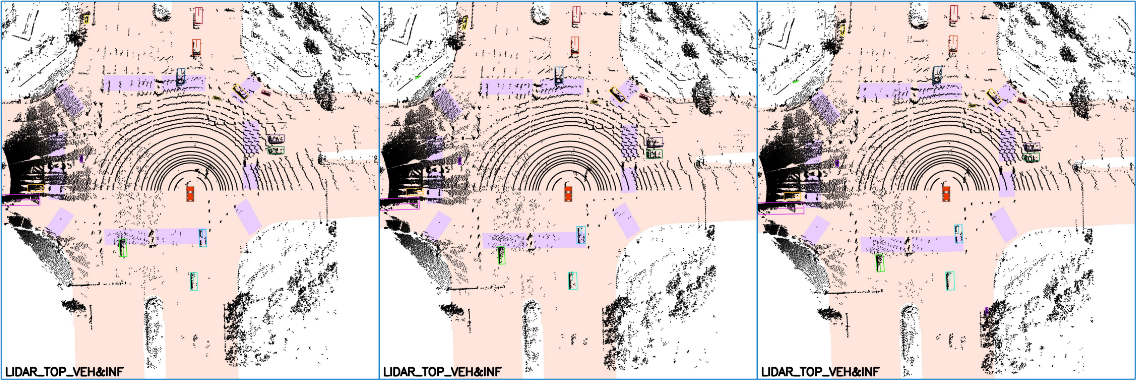}
        \subcaption{Ground Truth}
        \label{fig:v2x-seq-spd_ground-truth_seq}
    \end{subfigure}
    \caption{\textbf{Sequential visualization of temporal perception results based on V2X-Seq-SPD.}
    The left column shows single-vehicle (V) results, and the right column shows the corresponding V2X results. 
    Consistent colors indicate the same tracked object across frames. 
    Semitransparent dashed boxes denote false negatives, and semitransparent dashed circles denote false positives. 
    (m) shows the ground truth.}
    \label{fig:visul_v2x-seq-spd_seq}
\end{figure*}

To complement the quantitative evaluation, we present qualitative results to further assess the perception behavior of the proposed models under diverse V2X scenarios.
Representative inference results based on the V2X-Seq-SPD~\cite{yu2023v2x}, V2X-Sim-V2V, and V2X-Sim-V2I~\cite{li2022v2x} benchmarks are presented.
Owing to space limitations, only selected examples are shown in the main paper, while additional qualitative visualizations are provided in the Appendix.

Fig.~\ref{fig:visul_v2x-seq-spd} presents detailed single-frame qualitative results based on the real-world V2X-Seq-SPD benchmark, 
visualizing representative detection results under both single-vehicle (V) and vehicle-to-everything (V2X) settings. 
For the V setting, each subfigure includes the vehicle-side image and the vehicle-side point cloud, 
whereas for the V2X setting, each subfigure incorporates the infrastructure-side image, the vehicle-side image, 
and the fused vehicle-and-infrastructure point cloud projected into the vehicle coordinate system. 
Furthermore, Fig.~\ref{fig:visul_v2x-seq-spd_seq} presents a representative temporal qualitative comparison on the same benchmark,
illustrating the differences between single-vehicle and cooperative perception schemes, as well as between unimodal and multimodal designs. 
To facilitate a direct and fair comparison of temporal continuity, each subfigure shows the same three consecutive frames across all compared models. 
The visualizations compare the predicted 3-D detection and tracking results with the ground-truth annotations. 
Consistent colors across frames indicate the same tracked object over time, enabling an intuitive assessment of detection completeness and temporal continuity. 
Semitransparent dashed boxes indicate false negatives, while semitransparent dashed circles denote false positives,
providing visual cues for analyzing failure cases under occlusion, dense traffic, and long-term tracking conditions.

Additional qualitative visualizations for the V2X-Sim-V2V and V2X-Sim-V2I benchmarks are provided in Figs.~\ref{fig:visul_v2x-sim-v2v}--\ref{fig:visul_v2x-sim-v2i_seq}.

\subsubsection{Single-Vehicle vs. Cooperative Perception}
As illustrated in the qualitative visualizations,
single-vehicle perception models suffer from noticeable performance degradation in scenarios with extensive occlusion issues or limited sensor coverage.
Missed detections and fragmented trajectories frequently appear when targets are partially observed or temporarily occluded by surrounding objects.
By contrast, V2X-based cooperative perception benefits from complementary viewpoints provided by multiple agents,
enabling more complete object recovery and visibly smoother temporal associations for challenging traffic scenes.

\subsubsection{Single-Modal vs. Multimodal Perception}
The visual results further reveal the limitations of single-modal perception.
Camera-only methods often fail in regions with poor illumination or severe visual clutter,
while LiDAR-only approaches may miss distant or sparsely sampled targets.
By jointly leveraging visual semantics and geometric cues,
multimodal fusion effectively compensates for these weaknesses,
leading to more accurate object localization and improved trajectory continuity in the rendered visualizations.

\subsubsection{Joint Multiview and Multimodal Fusion}
When multiview collaboration and multimodal fusion are jointly enabled,
XET-V2X consistently produces the most stable and complete perception results.
As shown in the visual examples,
XET-V2X successfully recovers objects located in blind spots,
maintains consistent object identities across long temporal windows,
and reduces trajectory fragmentation in dense multiagent interactions.
These qualitative observations corroborate the quantitative improvements reported earlier and highlight the synergistic benefits of combining multiview and multimodal cues
for robust 3-D spatiotemporal perception in complex V2X scenarios.

The qualitative results indicate that the proposed XET-V2X framework not only improves detection accuracy but also enhances temporal consistency and robustness,
particularly under challenging conditions involving occlusions, dense traffic, and extended temporal tracking.

\section{Conclusion}
In this work, XET-V2X, an end-to-end spatiotemporal perception framework designed for multiview cooperative and multimodal fusion in V2X environments, is presented.
By jointly modeling temporal dynamics, cross-view collaboration, and multimodal feature interactions, 
the proposed approach effectively addresses the performance degradation issue commonly observed for single-view perception systems,
unimodal perception models, and conventional tracking-by-detection cooperative frameworks in cases with occlusions, limited fields of view, and calibration uncertainties.

Comprehensive experiments and systematic comparisons based on V2X-Seq-SPD~\cite{yu2023v2x}, V2X-Sim-V2I~\cite{li2022v2x},
and V2X-Sim-V2V~\cite{li2022v2x} demonstrate the effectiveness of the proposed framework from both quantitative and qualitative perspectives.
The quantitative results confirm that multiview collaboration and multimodal fusion each contribute substantially to performance gains in detection and tracking, 
whereas their unified integration in XET-V2X consistently achieves superior overall performance compared to that of other models.
Ablation studies further verify that the proposed cross-modal fusion design provides robust and stable improvements regardless of the feature extraction order.

Qualitative visualizations offer intuitive evidence that supports the quantitative findings.
Single-view or unimodal models frequently yield missed detections and unstable temporal associations in complex traffic scenarios, whereas cooperative models significantly alleviate these issues.
In particular, XET-V2X results in the fewest missed targets and the most consistent long-term tracking, highlighting the complementary advantages of multiview collaboration and multimodal fusion.

Despite these promising results, several limitations remain.
The current framework still incurs nontrivial communication and computation costs, especially when the number of collaborating agents increases.
Moreover, the latency modeling in the present experiments is based on frame-level delay simulation rather than full network-stack emulation, 
and no dedicated learned feature compression module is integrated into the current implementation.
Practical deployment also depends on sufficiently accurate calibration and localization across agents, as well as adequate onboard or edge computing resources.
Future research will therefore focus on integrating communication-aware feature compression and adaptive transmission strategies, 
conducting more realistic deployment-oriented evaluations under practical networking conditions, 
extending the framework to more heterogeneous V2X settings, and further improving efficiency for real-time cooperative perception.

Overall, the proposed XET-V2X framework demonstrates strong robustness and generalizability under both real-world and simulated V2X settings, 
making it a promising solution for scalable and reliable cooperative spatiotemporal perception in autonomous driving systems. 
The implementation of XET-V2X is publicly available at \url{https://github.com/yangzvv/XET-V2X}.

\appendix
\section{Qualitative Visualization based on V2X-Sim Datasets}
\begin{figure*}[hbp!] 
    \centering
    \begin{subfigure}[b]{0.14285\textwidth}
        \includegraphics[width=\linewidth]{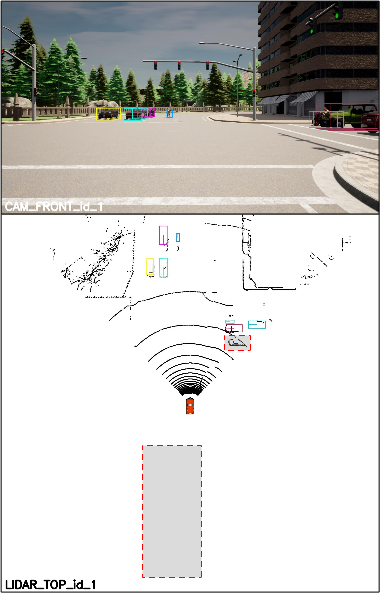}
        \subcaption{CET-V}
        \label{fig:v2x-sim-v2v_cet-v}
    \end{subfigure}%
    \begin{subfigure}[b]{0.14285\textwidth}
        \includegraphics[width=\linewidth]{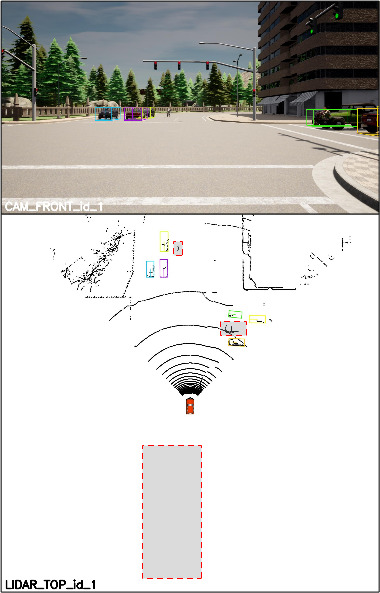}
        \subcaption{LET-V}
        \label{fig:v2x-sim-v2v_let-v}
    \end{subfigure}%
    \begin{subfigure}[b]{0.14285\textwidth}
        \includegraphics[width=\linewidth]{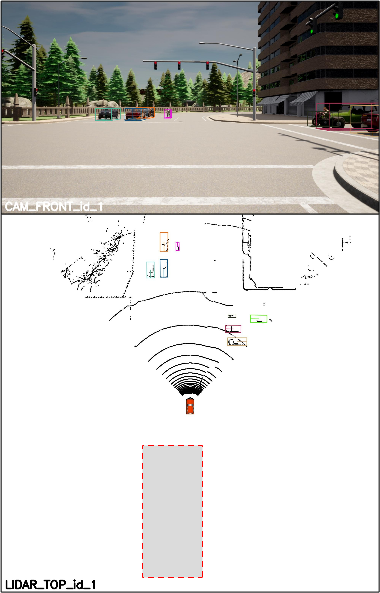}
        \subcaption{XET-V}
        \label{fig:v2x-sim-v2v_xet-v}
    \end{subfigure}%
    \begin{subfigure}[b]{0.14285\textwidth}
        \includegraphics[width=\linewidth]{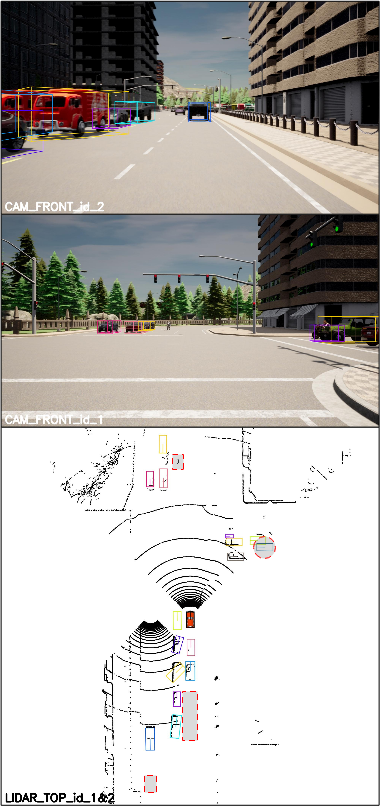}
        \subcaption{CET-V2X}
        \label{fig:v2x-sim-v2v_cet-v2x}
    \end{subfigure}%
    \begin{subfigure}[b]{0.14285\textwidth}
        \includegraphics[width=\linewidth]{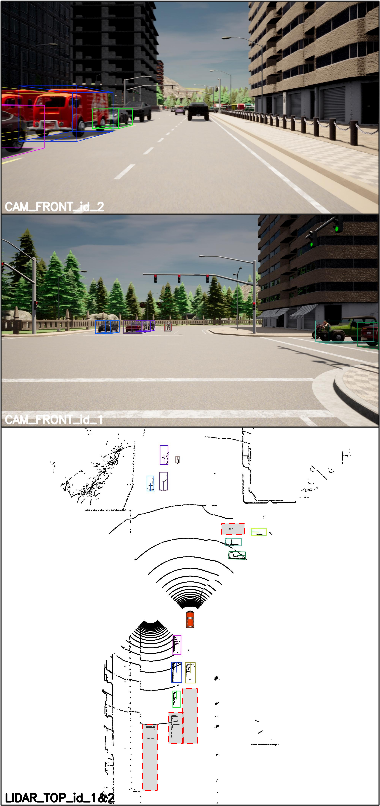}
        \subcaption{LET-V2X}
        \label{fig:v2x-sim-v2v_let-v2x}
    \end{subfigure}%
    \begin{subfigure}[b]{0.14285\textwidth}
        \includegraphics[width=\linewidth]{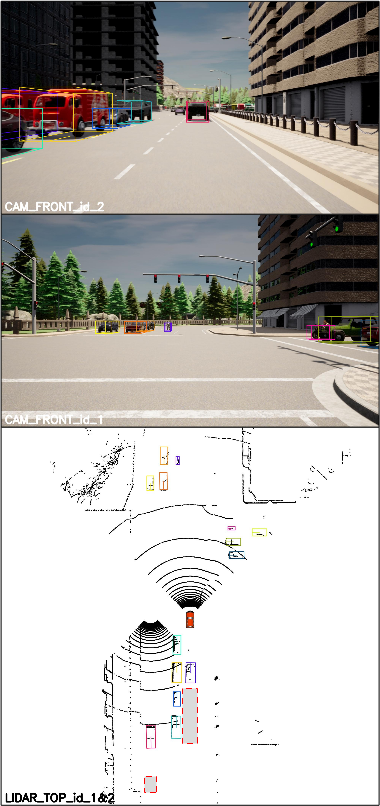}
        \subcaption{XET-V2X}
        \label{fig:v2x-sim-v2v_xet-v2x}
    \end{subfigure}%
    \begin{subfigure}[b]{0.14285\textwidth}
        \includegraphics[width=\linewidth]{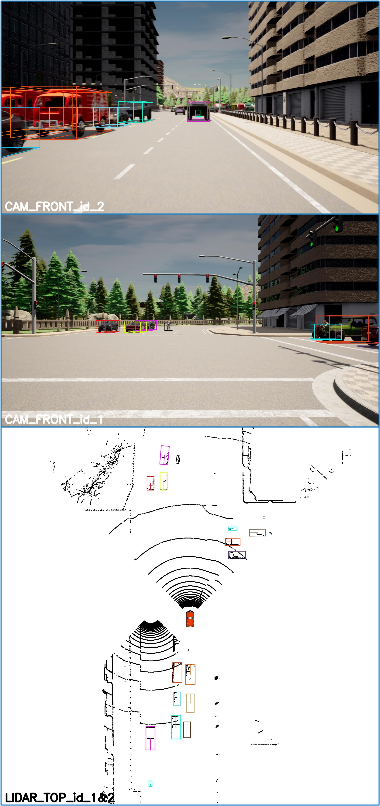}
        \subcaption{Ground Truth}
        \label{fig:v2x-sim-v2v_ground-truth}
    \end{subfigure}
    \caption{\textbf{Detailed visualization of the temporal perception results based on V2X-Sim-V2V.}
    Subfigures (a)-(f) and (g)-(l) present the results for the single-vehicle (V) and vehicle-to-everything (V2X) settings, respectively. (m) is the ground truth.
    The single-frame layout, sensor views, and detection annotations strictly follow the definitions in Fig.~\ref{fig:visul_v2x-seq-spd}.}
    \label{fig:visul_v2x-sim-v2v}

    \vspace{3em} 

    \begin{subfigure}[b]{0.14285\textwidth}
        \includegraphics[width=\linewidth]{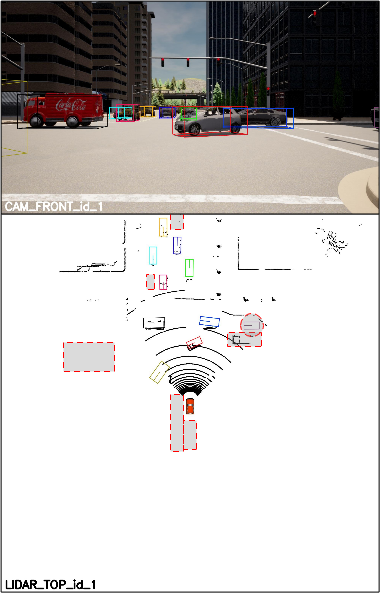}
        \subcaption{CET-V}
        \label{fig:v2x-sim-v2i_cet-v}
    \end{subfigure}%
    \begin{subfigure}[b]{0.14285\textwidth}
        \includegraphics[width=\linewidth]{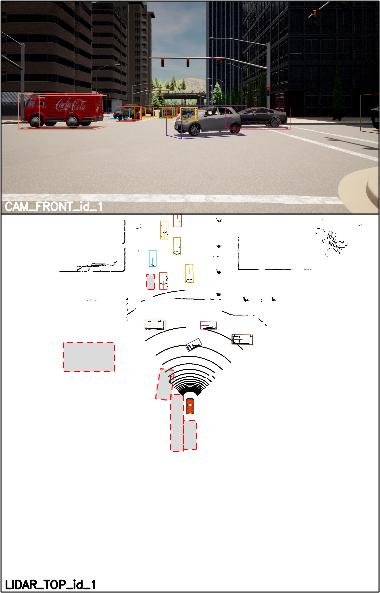}
        \subcaption{LET-V}
        \label{fig:v2x-sim-v2i_let-v}
    \end{subfigure}%
    \begin{subfigure}[b]{0.14285\textwidth}
        \includegraphics[width=\linewidth]{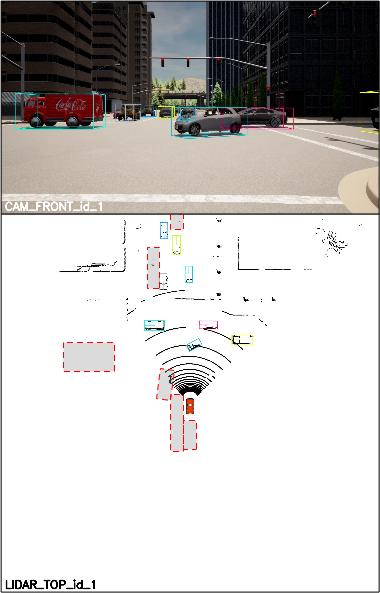}
        \subcaption{XET-V}
        \label{fig:v2x-sim-v2i_xet-v}
    \end{subfigure}%
    \begin{subfigure}[b]{0.14285\textwidth}
        \includegraphics[width=\linewidth]{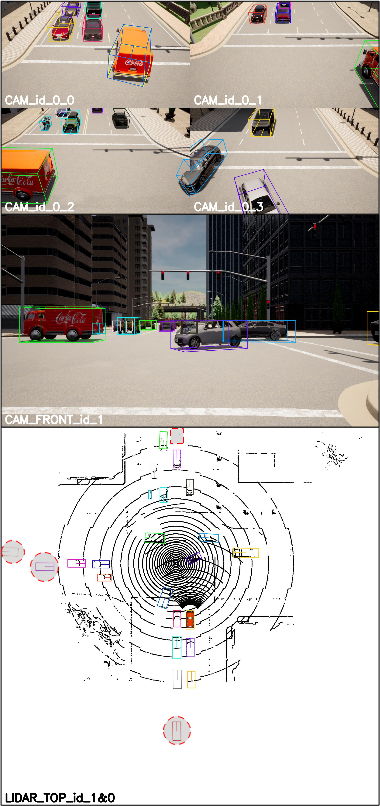}
        \subcaption{CET-V2X}
        \label{fig:v2x-sim-v2i_cet-v2x}
    \end{subfigure}%
    \begin{subfigure}[b]{0.14285\textwidth}
        \includegraphics[width=\linewidth]{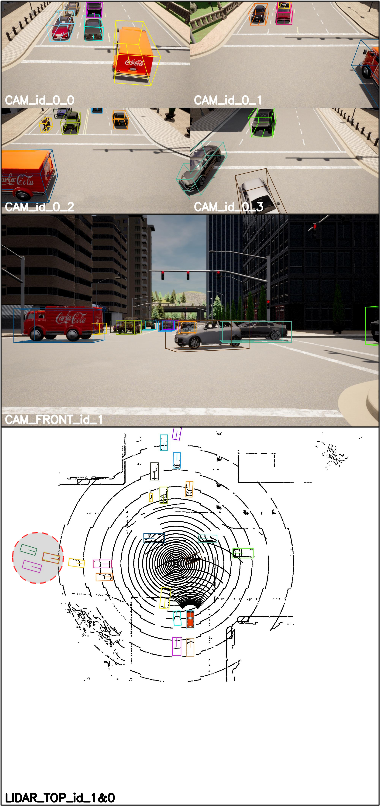}
        \subcaption{LET-V2X}
        \label{fig:v2x-sim-v2i_let-v2x}
    \end{subfigure}%
    \begin{subfigure}[b]{0.14285\textwidth}
        \includegraphics[width=\linewidth]{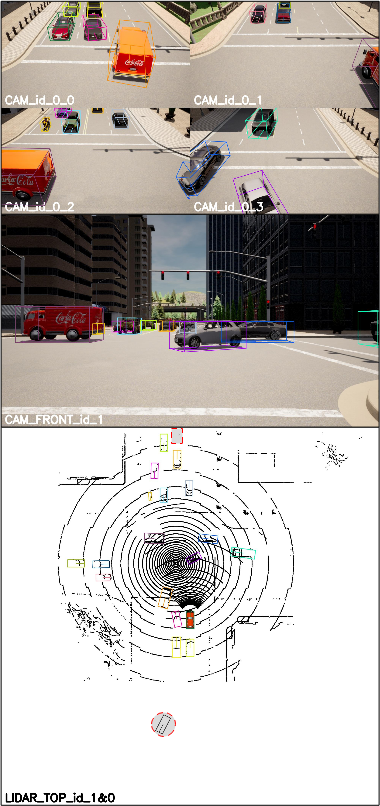}
        \subcaption{XET-V2X}
        \label{fig:v2x-sim-v2i_xet-v2x}
    \end{subfigure}%
    \begin{subfigure}[b]{0.14285\textwidth}
        \includegraphics[width=\linewidth]{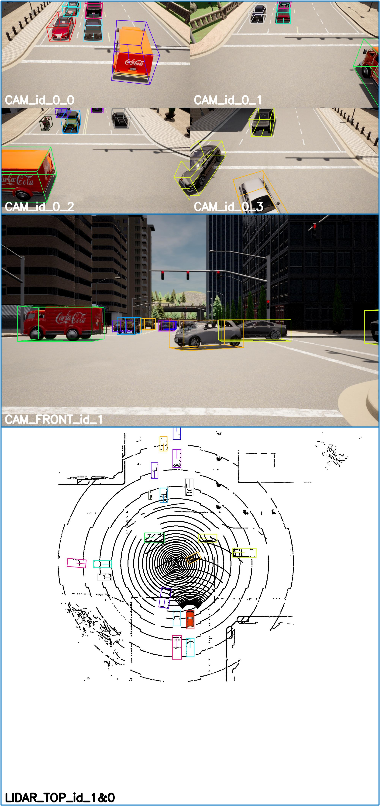}
        \subcaption{Ground Truth}
        \label{fig:v2x-sim-v2i_ground-truth}
    \end{subfigure}
    \caption{\textbf{Detailed visualization of the temporal perception results based on V2X-Sim-V2I.}
    Subfigures (a)-(f) and (g)-(l) present the results for the single-vehicle (V) and vehicle-to-everything (V2X) settings, respectively. (m) is the ground truth.
    The single-frame layout, sensor views, and detection annotations strictly follow the definitions in Fig.~\ref{fig:visul_v2x-seq-spd}.}
    \label{fig:visul_v2x-sim-v2i}
\end{figure*}

In this section, the qualitative results obtained for the V2X-Sim-V2V and V2X-Sim-V2I datasets~\cite{li2022v2x} are visualized.
The predicted 3-D detection and tracking results are compared with ground-truth annotations.
Consistent colors indicate the same tracked objects across time.
Semitransparent dashed boxes denote false negatives (missed detections), 
and semitransparent dashed circles indicate false positives.

\begin{figure}[htbp!]
    \centering
    \begin{subfigure}[b]{0.42\textwidth}
        \includegraphics[width=\linewidth]{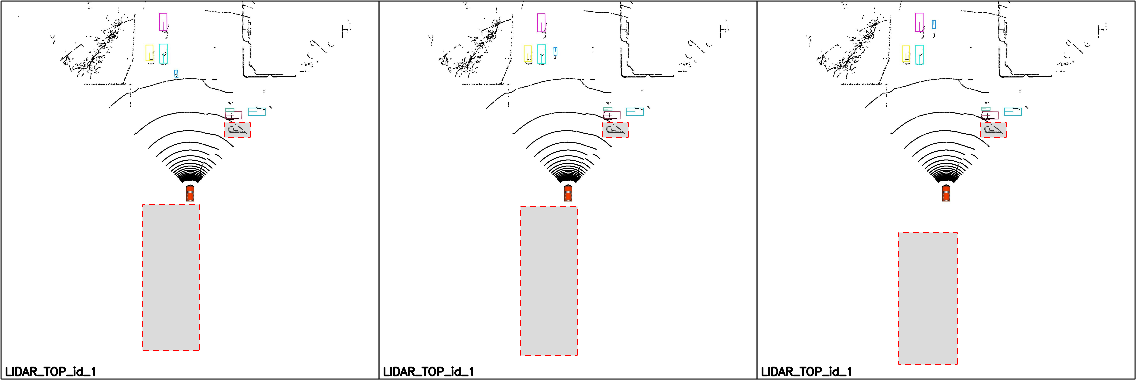}
        \subcaption{CET-V}
        \label{fig:v2x-sim-v2v_cet-v_seq}
    \end{subfigure}
    \\[2mm]
    \begin{subfigure}[b]{0.42\textwidth}
        \includegraphics[width=\linewidth]{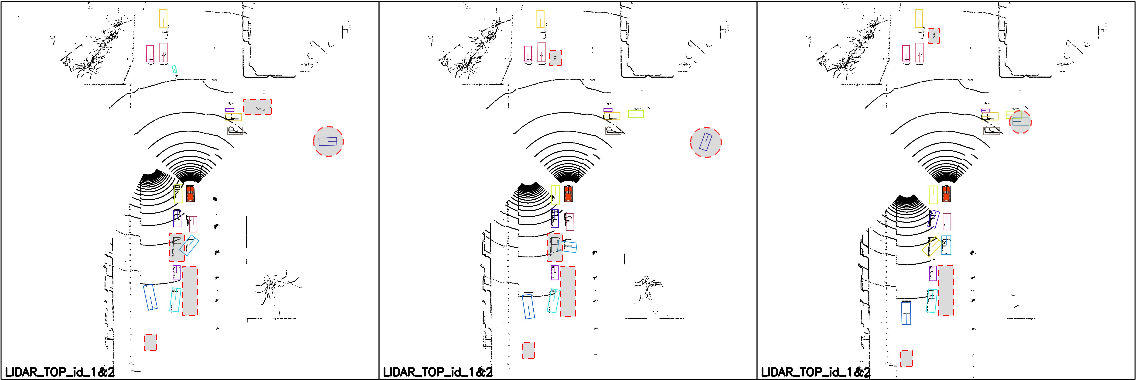}
        \subcaption{CET-V2X}
        \label{fig:v2x-sim-v2v_cet-v2x_seq}
    \end{subfigure}
    \\[2mm]
    \begin{subfigure}[b]{0.42\textwidth}
        \includegraphics[width=\linewidth]{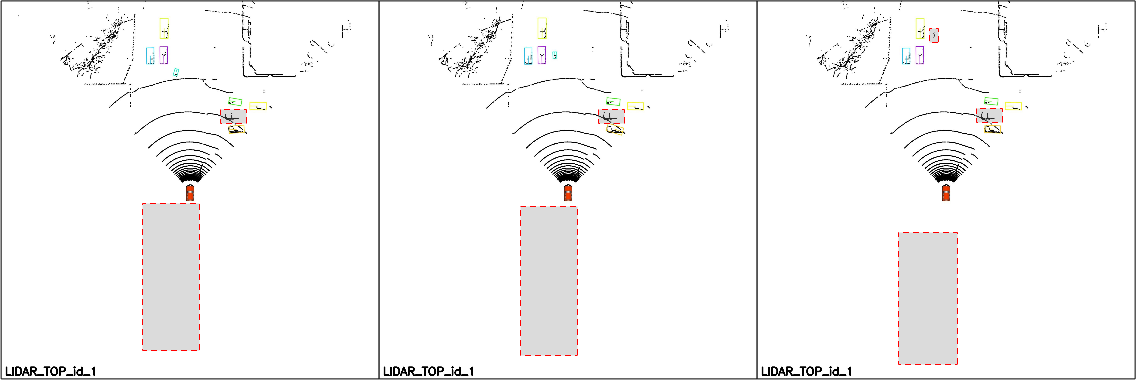}
        \subcaption{LET-V}
        \label{fig:v2x-sim-v2v_let-v_seq}
    \end{subfigure}
    \\[2mm]
    \begin{subfigure}[b]{0.42\textwidth}
        \includegraphics[width=\linewidth]{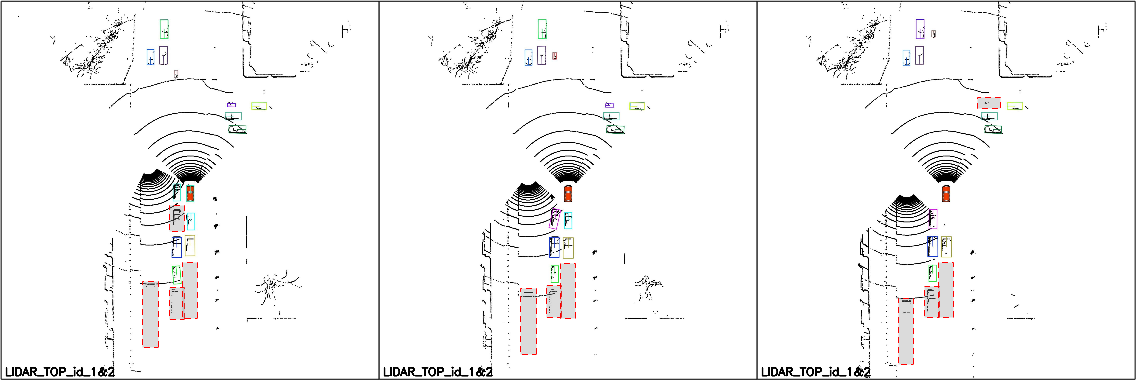}
        \subcaption{LET-V2X}
        \label{fig:v2x-sim-v2v_let-v2x_seq}
    \end{subfigure}
    \\[2mm]
    \begin{subfigure}[b]{0.42\textwidth}
        \includegraphics[width=\linewidth]{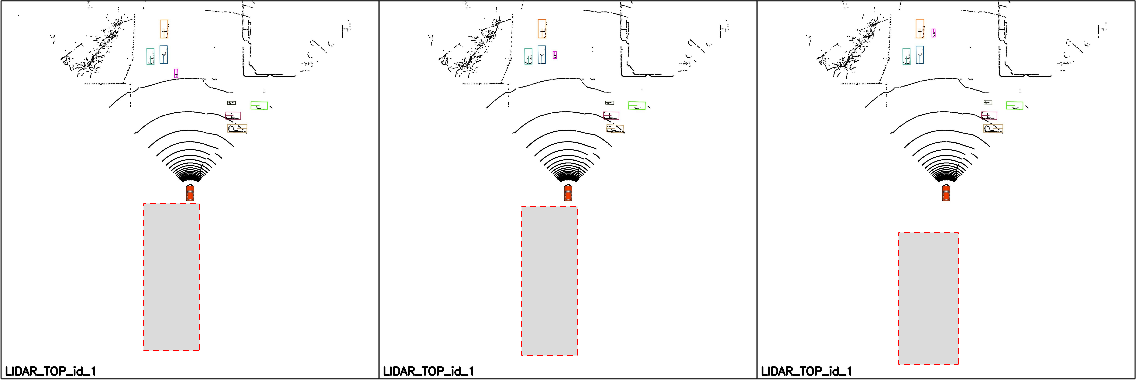}
        \subcaption{XET-V}
        \label{fig:v2x-sim-v2v_xet-v_seq}
    \end{subfigure}
    \\[2mm]
    \begin{subfigure}[b]{0.42\textwidth}
        \includegraphics[width=\linewidth]{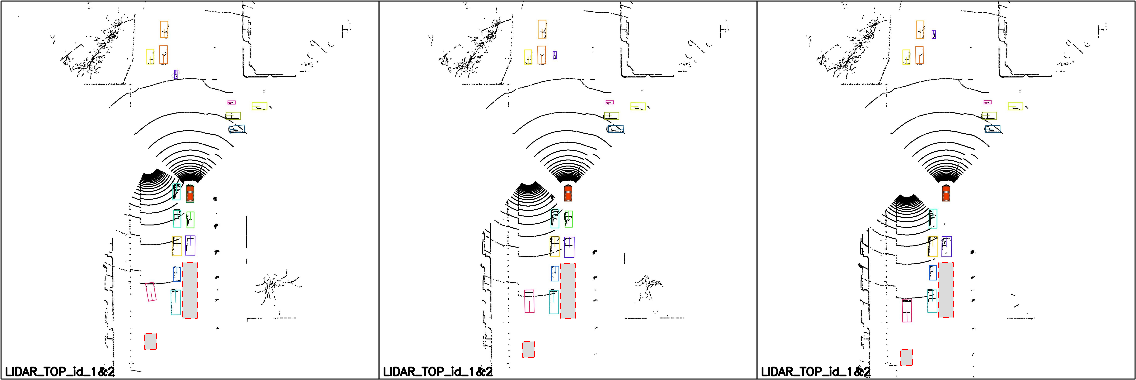}
        \subcaption{XET-V2X}
        \label{fig:v2x-sim-v2v_xet-v2x_seq}
    \end{subfigure}
    \\[2mm]
    \begin{subfigure}[b]{0.42\textwidth}
        \includegraphics[width=\linewidth]{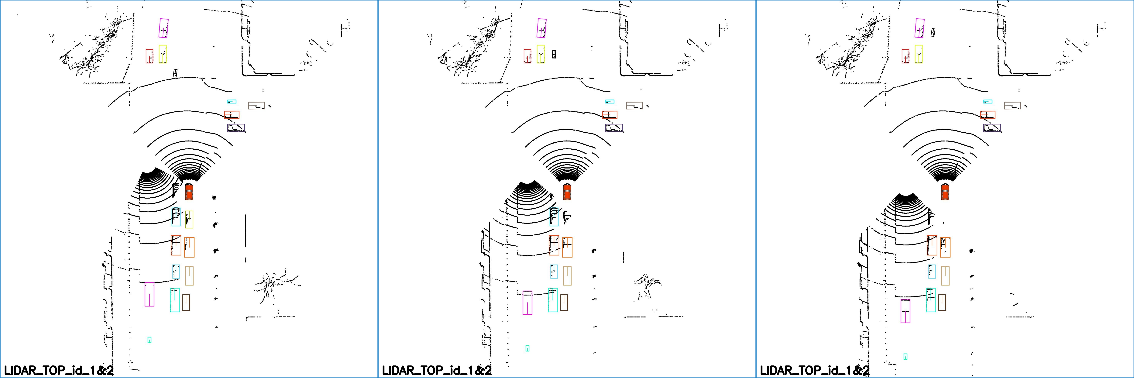}
        \subcaption{Ground Truth}
        \label{fig:v2x-sim-v2v_ground-truth_seq}
    \end{subfigure}
    \caption{\textbf{Sequential visualization of the temporal perception results based on V2X-Sim-V2V.} 
    Subfigures (a)-(c) and (d)-(f) present the results under the single-vehicle (V) and vehicle-to-everything (V2X) settings, respectively. (g) is the ground truth. 
    The three-frame sequence layout, sensor views, and tracking annotations strictly follow the definitions in Fig.~\ref{fig:visul_v2x-seq-spd_seq}.}
    \label{fig:visul_v2x-sim-v2v_seq}
\end{figure}

\begin{figure}[htbp!]
    \centering
    \begin{subfigure}[b]{0.42\textwidth}
        \includegraphics[width=\linewidth]{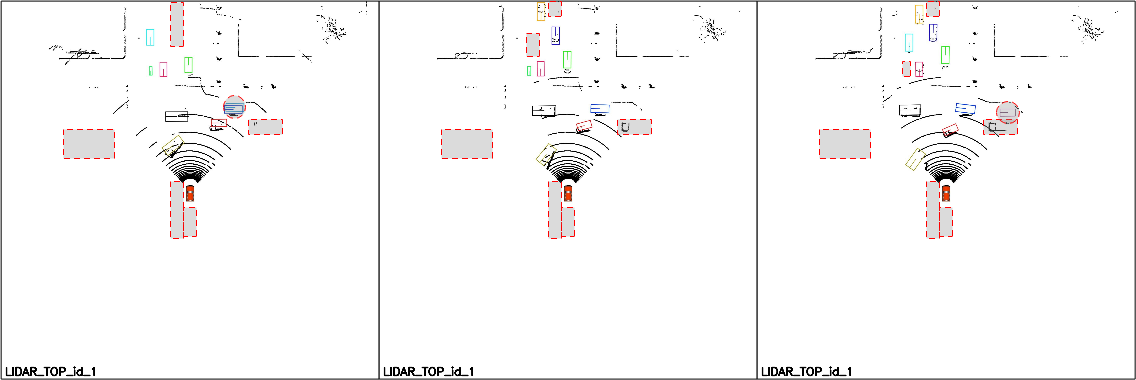}
        \subcaption{CET-V}
        \label{fig:v2x-sim-v2i_cet-v_seq}
    \end{subfigure}
    \\[2mm]
    \begin{subfigure}[b]{0.42\textwidth}
        \includegraphics[width=\linewidth]{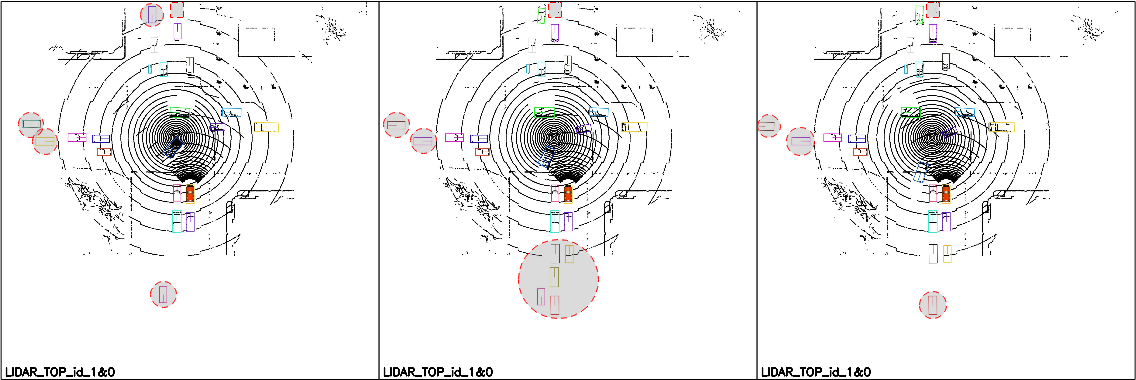}
        \subcaption{CET-V2X}
        \label{fig:v2x-sim-v2i_cet-v2x_seq}
    \end{subfigure}
    \\[2mm]
    \begin{subfigure}[b]{0.42\textwidth}
        \includegraphics[width=\linewidth]{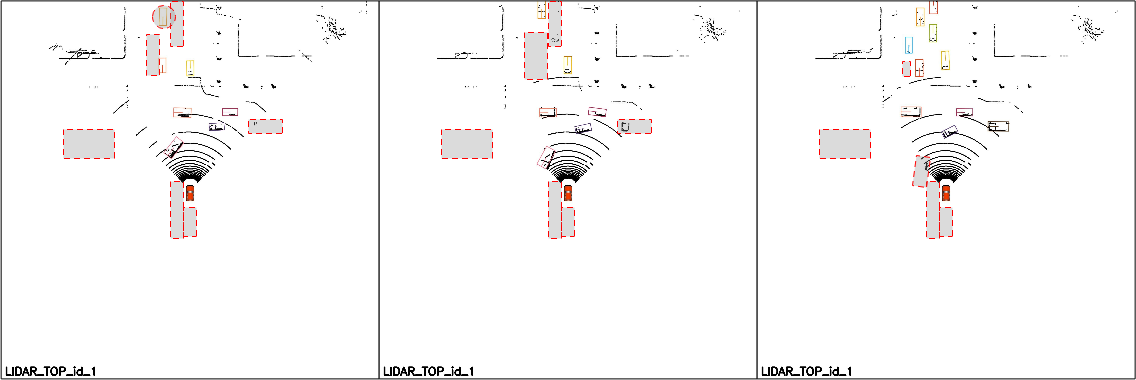}
        \subcaption{LET-V}
        \label{fig:v2x-sim-v2i_let-v_seq}
    \end{subfigure}
    \\[2mm]
    \begin{subfigure}[b]{0.42\textwidth}
        \includegraphics[width=\linewidth]{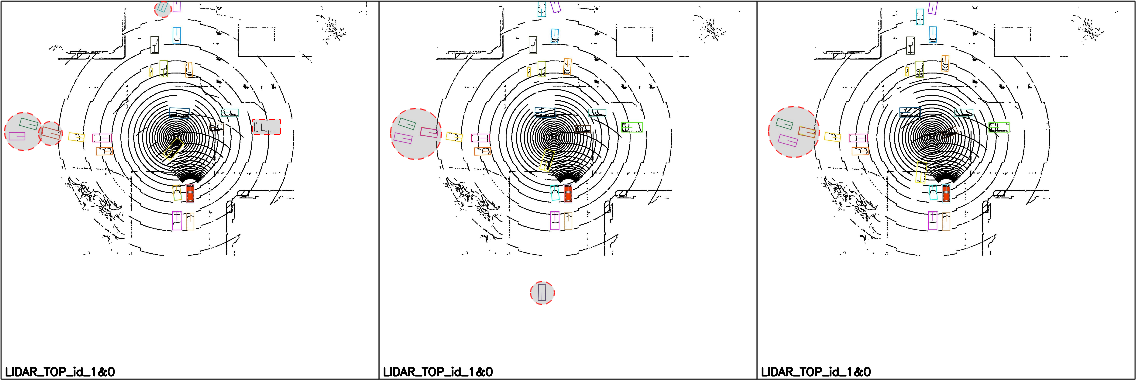}
        \subcaption{LET-V2X}
        \label{fig:v2x-sim-v2i_let-v2x_seq}
    \end{subfigure}
    \\[2mm]
    \begin{subfigure}[b]{0.42\textwidth}
        \includegraphics[width=\linewidth]{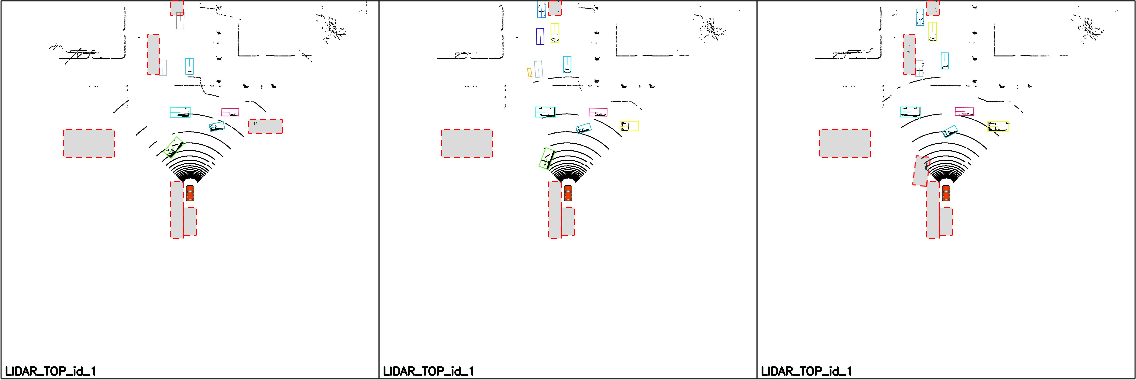}
        \subcaption{XET-V}
        \label{fig:v2x-sim-v2i_xet-v_seq}
    \end{subfigure}
    \\[2mm]
    \begin{subfigure}[b]{0.42\textwidth}
        \includegraphics[width=\linewidth]{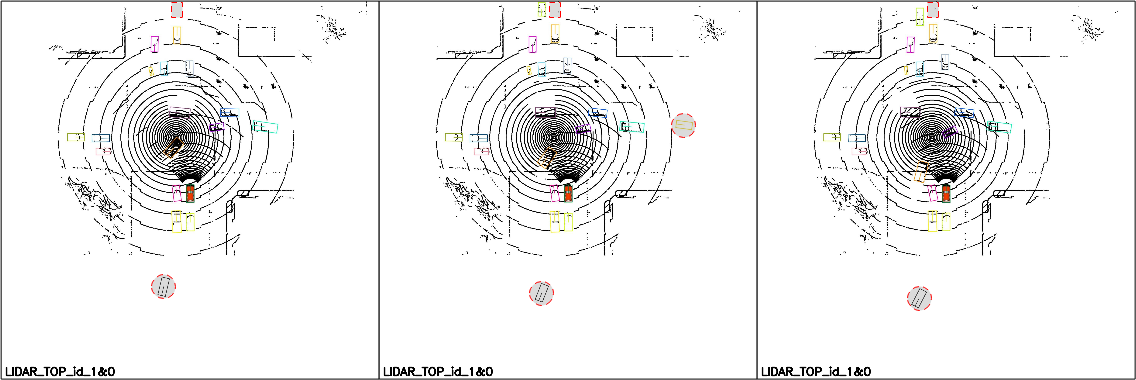}
        \subcaption{XET-V2X}
        \label{fig:v2x-sim-v2i_xet-v2x_seq}
    \end{subfigure}
    \\[2mm]
    \begin{subfigure}[b]{0.42\textwidth}
        \includegraphics[width=\linewidth]{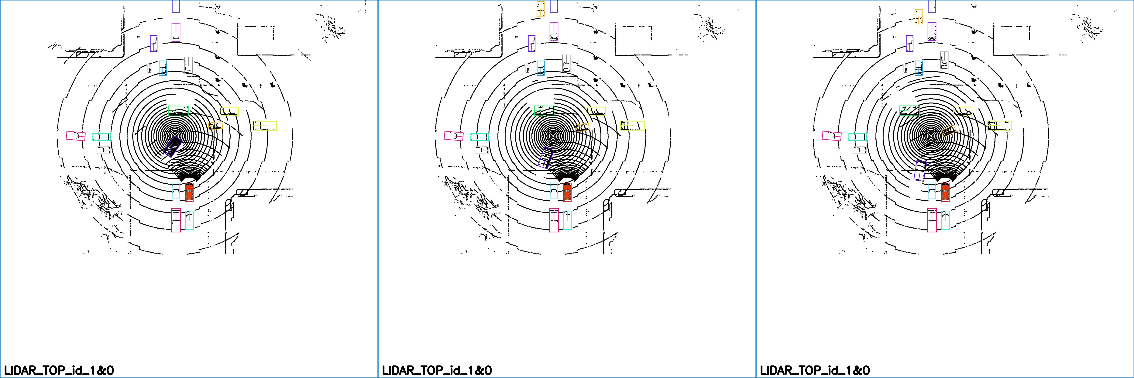}
        \subcaption{Ground Truth}
        \label{fig:v2x-sim-v2i_ground-truth_seq}
    \end{subfigure}
    \caption{\textbf{Sequential visualization of temporal perception results on V2X-Sim-V2I.} 
    Subfigures (a)-(c) and (d)-(f) present the results under the single-vehicle (V) and vehicle-to-everything (V2X) settings, respectively. (g) is the Ground Truth. 
    The three-frame sequence layout, sensor views, and tracking annotations strictly follow the definitions in Fig.~\ref{fig:visul_v2x-seq-spd_seq}.}
    \label{fig:visul_v2x-sim-v2i_seq}
\end{figure}

\clearpage

{
\small
\bibliographystyle{IEEEtran}
\bibliography{IEEEabrv,reference}

\begin{thebibliography}{10}
\providecommand{\url}[1]{#1}
\csname url@samestyle\endcsname
\providecommand{\newblock}{\relax}
\providecommand{\bibinfo}[2]{#2}
\providecommand{\BIBentrySTDinterwordspacing}{\spaceskip=0pt\relax}
\providecommand{\BIBentryALTinterwordstretchfactor}{4}
\providecommand{\BIBentryALTinterwordspacing}{\spaceskip=\fontdimen2\font plus
\BIBentryALTinterwordstretchfactor\fontdimen3\font minus \fontdimen4\font\relax}
\providecommand{\BIBforeignlanguage}[2]{{%
\expandafter\ifx\csname l@#1\endcsname\relax
\typeout{** WARNING: IEEEtran.bst: No hyphenation pattern has been}%
\typeout{** loaded for the language `#1'. Using the pattern for}%
\typeout{** the default language instead.}%
\else
\language=\csname l@#1\endcsname
\fi
#2}}
\providecommand{\BIBdecl}{\relax}
\BIBdecl

\bibitem{chen2019cooper}
Q.~Chen, S.~Tang, Q.~Yang, and S.~Fu, ``Cooper: Cooperative perception for connected autonomous vehicles based on 3d point clouds,'' in \emph{2019 IEEE 39th International Conference on Distributed Computing Systems (ICDCS)}.\hskip 1em plus 0.5em minus 0.4em\relax IEEE, 2019, pp. 514--524.

\bibitem{chen2019f}
Q.~Chen, X.~Ma, S.~Tang, J.~Guo, Q.~Yang, and S.~Fu, ``F-cooper: Feature based cooperative perception for autonomous vehicle edge computing system using 3d point clouds,'' in \emph{Proceedings of the 4th ACM/IEEE Symposium on Edge Computing}, Nov. 2019, pp. 88--100.

\bibitem{wang2020v2vnet}
T.-H. Wang, S.~Manivasagam, M.~Liang, B.~Yang, W.~Zeng, and R.~Urtasun, ``V2vnet: Vehicle-to-vehicle communication for joint perception and prediction,'' in \emph{Computer Vision--ECCV 2020: 16th European Conference, Glasgow, UK, August 23--28, 2020, Proceedings, Part II 16}.\hskip 1em plus 0.5em minus 0.4em\relax Springer, 2020, pp. 605--621.

\bibitem{hu2022where2comm}
Y.~Hu, S.~Fang, Z.~Lei, Y.~Zhong, and S.~Chen, ``Where2comm: Communication-efficient collaborative perception via spatial confidence maps,'' \emph{Advances in neural information processing systems}, vol.~35, pp. 4874--4886, 2022.

\bibitem{yu2023v2x}
H.~Yu, W.~Yang, H.~Ruan, Z.~Yang, Y.~Tang, X.~Gao, X.~Hao, Y.~Shi, Y.~Pan, N.~Sun \emph{et~al.}, ``V2x-seq: A large-scale sequential dataset for vehicle-infrastructure cooperative perception and forecasting,'' in \emph{Proceedings of the IEEE/CVF Conference on Computer Vision and Pattern Recognition}, 2023, pp. 5486--5495.

\bibitem{li2022v2x}
Y.~Li, D.~Ma, Z.~An, Z.~Wang, Y.~Zhong, S.~Chen, and C.~Feng, ``V2x-sim: Multi-agent collaborative perception dataset and benchmark for autonomous driving,'' \emph{IEEE Robotics and Automation Letters}, vol.~7, no.~4, pp. 10\,914--10\,921, 2022.

\bibitem{xu2023v2v4real}
R.~Xu, X.~Xia, J.~Li, H.~Li, S.~Zhang, Z.~Tu, Z.~Meng, H.~Xiang, X.~Dong, R.~Song \emph{et~al.}, ``V2v4real: A real-world large-scale dataset for vehicle-to-vehicle cooperative perception,'' in \emph{Proceedings of the IEEE/CVF Conference on Computer Vision and Pattern Recognition}, Jun. 2023, pp. 13\,712--13\,722.

\bibitem{seeliger2014advisory}
F.~Seeliger, G.~Weidl, D.~Petrich, F.~Naujoks, G.~Breuel, A.~Neukum, and K.~Dietmayer, ``Advisory warnings based on cooperative perception,'' in \emph{2014 IEEE Intelligent Vehicles Symposium Proceedings}.\hskip 1em plus 0.5em minus 0.4em\relax IEEE, 2014, pp. 246--252.

\bibitem{kim2020shared}
Y.~Kim, L.~Onesto, S.~Tay, L.~Yang, J.~Guanetti, S.~Savaresi, and F.~Borrelli, ``Shared perception for connected and automated vehicles,'' in \emph{2020 IEEE Intelligent Vehicles Symposium (IV)}.\hskip 1em plus 0.5em minus 0.4em\relax IEEE, Oct. 2020, pp. 21--26.

\bibitem{ye2020cooperative}
E.~Ye, P.~Spiegel, and M.~Althoff, ``Cooperative raw sensor data fusion for ground truth generation in autonomous driving,'' in \emph{2020 IEEE 23rd International Conference on Intelligent Transportation Systems (ITSC)}.\hskip 1em plus 0.5em minus 0.4em\relax IEEE, 2020, pp. 1--7.

\bibitem{arnold2020cooperative}
E.~Arnold, M.~Dianati, R.~de~Temple, and S.~Fallah, ``Cooperative perception for 3d object detection in driving scenarios using infrastructure sensors,'' \emph{IEEE Transactions on Intelligent Transportation Systems}, vol.~23, no.~3, pp. 1852--1864, 2020.

\bibitem{xiao2018multimedia}
Z.~Xiao, Z.~Mo, K.~Jiang, and D.~Yang, ``Multimedia fusion at semantic level in vehicle cooperactive perception,'' in \emph{2018 IEEE International Conference on Multimedia \& Expo Workshops (ICMEW)}.\hskip 1em plus 0.5em minus 0.4em\relax IEEE, 2018, pp. 1--6.

\bibitem{yee2018collaborative}
R.~Yee, E.~Chan, B.~Cheng, and G.~Bansal, ``Collaborative perception for automated vehicles leveraging vehicle-to-vehicle communications,'' in \emph{2018 IEEE Intelligent Vehicles Symposium (IV)}.\hskip 1em plus 0.5em minus 0.4em\relax IEEE, 2018, pp. 1099--1106.

\bibitem{miller2020cooperative}
A.~Miller, K.~Rim, P.~Chopra, P.~Kelkar, and M.~Likhachev, ``Cooperative perception and localization for cooperative driving,'' in \emph{2020 IEEE International Conference on Robotics and Automation (ICRA)}.\hskip 1em plus 0.5em minus 0.4em\relax IEEE, 2020, pp. 1256--1262.

\bibitem{liu2020who2com}
Y.-C. Liu, J.~Tian, C.-Y. Ma, N.~Glaser, C.-W. Kuo, and Z.~Kira, ``Who2com: Collaborative perception via learnable handshake communication,'' in \emph{2020 IEEE International Conference on Robotics and Automation (ICRA)}.\hskip 1em plus 0.5em minus 0.4em\relax IEEE, May--Aug. 2020, pp. 6876--6883.

\bibitem{liu2020when2com}
Y.-C. Liu, J.~Tian, N.~Glaser, and Z.~Kira, ``When2com: Multi-agent perception via communication graph grouping,'' in \emph{Proceedings of the IEEE/CVF Conference on computer vision and pattern recognition}, 2020, pp. 4106--4115.

\bibitem{hu2024communication}
Y.~Hu, J.~Peng, S.~Liu, J.~Ge, S.~Liu, and S.~Chen, ``Communication-efficient collaborative perception via information filling with codebook,'' in \emph{Proceedings of the IEEE/CVF Conference on Computer Vision and Pattern Recognition}, 2024, pp. 15\,481--15\,490.

\bibitem{huang2021bevdet}
\BIBentryALTinterwordspacing
J.~Huang, G.~Huang, Z.~Zhu, Y.~Ye, and D.~Du, ``Bevdet: High-performance multi-camera 3d object detection in bird-eye-view,'' \emph{arXiv preprint arXiv:2112.11790}, 2021. [Online]. Available: \url{https://arxiv.org/abs/2112.11790}
\BIBentrySTDinterwordspacing

\bibitem{li2023bevdepth}
Y.~Li, Z.~Ge, G.~Yu, J.~Yang, Z.~Wang, Y.~Shi, J.~Sun, and Z.~Li, ``Bevdepth: Acquisition of reliable depth for multi-view 3d object detection,'' in \emph{Proceedings of the AAAI conference on artificial intelligence}, vol.~37, no.~2, 2023, pp. 1477--1485.

\bibitem{bai2022pillargrid}
Z.~Bai, G.~Wu, M.~J. Barth, Y.~Liu, E.~A. Sisbot, and K.~Oguchi, ``Pillargrid: Deep learning-based cooperative perception for 3d object detection from onboard-roadside lidar,'' in \emph{2022 IEEE 25th International Conference on Intelligent Transportation Systems (ITSC)}.\hskip 1em plus 0.5em minus 0.4em\relax IEEE, Oct. 2022, pp. 1743--1749.

\bibitem{liu2022petr}
Y.~Liu, T.~Wang, X.~Zhang, and J.~Sun, ``Petr: Position embedding transformation for multi-view 3d object detection,'' in \emph{European conference on computer vision}.\hskip 1em plus 0.5em minus 0.4em\relax Springer, 2022, pp. 531--548.

\bibitem{liu2023petrv2}
Y.~Liu, J.~Yan, F.~Jia, S.~Li, A.~Gao, T.~Wang, and X.~Zhang, ``Petrv2: A unified framework for 3d perception from multi-camera images,'' in \emph{Proceedings of the IEEE/CVF international conference on computer vision}, Oct. 2023, pp. 3262--3272.

\bibitem{wang2022detr3d}
Y.~Wang, V.~C. Guizilini, T.~Zhang, Y.~Wang, H.~Zhao, and J.~Solomon, ``Detr3d: 3d object detection from multi-view images via 3d-to-2d queries,'' in \emph{Conference on Robot Learning}.\hskip 1em plus 0.5em minus 0.4em\relax PMLR, 2022, pp. 180--191.

\bibitem{li2024bevformer}
Z.~Li, W.~Wang, H.~Li, E.~Xie, C.~Sima, T.~Lu, Q.~Yu, and J.~Dai, ``Bevformer: learning bird's-eye-view representation from lidar-camera via spatiotemporal transformers,'' \emph{IEEE Transactions on Pattern Analysis and Machine Intelligence}, 2024.

\bibitem{yang2023bevformer}
C.~Yang, Y.~Chen, H.~Tian, C.~Tao, X.~Zhu, Z.~Zhang, G.~Huang, H.~Li, Y.~Qiao, L.~Lu \emph{et~al.}, ``Bevformer v2: Adapting modern image backbones to bird's-eye-view recognition via perspective supervision,'' in \emph{Proceedings of the IEEE/CVF conference on computer vision and pattern recognition}, 2023, pp. 17\,830--17\,839.

\bibitem{yu2022dair}
H.~Yu, Y.~Luo, M.~Shu, Y.~Huo, Z.~Yang, Y.~Shi, Z.~Guo, H.~Li, X.~Hu, J.~Yuan \emph{et~al.}, ``Dair-v2x: A large-scale dataset for vehicle-infrastructure cooperative 3d object detection,'' in \emph{Proceedings of the IEEE/CVF Conference on Computer Vision and Pattern Recognition}, 2022, pp. 21\,361--21\,370.

\bibitem{xu2022opv2v}
R.~Xu, H.~Xiang, X.~Xia, X.~Han, J.~Li, and J.~Ma, ``Opv2v: An open benchmark dataset and fusion pipeline for perception with vehicle-to-vehicle communication,'' in \emph{2022 International Conference on Robotics and Automation (ICRA)}.\hskip 1em plus 0.5em minus 0.4em\relax IEEE, 2022, pp. 2583--2589.

\bibitem{lei2022latency}
Z.~Lei, S.~Ren, Y.~Hu, W.~Zhang, and S.~Chen, ``Latency-aware collaborative perception,'' in \emph{European Conference on Computer Vision}.\hskip 1em plus 0.5em minus 0.4em\relax Springer, 2022, pp. 316--332.

\bibitem{yu2023flow}
H.~Yu, Y.~Tang, E.~Xie, J.~Mao, P.~Luo, and Z.~Nie, ``Flow-based feature fusion for vehicle-infrastructure cooperative 3d object detection,'' \emph{Advances in Neural Information Processing Systems}, vol.~36, pp. 34\,493--34\,503, 2023.

\bibitem{dao2024practical}
M.-Q. Dao, J.~S. Berrio, V.~Fr{\'e}mont, M.~Shan, E.~H{\'e}ry, and S.~Worrall, ``Practical collaborative perception: A framework for asynchronous and multi-agent 3d object detection,'' \emph{IEEE Transactions on Intelligent Transportation Systems}, vol.~25, no.~9, pp. 12\,163--12\,175, 2024.

\bibitem{lu2024heal}
Y.~Lu, Y.~Hu, Y.~Zhong, D.~Wang, Y.~Wang, and S.~Chen, ``Heal: An extensible framework for open heterogeneous collaborative perception,'' in \emph{International Conference on Learning Representations (LCLR 2024)}, 2024.

\bibitem{vora2020pointpainting}
S.~Vora, A.~H. Lang, B.~Helou, and O.~Beijbom, ``Pointpainting: Sequential fusion for 3d object detection,'' in \emph{Proceedings of the IEEE/CVF conference on computer vision and pattern recognition}, 2020, pp. 4604--4612.

\bibitem{sindagi2019mvx}
V.~A. Sindagi, Y.~Zhou, and O.~Tuzel, ``Mvx-net: Multimodal voxelnet for 3d object detection,'' in \emph{2019 International Conference on Robotics and Automation (ICRA)}.\hskip 1em plus 0.5em minus 0.4em\relax IEEE, 2019, pp. 7276--7282.

\bibitem{ku2018joint}
J.~Ku, M.~Mozifian, J.~Lee, A.~Harakeh, and S.~L. Waslander, ``Joint 3d proposal generation and object detection from view aggregation,'' in \emph{2018 IEEE/RSJ International Conference on Intelligent Robots and Systems (IROS)}.\hskip 1em plus 0.5em minus 0.4em\relax IEEE, Oct. 2018, pp. 1--8.

\bibitem{wang2021pointaugmenting}
C.~Wang, C.~Ma, M.~Zhu, and X.~Yang, ``Pointaugmenting: Cross-modal augmentation for 3d object detection,'' in \emph{Proceedings of the IEEE/CVF conference on computer vision and pattern recognition}, 2021, pp. 11\,794--11\,803.

\bibitem{nabati2021centerfusion}
R.~Nabati and H.~Qi, ``Centerfusion: Center-based radar and camera fusion for 3d object detection,'' in \emph{Proceedings of the IEEE/CVF Winter Conference on Applications of Computer Vision}, 2021, pp. 1527--1536.

\bibitem{chen2022deformable}
Z.~Chen, Z.~Li, S.~Zhang, L.~Fang, Q.~Jiang, and F.~Zhao, ``Deformable feature aggregation for dynamic multi-modal 3d object detection,'' in \emph{European conference on computer vision}.\hskip 1em plus 0.5em minus 0.4em\relax Springer, 2022, pp. 628--644.

\bibitem{li2022deepfusion}
Y.~Li, A.~W. Yu, T.~Meng, B.~Caine, J.~Ngiam, D.~Peng, J.~Shen, Y.~Lu, D.~Zhou, Q.~V. Le \emph{et~al.}, ``Deepfusion: Lidar-camera deep fusion for multi-modal 3d object detection,'' in \emph{Proceedings of the IEEE/CVF conference on computer vision and pattern recognition}, 2022, pp. 17\,182--17\,191.

\bibitem{pang2020clocs}
S.~Pang, D.~Morris, and H.~Radha, ``Clocs: Camera-lidar object candidates fusion for 3d object detection,'' in \emph{2020 IEEE/RSJ International Conference on Intelligent Robots and Systems (IROS)}.\hskip 1em plus 0.5em minus 0.4em\relax IEEE, 2020, pp. 10\,386--10\,393.

\bibitem{bai2022transfusion}
X.~Bai, Z.~Hu, X.~Zhu, Q.~Huang, Y.~Chen, H.~Fu, and C.-L. Tai, ``Transfusion: Robust lidar-camera fusion for 3d object detection with transformers,'' in \emph{Proceedings of the IEEE/CVF conference on computer vision and pattern recognition}, 2022, pp. 1090--1099.

\bibitem{chen2023futr3d}
X.~Chen, T.~Zhang, Y.~Wang, Y.~Wang, and H.~Zhao, ``Futr3d: A unified sensor fusion framework for 3d detection,'' in \emph{proceedings of the IEEE/CVF conference on computer vision and pattern recognition}, 2023, pp. 172--181.

\bibitem{yang2022deepinteraction}
Z.~Yang, J.~Chen, Z.~Miao, W.~Li, X.~Zhu, and L.~Zhang, ``Deepinteraction: 3d object detection via modality interaction,'' \emph{Advances in Neural Information Processing Systems}, vol.~35, pp. 1992--2005, 2022.

\bibitem{liu2023bevfusion}
Z.~Liu, H.~Tang, A.~Amini, X.~Yang, H.~Mao, D.~L. Rus, and S.~Han, ``Bevfusion: Multi-task multi-sensor fusion with unified bird's-eye view representation,'' in \emph{2023 IEEE international conference on robotics and automation (ICRA)}.\hskip 1em plus 0.5em minus 0.4em\relax IEEE, 2023, pp. 2774--2781.

\bibitem{yan2023cross}
J.~Yan, Y.~Liu, J.~Sun, F.~Jia, S.~Li, T.~Wang, and X.~Zhang, ``Cross modal transformer: Towards fast and robust 3d object detection,'' in \emph{Proceedings of the IEEE/CVF international conference on computer vision}, Oct. 2023, pp. 18\,268--18\,278.

\bibitem{wang2024unibev}
S.~Wang, H.~Caesar, L.~Nan, and J.~F. Kooij, ``Unibev: Multi-modal 3d object detection with uniform bev encoders for robustness against missing sensor modalities,'' in \emph{2024 IEEE Intelligent Vehicles Symposium (IV)}.\hskip 1em plus 0.5em minus 0.4em\relax IEEE, Jun. 2024, pp. 2776--2783.

\bibitem{li2024gafusion}
X.~Li, B.~Fan, J.~Tian, and H.~Fan, ``Gafusion: Adaptive fusing lidar and camera with multiple guidance for 3d object detection,'' in \emph{Proceedings of the IEEE/CVF Conference on Computer Vision and Pattern Recognition}, Jun. 2024, pp. 21\,209--21\,218.

\bibitem{he2025mdnet}
J.~He, X.~Deng, J.~Gui, T.~Zhang, and X.~He, ``Mdnet: Multimodal cooperative perception via spatial alignment of modal decision-making,'' \emph{IEEE Internet of Things Journal}, vol.~12, no.~11, pp. 16\,142--16\,154, 2025.

\bibitem{bewley2016simple}
A.~Bewley, Z.~Ge, L.~Ott, F.~Ramos, and B.~Upcroft, ``Simple online and realtime tracking,'' in \emph{2016 IEEE international conference on image processing (ICIP)}.\hskip 1em plus 0.5em minus 0.4em\relax IEEE, 2016, pp. 3464--3468.

\bibitem{wojke2017simple}
N.~Wojke, A.~Bewley, and D.~Paulus, ``Simple online and realtime tracking with a deep association metric,'' in \emph{2017 IEEE international conference on image processing (ICIP)}.\hskip 1em plus 0.5em minus 0.4em\relax IEEE, 2017, pp. 3645--3649.

\bibitem{weng2020ab3dmot}
\BIBentryALTinterwordspacing
X.~Weng, J.~Wang, D.~Held, and K.~Kitani, ``Ab3dmot: A baseline for 3d multi-object tracking and new evaluation metrics,'' \emph{arXiv preprint arXiv:2008.08063}, 2020. [Online]. Available: \url{https://arxiv.org/abs/2008.08063}
\BIBentrySTDinterwordspacing

\bibitem{benbarka2021score}
N.~Benbarka, J.~Schr{\"o}der, and A.~Zell, ``Score refinement for confidence-based 3d multi-object tracking,'' in \emph{2021 IEEE/RSJ International Conference on Intelligent Robots and Systems (IROS)}.\hskip 1em plus 0.5em minus 0.4em\relax IEEE, Sep.--Oct. 2021, pp. 8083--8090.

\bibitem{pang2022simpletrack}
Z.~Pang, Z.~Li, and N.~Wang, ``Simpletrack: Understanding and rethinking 3d multi-object tracking,'' in \emph{European Conference on Computer Vision}.\hskip 1em plus 0.5em minus 0.4em\relax Springer, 2022, pp. 680--696.

\bibitem{wang2020towards}
Z.~Wang, L.~Zheng, Y.~Liu, Y.~Li, and S.~Wang, ``Towards real-time multi-object tracking,'' in \emph{European conference on computer vision}.\hskip 1em plus 0.5em minus 0.4em\relax Springer, 2020, pp. 107--122.

\bibitem{zhang2021fairmot}
Y.~Zhang, C.~Wang, X.~Wang, W.~Zeng, and W.~Liu, ``Fairmot: On the fairness of detection and re-identification in multiple object tracking,'' \emph{International journal of computer vision}, vol. 129, pp. 3069--3087, 2021.

\bibitem{yin2021center}
T.~Yin, X.~Zhou, and P.~Krahenbuhl, ``Center-based 3d object detection and tracking,'' in \emph{Proceedings of the IEEE/CVF conference on computer vision and pattern recognition}, 2021, pp. 11\,784--11\,793.

\bibitem{zaech2022learnable}
J.-N. Zaech, A.~Liniger, D.~Dai, M.~Danelljan, and L.~Van~Gool, ``Learnable online graph representations for 3d multi-object tracking,'' \emph{IEEE Robotics and Automation Letters}, vol.~7, no.~2, pp. 5103--5110, 2022.

\bibitem{zhang2023motiontrack}
C.~Zhang, C.~Zhang, Y.~Guo, L.~Chen, and M.~Happold, ``Motiontrack: end-to-end transformer-based multi-object tracking with lidar-camera fusion,'' in \emph{Proceedings of the IEEE/CVF Conference on Computer Vision and Pattern Recognition}, 2023, pp. 151--160.

\bibitem{ding2024ada}
S.~Ding, L.~Schneider, M.~Cordts, and J.~Gall, ``Ada-track: End-to-end multi-camera 3d multi-object tracking with alternating detection and association,'' in \emph{Proceedings of the IEEE/CVF Conference on Computer Vision and Pattern Recognition}, Jun. 2024, pp. 15\,184--15\,194.

\bibitem{meinhardt2022trackformer}
T.~Meinhardt, A.~Kirillov, L.~Leal-Taixe, and C.~Feichtenhofer, ``Trackformer: Multi-object tracking with transformers,'' in \emph{Proceedings of the IEEE/CVF conference on computer vision and pattern recognition}, 2022, pp. 8844--8854.

\bibitem{zeng2022motr}
F.~Zeng, B.~Dong, Y.~Zhang, T.~Wang, X.~Zhang, and Y.~Wei, ``Motr: End-to-end multiple-object tracking with transformer,'' in \emph{European Conference on Computer Vision}.\hskip 1em plus 0.5em minus 0.4em\relax Springer, 2022, pp. 659--675.

\bibitem{piergiovanni20214d}
A.~Piergiovanni, V.~Casser, M.~S. Ryoo, and A.~Angelova, ``4d-net for learned multi-modal alignment,'' in \emph{Proceedings of the IEEE/CVF International Conference on Computer Vision}, 2021, pp. 15\,435--15\,445.

\bibitem{chang2024recurrentbev}
M.~Chang, X.~Zhang, R.~Zhang, Z.~Zhao, G.~He, and S.~Liu, ``Recurrentbev: A long-term temporal fusion framework for multi-view 3d detection,'' in \emph{European Conference on Computer Vision}.\hskip 1em plus 0.5em minus 0.4em\relax Springer, 2024, pp. 131--147.

\bibitem{zhou2024v2xpnp}
\BIBentryALTinterwordspacing
Z.~Zhou, H.~Xiang, Z.~Zheng, S.~Z. Zhao, M.~Lei, Y.~Zhang, T.~Cai, X.~Liu, J.~Liu, M.~Bajji \emph{et~al.}, ``V2xpnp: Vehicle-to-everything spatio-temporal fusion for multi-agent perception and prediction,'' \emph{arXiv preprint arXiv:2412.01812}, 2024. [Online]. Available: \url{https://arxiv.org/abs/2412.01812}
\BIBentrySTDinterwordspacing

\bibitem{wang2023exploring}
S.~Wang, Y.~Liu, T.~Wang, Y.~Li, and X.~Zhang, ``Exploring object-centric temporal modeling for efficient multi-view 3d object detection,'' in \emph{Proceedings of the IEEE/CVF international conference on computer vision}, 2023, pp. 3621--3631.

\bibitem{lin2023sparse4dv3}
\BIBentryALTinterwordspacing
X.~Lin, Z.~Pei, T.~Lin, L.~Huang, and Z.~Su, ``Sparse4d v3: Advancing end-to-end 3d detection and tracking,'' \emph{arXiv preprint arXiv:2311.11722}, 2023. [Online]. Available: \url{https://arxiv.org/abs/2311.11722}
\BIBentrySTDinterwordspacing

\bibitem{yang2025letvic}
Z.~Yang, J.~Mao, W.~Yang, Y.~Ai, Y.~Kong, H.~Yu, and W.~Zhang, ``Lidar-based end-to-end temporal perception for vehicle-infrastructure cooperation,'' \emph{IEEE INTERNET OF THINGS JOURNAL}, vol.~12, no.~13, pp. 22\,862--22\,874, Jul. 2025.

\bibitem{lang2019pointpillars}
A.~H. Lang, S.~Vora, H.~Caesar, L.~Zhou, J.~Yang, and O.~Beijbom, ``Pointpillars: Fast encoders for object detection from point clouds,'' in \emph{Proceedings of the IEEE/CVF conference on computer vision and pattern recognition}, Jun. 2019, pp. 12\,697--12\,705.

\bibitem{he2016deep}
K.~He, X.~Zhang, S.~Ren, and J.~Sun, ``Deep residual learning for image recognition,'' in \emph{Proceedings of the IEEE conference on computer vision and pattern recognition}, 2016, pp. 770--778.

\bibitem{fan2023quest}
\BIBentryALTinterwordspacing
S.~Fan, H.~Yu, W.~Yang, J.~Yuan, and Z.~Nie, ``Quest: Query stream for vehicle-infrastructure cooperative perception,'' \emph{arXiv preprint arXiv:2308.01804}, 2023. [Online]. Available: \url{https://arxiv.org/abs/2308.01804}
\BIBentrySTDinterwordspacing

\bibitem{chen2023transiff}
Z.~Chen, Y.~Shi, and J.~Jia, ``Transiff: An instance-level feature fusion framework for vehicle-infrastructure cooperative 3d detection with transformers,'' in \emph{Proceedings of the IEEE/CVF International Conference on Computer Vision}, Oct. 2023, pp. 18\,205--18\,214.

\bibitem{feng2023dense}
S.~Feng, H.~Sun, X.~Yan, H.~Zhu, Z.~Zou, S.~Shen, and H.~X. Liu, ``Dense reinforcement learning for safety validation of autonomous vehicles,'' \emph{Nature}, vol. 615, no. 7953, pp. 620--627, 2023.

\bibitem{vaswani2017attention}
A.~Vaswani, N.~Shazeer, N.~Parmar, J.~Uszkoreit, L.~Jones, A.~N. Gomez, {\L}.~Kaiser, and I.~Polosukhin, ``Attention is all you need,'' \emph{Advances in neural information processing systems}, vol.~30, 2017.

\bibitem{rukhovich2022imvoxelnet}
D.~Rukhovich, A.~Vorontsova, and A.~Konushin, ``Imvoxelnet: Image to voxels projection for monocular and multi-view general-purpose 3d object detection,'' in \emph{Proceedings of the IEEE/CVF winter conference on applications of computer vision}, 2022, pp. 2397--2406.

\end{thebibliography}
}

\end{document}